\documentclass{article}

\usepackage[main, final]{neurips_2026}

\makeatletter
\renewcommand{\@noticestring}{}
\makeatother

\usepackage[utf8]{inputenc} 
\usepackage[T1]{fontenc}    
\usepackage{hyperref}       
\usepackage{url}            
\usepackage{booktabs}       
\usepackage{amsfonts}       
\usepackage{nicefrac}       
\usepackage{microtype}      
\usepackage{xcolor}         
\usepackage{amsmath}
\usepackage{booktabs}    
\usepackage{tabularx}    
\usepackage{multirow}    
\usepackage{array}       
\usepackage{colortbl}    
\usepackage{xcolor}      
\usepackage{arydshln}    
\usepackage{booktabs}
\usepackage{tabularx}
\usepackage{multirow}
\usepackage[table]{xcolor}
\usepackage{arydshln}
\usepackage{booktabs}
\usepackage{multirow}
\usepackage[table]{xcolor}
\usepackage[most]{tcolorbox}
\usepackage{makecell}
\usepackage{array}
\usepackage{adjustbox}
\usepackage{array}
\usepackage{graphicx}
\usepackage[normalem]{ulem}
\usepackage{booktabs}
\usepackage{pifont}
\usepackage{xcolor}
\usepackage{tabularx}
\newcommand{\cmark}{\color{green}\ding{51}}%
\newcommand{\xmark}{\color{red}\ding{55}}%
\usepackage{xcolor}

\definecolor{groupgray}{RGB}{245,246,248}
\definecolor{gainbg}{RGB}{234,247,238}
\definecolor{lossbg}{RGB}{253,238,238}
\definecolor{gainfg}{RGB}{46,125,80}
\definecolor{lossfg}{RGB}{180,70,70}

\newtcbox{\gainboxfixed}{
  on line,
  boxrule=0pt,
  colback=gainbg,
  colframe=gainbg,
  arc=2.2pt,
  left=1pt,right=1pt,top=0.8pt,bottom=0.8pt,
  boxsep=0pt,
  width=3.4em,
  halign=center,
  nobeforeafter
}
\newtcbox{\lossboxfixed}{
  on line,
  boxrule=0pt,
  colback=lossbg,
  colframe=lossbg,
  arc=2.2pt,
  left=1pt,right=1pt,top=0.8pt,bottom=0.8pt,
  boxsep=0pt,
  width=3.4em,
  halign=center,
  nobeforeafter
}
\newcommand{\pctgain}[1]{{\scriptsize\gainboxfixed{\textcolor{gainfg}{+#1\%}}}}
\newcommand{\pctloss}[1]{{\scriptsize\lossboxfixed{\textcolor{lossfg}{-#1\%}}}}

\definecolor{groupbackground}{RGB}{245,245,245}
\definecolor{highlightcolor}{RGB}{230,245,255}
\definecolor{lightblueweak}{RGB}{242,249,253}

\definecolor{dlpink}{RGB}{255, 105, 180}

\lstdefinestyle{plainoutput}{
  basicstyle=\ttfamily\small,      
  breaklines=true,                 
  columns=fullflexible,            
  showstringspaces=false,          
  keepspaces=true,                 
  frame=none,                      
  numbers=none,                    
  morekeywords={},                 
  keywordstyle={},                 
  commentstyle={},                 
  stringstyle={},                  
  escapeinside={},                 
   breakindent=0pt                  
}
\usepackage[linesnumbered,ruled,vlined]{algorithm2e}
\usepackage{placeins}

\title{
\begin{minipage}[c]{0.07\textwidth}
\centering
\includegraphics[height=2.1cm]{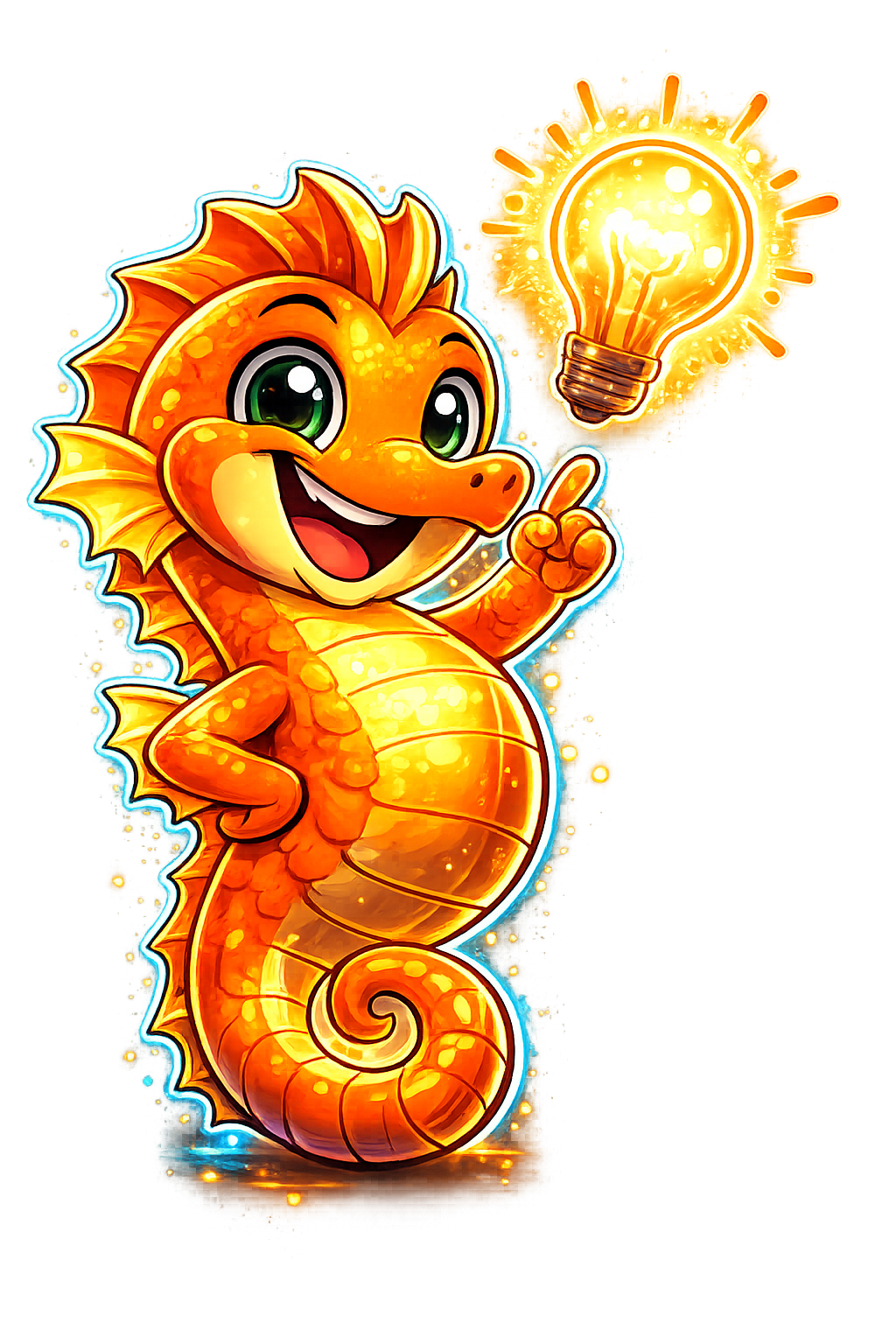}
\end{minipage}
\hfill
\begin{minipage}[c]{0.90\textwidth}
\raggedright
\LARGE \textbf{\textsc{HippoSpark}: An On-Demand Experience System for LLM Reasoning}
\end{minipage}
}

\author{%
\begin{tabular}{c}
Jingyao Liu\textsuperscript{1,3} \quad
Danling Meng\textsuperscript{3,4} \quad
Chen Huang\textsuperscript{5} \quad
Yukun Yan\textsuperscript{2}\thanks{Corresponding authors.} \\
Zhenghao Liu\textsuperscript{6} \quad
Wenqiang Lei\textsuperscript{1,3}\footnotemark[1] \quad
See-Kiong Ng\textsuperscript{5} \quad
Maosong Sun\textsuperscript{2}
\end{tabular}
\vspace{0.5em}
\\
\begin{tabular}{c}
\small \textsuperscript{1}College of Computer Science, Sichuan University \\
\small \textsuperscript{2}Department of Computer Science and Technology, Tsinghua University \\
\small \textsuperscript{3}Engineering Research Center of Machine Learning and Industry Intelligence,\\
\small Ministry of Education, China \\
\small \textsuperscript{4}College of Software Engineering, Sichuan University \\
\small \textsuperscript{5}Institute of Data Science, National University of Singapore \\
\small \textsuperscript{6}Northeastern University \\
\small \texttt{liujingyao1@stu.scu.edu.cn}
\end{tabular}
}

\begin{document}

\maketitle

\begin{abstract}
Distilling historical trajectories into reusable experience to enhance future problem-solving has become a focal point of recent LLM research. However, existing methods predominantly operate at the task level, leveraging general summaries or rules under the assumption that analogous tasks share universal solution patterns. This approach often fails in complex reasoning, which typically falters at local bottlenecks that require precise, state-specific guidance rather than broad heuristics. We introduce \textsc{HippoSpark}, a state-level experience system that performs on-demand retrieval tailored to the immediate needs of the current reasoning state. Across mathematical, scientific, and programming benchmarks, \textsc{HippoSpark} consistently outperforms both standard prompting and task-level experience baselines. Our findings reveal that the most effective experience systems are those that provide actionable guidance at critical bottlenecks rather than serving as generic task-level context. Our code is available at \url{https://github.com/DanlingMeng/HippoSpark}.
\end{abstract}

\section{Introduction}
\label{sec:intro}

Recent advances in large language models (LLMs) have driven impressive performance on complex reasoning tasks such as mathematical reasoning and scientific question answering~\citep{wei2022chain,yang2025qwen3,guo2025deepseekr1}.
However, most current systems treat each problem in isolation, discarding the reasoning trajectory once a final answer is generated. This represents a missed opportunity: past trajectories often contain problem-solving experience that can support future reasoning, such as useful directions, effective transformations, and failure cases under similar difficulties~\citep{gick1980analogical,gentner1983structure,kolodner1993case}.
This motivates the need for an \textbf{experience system}~\citep{zhao2024expel,wang2025awm,liu2025cer,ma2025agentrr}, where past trajectories are retained and distilled into experience to support future problem solving.

However, building a robust experience system requires moving beyond the \textbf{task-level} reuse seen in existing work~\citep{shinn2023reflexion,zhao2024expel,wang2025awm,liu2025cer,ma2025agentrr,chen2024automanual,suzgun2026dynamic,ouyang2025reasoningbank,fang2025memp,wu2025evolver}. Traditionally, existing systems employ general summaries or rules from past trajectories as experience, assuming that similar tasks share identical high-level solution patterns. While useful, this broad approach is ill-suited for complex reasoning. In particular, reasoning unfolds through intermediate states~\citep{newell1972human,schank1982dynamic,kolodner1993case,yao2023tree}, where experience is often {needed only at a local bottleneck: a state in which several next moves appear plausible, but it is unclear which one can make progress~\citep{yao2023tree,wang2022selfconsistency}. Therefore, rather than characterizing macroscopic task-level similarity, the system must reflect \textbf{state-level} similarity to provide step-specific guidance that helps the model advance. As illustrated in Figure~\ref{fig:intro}, consider an LLM solving a constrained optimization problem. The model may become stuck at an intermediate reasoning state $S_1$, unsure of how to resolve a specific sub-problem (i.e., \textit{how to transform the exponential expression}). Retrieving task-level experience in this scenario might yield generic insights about constrained optimization, but it will likely lack the precise tactical guidance required to overcome the specific bottleneck faced at state $S_1$. Therefore, \textbf{compared to overly coarse task-level reuse, state-level experience provides far more targeted and instructive guidance}.


Mirroring the human \underline{hippo}campal system, which retrieves relevant past experiences in response to ongoing cognitive demands \citep{miller2001integrative,preston2013interplay}, we propose \textsc{\textbf{HippoSpark}}, a state-level experience system that triggers on-demand retrieval based on the specific needs of the current state to \uline{spark} the reasoning. Specifically, as illustrated in Figure~\ref{fig:intro}(b), \textsc{HippoSpark} consists of two primary modules: \textit{on-demand experience utilization (OEU)} and \textit{bottleneck-focused experience construction (BEC)}. The former monitors the uncertainty of the model's next move at the current state and triggers retrieval accordingly. The latter establishes the system's foundation by distilling past trajectories into state-level experience bank, which is achieved by extracting bottleneck reasoning states and mapping them to their corresponding decision cues and supporting details. As such, \textsc{HippoSpark} helps the model successfully navigate bottleneck states where traditional task-level systems would fail.

We evaluate \textsc{HippoSpark} on mathematical, scientific, and code reasoning benchmarks across various LLM backbones. Our experimental results demonstrate that \textsc{HippoSpark} consistently outperforms task-level experience baselines (+{43.48}\% on average). Further analyses validate that the OEU module facilitates more deliberate decision-making by leveraging retrieved experience at critical reasoning state, while the BEC module enhances actionability by providing decision cues with execution-level support. These findings underscore that experience is most effective when grounded in local state transitions where the model's next move is most uncertain.

\begin{figure}[t]
    \centering
    \includegraphics[width=\textwidth]{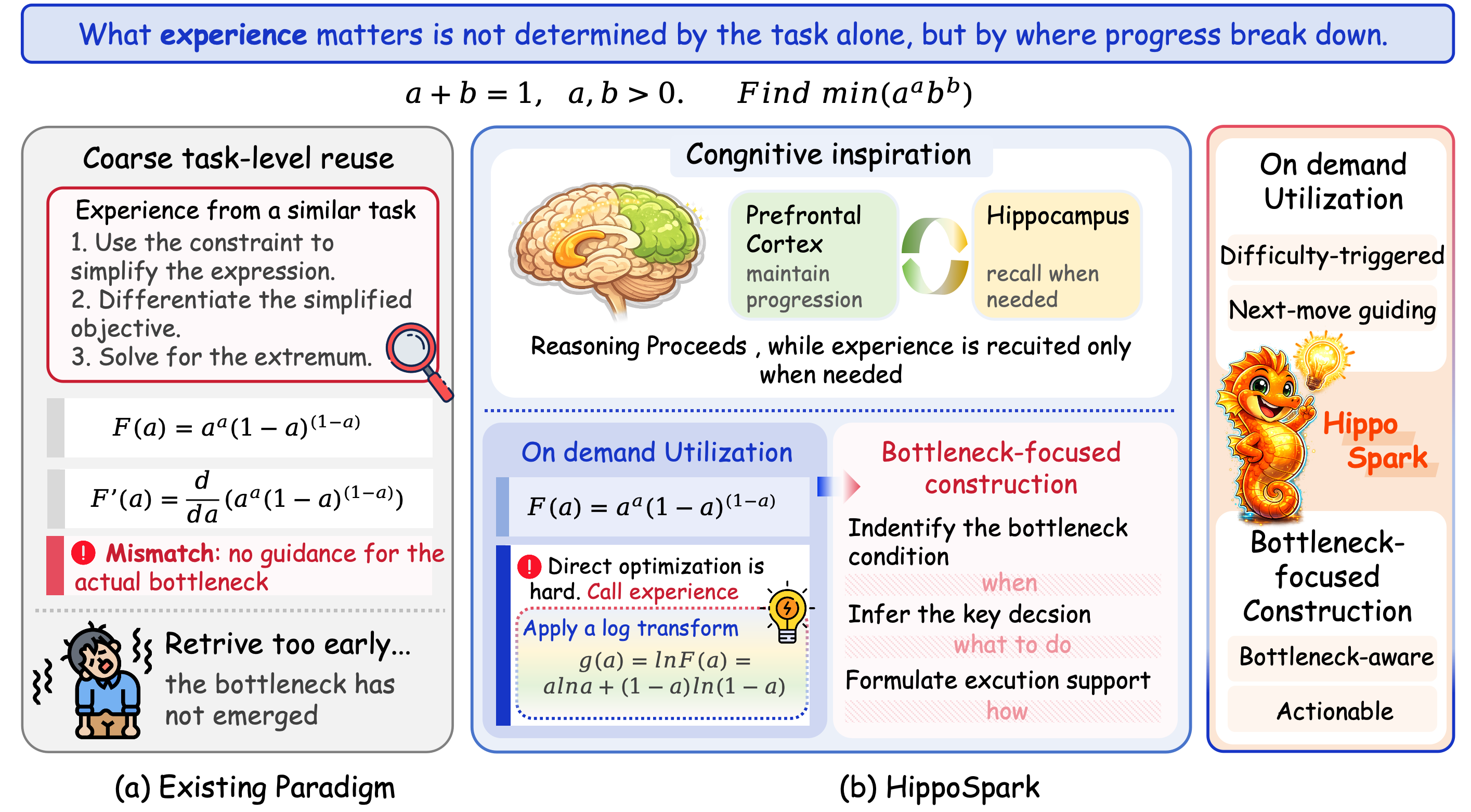}
    \caption{
\textbf{Motivation and overview of \textsc{HippoSpark}.}
Existing experience systems rely on task-level reuse, which is often too coarse to help when reasoning gets stuck at a specific intermediate state.
Inspired by hippocampal recall under uncertainty, \textsc{HippoSpark} provides targeted state-level guidance by invoking experience only when the next move is unclear and constructing actionable experience for passing the current bottleneck.
}
    \label{fig:intro}
\end{figure}

\section{Related work}
\label{sec:related_work}

\begin{table}[t]
\centering
\caption{Comparison of representative experience-based methods.
\textit{Type} indicates the level at which experience supports reasoning:
\textbf{T} = task-level and \textbf{S} = state-level.
\textit{Dem.} denotes on-demand experience use, and
\textit{Bott.} denotes bottleneck-focused experience construction.
}
\label{tab:related_work_compare}
\small
\setlength{\tabcolsep}{3.0pt}
\renewcommand{\arraystretch}{1.08}
\begin{tabularx}{\columnwidth}{lc cc X}
\toprule
\textbf{Method} & \textbf{Type} & \textbf{Dem.} & \textbf{Bott.} & \textbf{Notes} \\
\midrule
\rowcolor{groupbackground}
\multicolumn{5}{c}{\textit{Trial-derived feedback}} \\
Reflexion~\citep{shinn2023reflexion}
& T & - & - 
& Store reflections; reuse them across trials \\
Self-Refine~\citep{madaan2023selfrefine}
& T & - & - 
& Generate feedback for iterative refinement \\

\rowcolor{groupbackground}
\multicolumn{5}{c}{\textit{Trajectory experience}} \\
ExpRAG~\citep{ferraz2026exprag}
& T & \xmark & \xmark 
& Past trajectories, task-level support \\
AWM~\citep{wang2025awm}
& T & \xmark & \xmark 
& Summarize workflows, task-level support \\
Mem\textsuperscript{p}~\citep{fang2025memp}
& T & \xmark & \xmark 
& Distill procedures, task-level support \\

\rowcolor{groupbackground}
\multicolumn{5}{c}{\textit{Distilled experience}} \\
Dyn. Cheatsheet~\citep{suzgun2026dynamic}
& T & \xmark  & \cmark
& Strategy snippets, task-level support \\
ReasoningBank~\citep{ouyang2025reasoningbank}
& T & \xmark  & \cmark 
& Reasoning strategies, task-level support \\

\rowcolor{highlightcolor}
\textbf{\textsc{HippoSpark} (Ours)}
& S & \cmark & \cmark 
& Bottleneck-centered, on-demand support \\
\bottomrule
\end{tabularx}
\end{table}

Our work is closely related to experience systems that retain past reasoning or agent trajectories for future problem solving.
Building an effective experience system involves two central questions: 
\emph{experience utilization}, which concerns when and how experience is retrieved and used, and 
\emph{experience construction}, which concerns what trajectory information is distilled into experience.
Table~\ref{tab:related_work_compare} summarizes how existing methods and \textsc{HippoSpark} differ along these two aspects.

\noindent\textbf{Experience Utilization.}
Existing experience systems generally follow \emph{task-level reuse}, where experience from the same or similar tasks is used as context to guide generation.
This includes intra-task feedback, such as reflections from previous attempts~\citep{shinn2023reflexion,madaan2023selfrefine}, and inter-task retrieval based on overall task similarity~\citep{zhao2024expel,wang2025awm,liu2025cer,ma2025agentrr,ferraz2026exprag,chen2024automanual,suzgun2026dynamic,ouyang2025reasoningbank,fang2025memp,wu2025evolver}.
Despite different sources, these methods provide experience as task-level context for the current task.
In contrast, \textsc{HippoSpark} invokes experience on demand at uncertain intermediate states and uses it only to guide the immediate next move.


\noindent\textbf{Experience Construction.}
Within the paradigm of task-level reuse, experience construction has evolved through a spectrum of increasing abstraction, which leads to three types of methods. At the most specific level, \uline{trial-derived feedback} captures reflections tied to individual task revisions~\citep{shinn2023reflexion,madaan2023selfrefine}. More modularly, \uline{trajectory experience} abstracts solutions into structural workflows or procedures, though it remains dependent on task similarity~\citep{zhao2024expel,wang2025awm,liu2025cer,ma2025agentrr,ferraz2026exprag}. Most recently, \uline{distilled experience} extracts compact strategy snippets or procedural instructions to maximize cross-task portability~\citep{chen2024automanual,suzgun2026dynamic,ouyang2025reasoningbank,fang2025memp,wu2025evolver}. Despite this progression, these constructions are fundamentally designed for whole-task retrieval. \textsc{HippoSpark} departs from this by reorienting construction toward state-level bottlenecks, distilling trajectories into actionable units that specifically address the immediate demands of local reasoning states.

\section{\textsc{HippoSpark}}
\label{sec:method}

\begin{figure}[t]
    \centering
    \includegraphics[width=\textwidth]{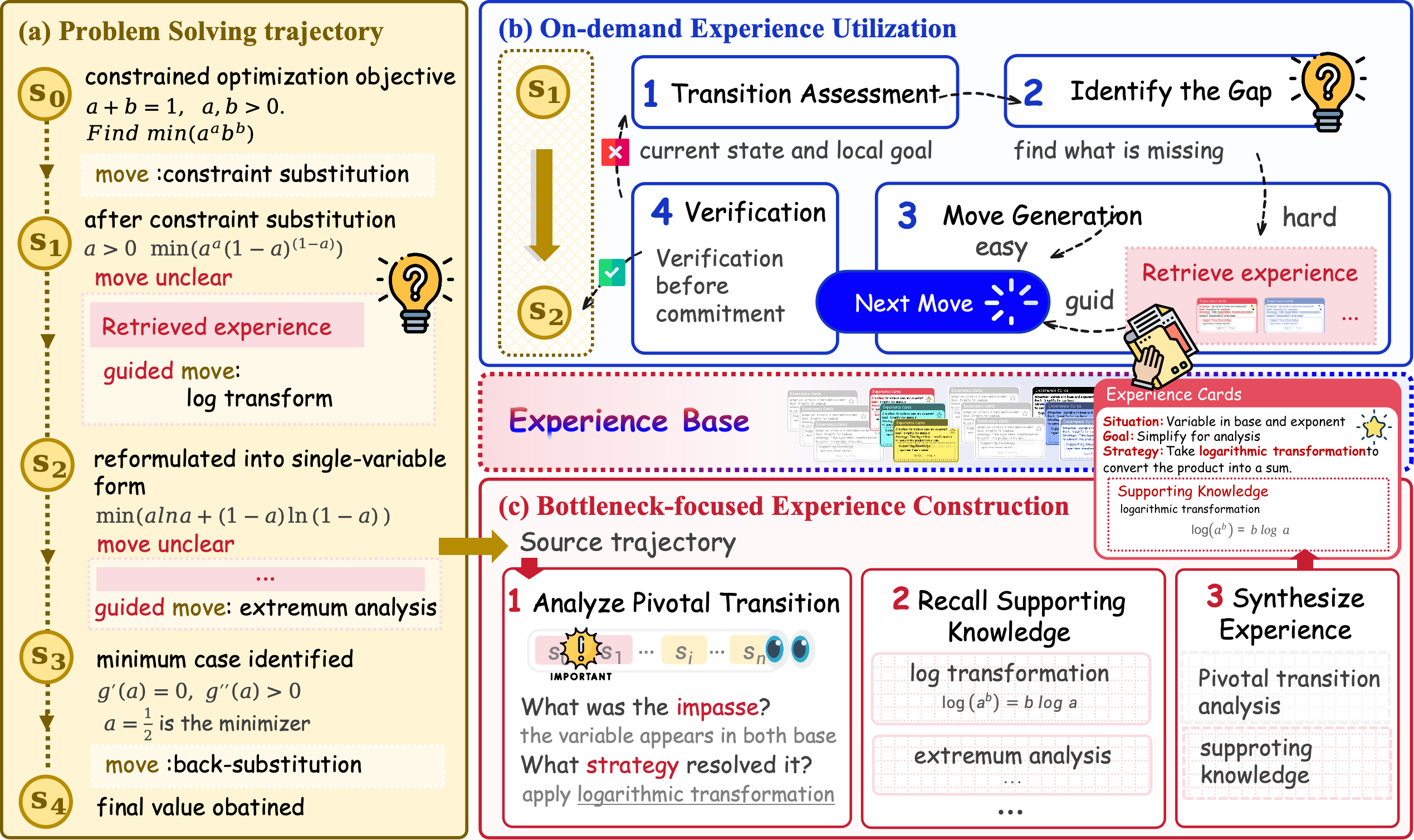}
    \caption{
\textbf{Overview of HippoSpark.}
\textbf{(a)} Reasoning proceeds through states and moves.
\textbf{(b)} Experience guides next moves at local transitions.
\textbf{(c)} Pivotal prior transitions are distilled into two layers: decision cards and execution knowledge.
}
    \label{fig:framework_overview}
\end{figure}

\noindent\textbf{Problem Formulation \& Overview.}
Given an input problem $x$, a target goal $G$, and an experience bank $\mathcal{E}$, an experience-augmented LLM reasoner generates a reasoning trajectory $\tau$ through a utilization function $U$:
$\tau = U(x,G,\mathcal{E})=(s_0,m_0,s_1,\ldots,m_{T-1},s_T)$,
where $s_t$ denotes the intermediate reasoning state at step $t$, and $m_t$ represents the move that effects the transition $s_t \xrightarrow{m_t} s_{t+1}$. The terminal state $s_T$ provides the final solution. Additionally, the experience bank $\mathcal{E}$ is derived from a corpus of historical reasoning trajectories $\mathcal{T}$ via a construction function $C$, such that $\mathcal{E}=C(\mathcal{T})=\{e_i\}_{i=1}^{N}$. During this process, the fundamental challenge lies in designing $U$ and $C$ to ensure that experience facilitates reasoning at the granular level of state transitions. As illustrated in Figure~\ref{fig:framework_overview}, \textsc{HippoSpark} addresses this by employing two coupled components: 1) The OEU module instantiates the function $U$. It dynamically determines the necessity of experience at any given state $s_t$ and integrates retrieved knowledge to guide the transition $m_t$. 2) The BEC module instantiates the function $C$. It distills $\mathcal{T}$ into a set of actionable experience units $\{e_i\}$, specifically targeting reusable bottleneck transitions where reasoning typically falters.

\subsection{On-demand Experience Utilization (OEU)}
\label{sec:experience_utilization}


The OEU module determines when experience is necessary through a four-stage process, illustrated in Figure~\ref{fig:framework_overview}. First, it evaluates the current state $s_t$ to produce a sub-goal. Second, it identifies if the next move is uncertain by measuring the gaps of state transition; if no gap exists, it generates $m_t$ independently. Otherwise, it retrieves relevant experience before producing the next move. Finally, a verifier ensures the proposed transition is grounded and productive before it is committed.
\noindent\textbf{Transition assessment.}
At each step, the OEU first summarizes what has been established in the current reasoning state and what should be achieved next.
Given the current state $s_t$ and the final goal $G$, it produces
\[
(c_t,g_t)=L_{\mathrm{assess}}(s_t,G),
\]
where $c_t$ describes the current condition, i.e., what is known or has been derived so far, and $g_t$ specifies the next sub-goal that should be achieved toward $G$.
This assessment provides the basis for deciding whether the next move is already clear or requires experience support.

For example, in Figure~\ref{fig:framework_overview}(a), at state $s_1$, the solver has rewritten the objective as $a^a(1-a)^{1-a}$, while the final goal is to find its minimum value.
In this state, $c_t$ records that the current objective is a product with variables appearing in the exponents, and $g_t$ is to find a transformation that makes the extremum easier to analyze.

\noindent\textbf{Gap identification.}
After identifying what has been established and what should be achieved next, the OEU decides whether the next move is clear enough to proceed directly.
Given the current condition $c_t$ and the next sub-goal $g_t$, it analyzes how to move from $c_t$ to $g_t$:
\[
\Delta_t=L_{\mathrm{gap}}(c_t,g_t), 
\qquad
r_t=L_{\mathrm{route}}(\Delta_t)\in\{\texttt{direct},\texttt{experience}\}.
\]
Here, $\Delta_t$ summarizes whether there is a clear operation for reaching $g_t$ from $c_t$.
If such an operation exists, such as routine simplification, substitution, numerical computation, or direct formula application, the solver takes the \texttt{direct} route.
Otherwise, it takes the \texttt{experience} route to seek guidance for the next move.

For example, in Figure~\ref{fig:framework_overview}(a), the solver has obtained $a^a(1-a)^{1-a}$, and the next sub-goal is to make it suitable for extremum analysis.
This step requires identifying a useful reformulation strategy, rather than applying a routine operation.
Since the strategy is not immediately clear, the OEU routes this step to experience retrieval.

\noindent\textbf{Move generation.}
Given the route $r_t$, the OEU generates a candidate move $\tilde{m}_t$.
For the \texttt{direct} route, the move is generated from the assessed condition and local goal:
\[
\tilde{m}_t=L_{\mathrm{dir}}(c_t,g_t).
\]
If $r_t=\texttt{experience}$, the OEU retrieves experience from similar situation--goal pairs and uses it to generate the move:
\[
E_t=\operatorname{TopK}\big(\operatorname{Retrieve}((c_t,g_t),\mathcal{E})\big),
\qquad
\tilde{m}_t=L_{\mathrm{exp}}(c_t,g_t,E_t).
\]
Here, $(c_t,g_t)$ serves as the retrieval query, so the retrieved experience comes from similar state transitions, where a solver moved from a similar current condition toward a similar next sub-goal.

For example, in Figure~\ref{fig:framework_overview}(a), the retrieved experience suggests using a logarithmic transformation when optimizing products with variables in the exponents.
Guided by this experience, the OEU sets $y=a^a(1-a)^{1-a}$ and rewrites the objective as
$\log y = a\log a + (1-a)\log(1-a)$,
making it amenable to standard extremum analysis.

Importantly, the retrieved experience is used only to generate the current move.
After $\tilde{m}_t$ is produced, $E_t$ is not kept in the subsequent reasoning context.
Only the generated move is considered for commitment to the trajectory.
This keeps experience local to the current state and reduces distraction from irrelevant retrieved context~\citep{shi2023large,yoran2024making}.

\noindent\textbf{Verification before commitment.}
Before committing a generated move to the reasoning trajectory, the OEU checks whether the move is grounded in the current state and productive for the next sub-goal.
Groundedness means that the move must be valid under the current state, such as satisfying positivity conditions for logarithmic transformations or handling boundary cases in optimization.
Productiveness means that the move should make progress toward $g_t$, rather than introduce irrelevant manipulations.
We use
\[
v_t=L_{\mathrm{verify}}(c_t,g_t,\tilde{m}_t),
\]
where $v_t\in\{\texttt{accept},\texttt{reject}\}$ indicates whether the move passes verification.
If accepted, the move is committed to the reasoning trajectory:
\[
s_{t+1}=\mathrm{Apply}(s_t,\tilde{m}_t).
\]
Here, $\mathrm{Apply}$ appends the accepted move to the reasoning chain and updates the state; if rejected, verifier feedback is used to regenerate the move.

Overall, the the OEU module realizes on-demand experience use.
Experience is invoked only when the current step lacks a reliable next move, used only to guide that immediate move, and discarded after the move is generated.
Refer to Algorithm~\ref{alg:utilization} for the pseudo-code.

\subsection{Bottleneck-focused Experience Construction (BEC)}
\label{sec:experience_construction}


To maximize utility, experience must be \textit{bottleneck-focused} to target only those critical states where reasoning stalls, and \textit{actionable} to provide the specific tactical guidance required for execution. This dual requirement ensures the system avoids redundant retrieval during trivial steps while providing enough granular detail to bridge complex transition gaps. To this end, the BEC module is designed to transform historical trajectories $\mathcal{T}$ to an experience bank $\mathcal{E}=\{e_i\}_{i=1}^{N}$ of experience cards, defined as
\[
e_i=\big((c_i,g_i),\,d_i,\,\mathcal{K}_i\big),
\]
where $(c_i,g_i)$ specifies the situation--goal anchor, $d_i$ denotes the strategy cue for the next move, and $\mathcal{K}_i$ is the {supporting knowledge} (e.g., rules or constraints) required for execution. This structure directly aligns construction with utilization: the anchor identifies the bottleneck, while the cue and knowledge drive the transition. For example, a card may anchor a "\textit{variables in exponents}" situation to a "\textit{logarithmic transformation}" strategy, supplemented by the specific log rules needed to simplify the expression (Figure~\ref{fig:framework_overview}c). Formally, our experience cards are constructed through the following three-stage pipeline.



\noindent\textbf{Analyze pivotal transitions.}
The BEC first analyzes complete historical trajectories to identify pivotal transitions, i.e., local state transitions where the next move is non-trivial and necessary for solution progress. Formally, The BEC produces a list of transition analyses:
\[
\{\alpha_i\}_{i=1}^{M}=L_{\mathrm{analyze}}(\tau,G).
\]
Each $\alpha_i$ records the information needed for later card construction, including the bottleneck context, the local goal, the strategy cue for resolving the transition, and the supporting knowledge topics $\mathcal{H}_i$ required for executing the strategy.

\noindent\textbf{Obtain supporting knowledge.}
The BEC then obtains the execution support needed to apply the extracted strategy.
Given the supporting knowledge topics $\mathcal{H}_i$ from $\alpha_i$, \textsc{HippoSpark} retrieves the corresponding knowledge entries:
\[
\mathcal{K}_i=\{L_{\mathrm{support}}(h)\mid h\in\mathcal{H}_i\}.
\]
The obtained support may include formulas, transformation rules, constraints, or diagnostic checks.

\noindent\textbf{Synthesize experience cards.}
Finally, the BEC rewrites the transition analyses into generalized experience cards and links them with the obtained support:
\[
\mathcal{E}_{\tau}
=
L_{\mathrm{card}}(\{\alpha_i\}_{i=1}^{M}, \{\mathcal{K}_i\}_{i=1}^{M})
=
\{e_i\}_{i=1}^{M},
\qquad
e_i=\big((c_i,g_i),\,d_i,\,\mathcal{K}_i\big).
\]
Here, $L_{\mathrm{card}}$ abstracts away task-specific details from each transition analysis and produces a reusable situation--goal anchor $(c_i,g_i)$ and strategy cue $d_i$.
The supporting knowledge $\mathcal{K}_i$ is linked to the corresponding card as the execution support for applying the strategy.

\section{Experiment}
\begin{table*}[t]
\centering
\caption{
Main results.
AIME uses five generations per problem; GPQA-Diamond and BigCodeBench-Hard use three runs.
Overall averages AIME24 Avg@1, AIME25 Avg@1, GPQA-Diamond aggregate accuracy, and BigCodeBench Pass@1. The
best and second-best are \textbf{bold} and \uline{underlined}.
}
\label{tab:main_results}
\renewcommand{\arraystretch}{1.12}
\setlength{\tabcolsep}{3.8pt}

\resizebox{0.98\textwidth}{!}{
\begin{tabular}{lcccccccccc}
\toprule

\multirow{3}{*}{\textbf{Method}}
& \multicolumn{4}{c}{\textbf{Math Reasoning}}
& \multicolumn{3}{c}{\textbf{Scientific Reasoning}}
& \multicolumn{2}{c}{\textbf{Code Generation}}
& \multirow{3}{*}{\textbf{Overall}} \\

\cmidrule(lr){2-5} \cmidrule(lr){6-8} \cmidrule(lr){9-10}

& \multicolumn{2}{c}{\textbf{AIME24}}
& \multicolumn{2}{c}{\textbf{AIME25}}
& \multicolumn{3}{c}{\textbf{GPQA-Diamond}}
& \multicolumn{2}{c}{\textbf{BigCodeBench}} \\

\cmidrule(lr){2-3} \cmidrule(lr){4-5} \cmidrule(lr){6-8} \cmidrule(lr){9-10}

& \textbf{Avg@1}$^{\uparrow}$ & \textbf{Best@5}$^{\uparrow}$
& \textbf{Avg@1}$^{\uparrow}$ & \textbf{Best@5}$^{\uparrow}$
& $\mathbf{Acc}_{\text{Phy.}}^{\uparrow}$ & $\mathbf{Acc}_{\text{Chem.}}^{\uparrow}$ & $\mathbf{Acc}_{\text{Bio.}}^{\uparrow}$
& \textbf{Pass@1}$^{\uparrow}$ & \textbf{Pass@3}$^{\uparrow}$ \\
\midrule

\rowcolor{groupbackground}
\multicolumn{11}{c}{\textbf{\textit{Qwen3-32B}}} \\
Vanilla
& 0.253 & 0.267 & 0.147 & 0.200
 & 0.705 & 0.313 & 0.643
& 0.231 & 0.352
& 0.287 \\
CoT
& 0.280 & 0.300 & 0.180 & 0.200
 & \underline{0.716} & 0.328 & \underline{0.738}
& 0.222 & 0.315
& 0.305 \\
Self-Refine
& 0.307 & \underline{0.367} & 0.247 & \underline{0.300}
 & 0.689 & 0.354 & \underline{0.738}
& 0.176 & 0.222
& 0.317 \\
\cdashline{1-11}
Trajectory Exp.
& 0.267 & 0.367 & 0.193 & 0.233
 & 0.503 & \underline{0.364} & 0.667
& \underline{0.244} & 0.343
& 0.290 \\
ReasoningBank
& 0.260 & 0.267 & 0.207 & 0.233
 & 0.536 & 0.253 & 0.643
& 0.167 & 0.222
& 0.262 \\
DC-Cu
& 0.307 & 0.333 & 0.227 & \underline{0.300}
& 0.519 & 0.354 & \textbf{0.762}
& 0.176 & 0.278
& 0.294 \\
DC-RS
& 0.293 & 0.333 & \underline{0.253} & \underline{0.300}
& 0.530 & 0.333 & 0.667
& 0.185 & 0.333
& 0.296 \\
\rowcolor{lightblueweak}
Ours (w/o Exp)
& \underline{0.393} & \underline{0.400} & \underline{0.253} & 0.267
& 0.699 & 0.323 & 0.643
& 0.231 & \underline{0.361}
& \underline{0.349} \\
\rowcolor{highlightcolor}
\textbf{Ours}
& \textbf{0.487} & \textbf{0.533} & \textbf{0.313} & \textbf{0.367}
& \textbf{0.732} & \textbf{0.369} & \underline{0.738}
& \textbf{0.272} & \textbf{0.398}
& \textbf{0.409} \\

\midrule

\rowcolor{groupbackground}
\multicolumn{11}{c}{\textbf{\textit{Qwen3-14B}}} \\
Vanilla
& 0.240 & 0.267 & 0.213 & 0.267
& 0.579 & 0.359 & 0.548
& \underline{0.207} & 0.287
& 0.283 \\
CoT
& 0.320 & 0.333 & 0.227 & 0.267
& \underline{0.639} & 0.318 & 0.619
& 0.185 & 0.259
& 0.305 \\
Self-Refine
& 0.287 & 0.300 & 0.233 & \underline{0.300}
& 0.590 & 0.333 & \underline{0.714}
& 0.176 & 0.222
& 0.295 \\
\cdashline{1-11}
Trajectory Exp.
& 0.227 & 0.267 & 0.200 & 0.233
& 0.536 & 0.323 & 0.571
& 0.176 & 0.231
& 0.261 \\
ReasoningBank
& 0.207 & 0.267 & 0.233 & \underline{0.300}
& 0.536 & 0.313 & 0.619
& 0.145 & 0.185
& 0.256 \\
DC-Cu
& 0.327 & 0.367 & \underline{0.253} & 0.267
& 0.530 & 0.298 & 0.595
& 0.157 & 0.213
& 0.291 \\
DC-RS
& 0.247 & 0.333 & 0.247 & 0.267
& 0.443 & \underline{0.374} & \textbf{0.786}
& 0.176 & 0.269
& 0.279 \\
\rowcolor{lightblueweak}
\textbf{Ours (w/o Exp)}
& \underline{0.387} & \underline{0.433} & 0.247 & \underline{0.300}
& 0.623 & 0.323 & 0.690
& 0.194 & \underline{0.306}
& \underline{0.330} \\
\rowcolor{highlightcolor}
\textbf{Ours}
& \textbf{0.420} & \textbf{0.467} & \textbf{0.300} & \textbf{0.333}
& \textbf{0.694} & \textbf{0.389} & 0.595
& \textbf{0.210} & \textbf{0.343}
& \textbf{0.368} \\

\midrule

\rowcolor{groupbackground}
\multicolumn{11}{c}{\textbf{\textit{GPT-5.4 mini}}} \\
Vanilla
& 0.273 & 0.300 & 0.253 & 0.300
& 0.634 & 0.414 & \underline{0.667}
& 0.244 & 0.306
& 0.326 \\
CoT
& 0.353 & 0.400 & 0.327 & 0.367
& 0.634 & 0.444 & \underline{0.667}
& 0.278 & 0.315
& 0.377 \\
Self-Refine
& 0.313 & 0.367 & 0.300 & 0.333
& 0.590 & 0.429 & 0.595
& 0.281 & 0.352
& 0.352 \\
\cdashline{1-11}
Trajectory Exp.
& 0.133 & 0.200 & 0.180 & 0.233
& 0.459 & 0.394 & 0.619
& 0.262 & 0.333
& 0.255 \\
ReasoningBank
& 0.320 & 0.367 & 0.260 & 0.267
& 0.393 & 0.399 & 0.476
& 0.241 & 0.278
& 0.306 \\
DC-Cu
& 0.360 & 0.400 & 0.227 & 0.267
& 0.454 & 0.404 & 0.643
& 0.241 & 0.315
& 0.319 \\
DC-RS
& 0.306 & 0.333 & 0.187 & 0.267
& 0.574 & 0.359 & \textbf{0.738}
& 0.287 & 0.352
& 0.318 \\
\rowcolor{lightblueweak}
\textbf{Ours (w/o Exp)}
& \underline{0.387} & \underline{0.500} & \underline{0.347} & \underline{0.400}
& \underline{0.776} & \underline{0.455} & 0.619
& \underline{0.290} & \underline{0.379}
& \underline{0.409} \\
\rowcolor{highlightcolor}
\textbf{Ours}
& \textbf{0.507} & \textbf{0.600} & \textbf{0.413} & \textbf{0.500}
& \textbf{0.852} & \textbf{0.510} & \underline{0.667}
& \textbf{0.336} & \textbf{0.444}
& \textbf{0.483} \\

\bottomrule
\end{tabular}
}

\end{table*}

\subsection{Experimental Setup}
\label{sec:exp_setup}
Full benchmark details, implementation settings, and prompts are provided in Appendix~\ref{app:implementation}.

\noindent\textbf{Benchmarks \& Experience Construction.}
{We evaluate \textsc{HippoSpark} on four benchmarks covering mathematical reasoning, scientific reasoning, and code generation: \uline{AIME24} and \uline{AIME25}~\citep{di_zhang_2025,huggingfaceh4_aime_2024,huggingfaceh4_aime_2025}, \uline{GPQA-Diamond}~\citep{rein2024gpqa}, and \uline{BigCodeBench-Hard}~\citep{zhuo2025bigcodebench}.
For experience construction, we follow the protocol of \citet{ouyang2025reasoningbank}: experience is derived from past problem-solving instances and evaluation is conducted strictly on a disjoint set of unseen data.
See Appendix~\ref{app:bench_data} for details.}



\noindent\textbf{Baselines \& Backbones.} In addition to standard prompting (denoted as \uline{Vanilla}) and chain-of-thought reasoning (\uline{CoT})~\citep{wei2022chain}, our baselines encompasses all three categories (cf. Table \ref{tab:related_work_compare}), specifically including \uline{Self-Refine}~\citep{madaan2023selfrefine}, \uline{Trajectory Experience}, \uline{ReasoningBank}~\citep{ouyang2025reasoningbank}, \uline{DC-Cu}~\citep{suzgun2026dynamic}, and \uline{DC-RS}~\citep{suzgun2026dynamic}. For our ablation analysis, we report both the transition-centric framework without external experience (\uline{HippoSpark w/o Exp}) and the full experience-enhanced version (\uline{HippoSpark}). We conduct experiments on two backbone families, {Qwen3} and {GPT-5.4}.

\noindent\textbf{Metrics.}
Following~\citet{lewkowycz2022solving},
we report accuracy-based metrics: \uline{Avg@1}/\uline{Best@5} over five generations on AIME, domain-wise accuracy averaged over three runs on GPQA-Diamond, and \uline{Pass@1}/\uline{Pass@3} on BigCodeBench-Hard.
We also report an \uline{Overall} score by averaging AIME24 Avg@1, AIME25 Avg@1, GPQA-Diamond aggregate accuracy, and BigCodeBench-Hard Pass@1.

\subsection{Main Results}
\label{sec:main_results}

{Table~\ref{tab:main_results} demonstrates that \textsc{HippoSpark} achieves the best Overall score across all evaluated backbones (Qwen3-32B, Qwen3-14B, and GPT-5.4-mini), improving upon the best baselines by {+29\%, +21\%, and +28\%}, respectively. Unlike other experience-augmented methods that sometimes underperform the Vanilla or CoT, our on-demand retrieval and experience construction designs guarantee stable improvements, an advantage we analyze in depth in Section~\ref{sec:analysis}. While our method is robust across most benchmarks, GPQA-Biology remains a minor exception. This is primarily due to the benchmark's small dataset size, which limits experience construction and results in sparse coverage and less reliable retrieval (see Appendix~\ref{app:gpqa_biology_analysis} for further analysis). } 
\subsection{Ablation Study}
\label{sec:ablation_study}




\begin{table*}[t]
\centering
\caption{
Ablation study.
We report Avg@1 and Best@5 to evaluate the contribution of the transition-based reasoning framework, verification before commitment, and experience guidance.
}
\label{tab:math_ablation_main}
\footnotesize
\setlength{\tabcolsep}{2.2pt}
\renewcommand{\arraystretch}{1.06}
\begin{adjustbox}{width=0.95\textwidth}
\begin{tabular}{>{\raggedright\arraybackslash}p{2.0cm}*{8}{r@{\hspace{2pt}}l}}
\toprule
\multirow{3}{*}{\textbf{Method}}
& \multicolumn{8}{c}{\textbf{Qwen3-32B}}
& \multicolumn{8}{c}{\textbf{GPT-5.4 mini}} \\
\cmidrule(lr){2-9} \cmidrule(lr){10-17}
& \multicolumn{4}{c}{\textbf{AIME24}}
& \multicolumn{4}{c}{\textbf{AIME25}}
& \multicolumn{4}{c}{\textbf{AIME24}}
& \multicolumn{4}{c}{\textbf{AIME25}} \\
\cmidrule(lr){2-5} \cmidrule(lr){6-9}
\cmidrule(lr){10-13} \cmidrule(lr){14-17}
& \multicolumn{2}{c}{\small\textbf{Avg@1}$^{\uparrow}$}
& \multicolumn{2}{c}{\small\textbf{Best@5}$^{\uparrow}$}
& \multicolumn{2}{c}{\small\textbf{Avg@1}$^{\uparrow}$}
& \multicolumn{2}{c}{\small\textbf{Best@5}$^{\uparrow}$}
& \multicolumn{2}{c}{\small\textbf{Avg@1}$^{\uparrow}$}
& \multicolumn{2}{c}{\small\textbf{Best@5}$^{\uparrow}$}
& \multicolumn{2}{c}{\small\textbf{Avg@1}$^{\uparrow}$}
& \multicolumn{2}{c}{\small\textbf{Best@5}$^{\uparrow}$} \\
\midrule

\rowcolor{groupgray}
\multicolumn{17}{l}{\textit{\textbf{Framework ablations}}} \\

Vanilla
& 0.253 & & 0.267 & & 0.147 & & 0.200 &
& 0.273 & & 0.300 & & 0.253 & & 0.300 & \\

CoT
& 0.280 & & 0.300 & & 0.180 & & 0.200 &
& 0.353 & & 0.400 & & 0.327 & & 0.367 & \\
\cdashline{1-17}

+ Transition
& 0.333 & & 0.367 & & 0.220 & & 0.267 &
& 0.347 & & 0.467 & & 0.327 & & 0.400 & \\

+ Verification
& 0.393 & & 0.400 & & 0.253 & & 0.267 &
& 0.387 & & 0.500 & & 0.347 & & 0.400 & \\

\textbf{Ours}
& \textbf{0.487} & \pctgain{24}
& \textbf{0.533} & \pctgain{33}
& \textbf{0.313} & \pctgain{24}
& \textbf{0.367} & \pctgain{38}
& \textbf{0.507} & \pctgain{31}
& \textbf{0.600} & \pctgain{20}
& \textbf{0.413} & \pctgain{19}
& \textbf{0.500} & \pctgain{25} \\
\bottomrule
\end{tabular}
\end{adjustbox}
\end{table*}

We conduct ablation studies on \textsc{HippoSpark}, focusing on the components of its {OEU} module that directly affect reasoning-time performance.
The results are shown in Table~\ref{tab:math_ablation_main}.
Starting from Vanilla, ``+Transition'' corresponds to \emph{Transition Assessment + Move Generation}, which explicitly assesses the current reasoning state before generating each move.
``+Transition'' improves over Vanilla across most settings.
Compared with CoT, it also achieves better performance with only a modest increase in output token consumption, e.g., from 3041 to 3414 average output tokens on AIME with Qwen3-32B.
This suggests that \uline{explicit state assessment helps the model make more grounded move decisions}.
The gain is less stable for stronger models such as GPT-5.4 mini, where CoT may already perform strong implicit state assessment.
``+Verification'' further adds \emph{Verification before Commitment} on top of ``+Transition''.
It consistently improves Avg@1 across models and benchmarks, with smaller gains on Best@5.
This suggests that verification mainly prevents ungrounded or unproductive moves from entering the reasoning trajectory, such as steps caused by overlooked assumptions or boundary cases.
Thus, \uline{verification mainly improves reasoning consistency by preventing invalid moves from being committed, rather than substantially expanding problem coverage}.
Finally, the full {OEU} module enables experience-guided move generation when the next move is unclear, bringing consistent gains in both Avg@1 and Best@5.
This suggests that the \uline{OEU solves additional cases through experience guidance while preserving verification's stability benefit}.
Further analysis of why \textsc{HippoSpark} works as an effective experience system is provided in Section~\ref{sec:analysis}.


\begin{figure}[t]
    \centering
    \includegraphics[width=\textwidth]{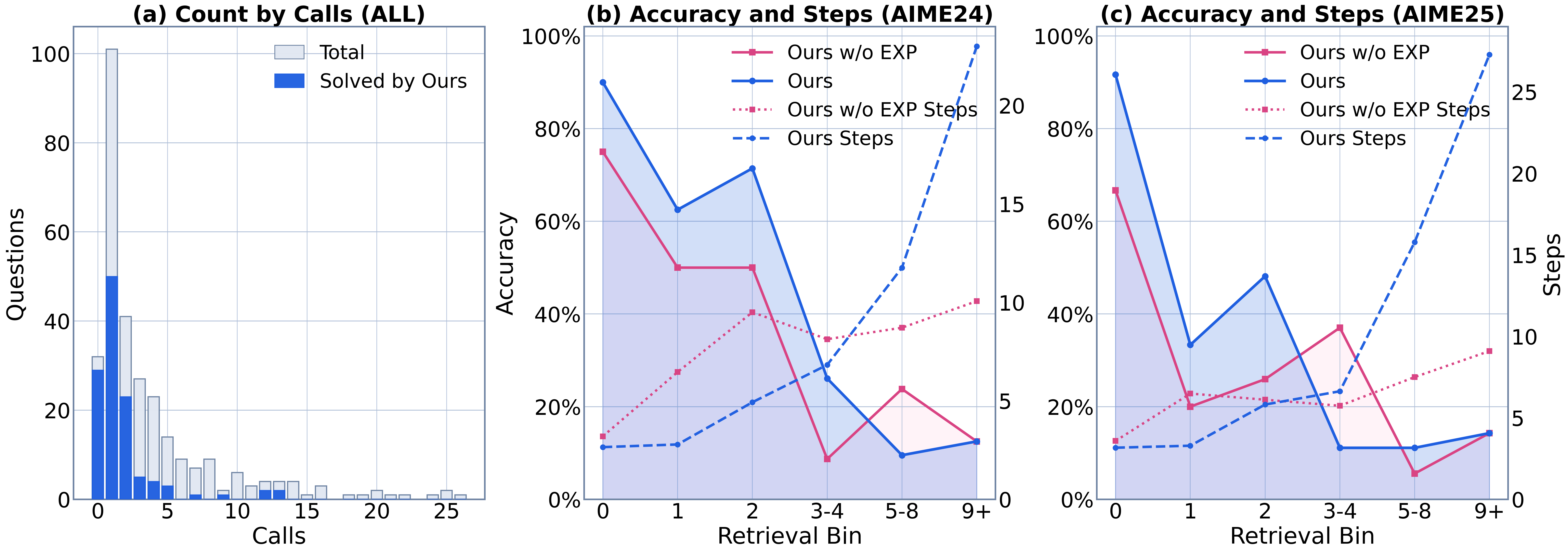}
    \caption{Difficulty-triggered retrieval on AIME24/25. (a) Retrieval-call distribution. (b,c) Accuracy by retrieval-frequency bin; dashed curves on the secondary axis show the average reasoning steps.}
    \label{fig:exp1}
\end{figure}

\subsection{Understanding the Effectiveness of \textsc{HippoSpark}}
\label{sec:analysis}

We analyze the mechanisms behind \textsc{HippoSpark}'s effectiveness along three dimensions: when experience is invoked, how retrieved experience is integrated, and what information an experience unit must capture. Unless otherwise specified, analyses are conducted with Qwen3-32B on AIME24/25 using 300 problem-solving instances. We also analyze the token cost of \textsc{HippoSpark}'s experience construction and on-demand utilization in Appendices~\ref{app:experience_construction_analysis} and~\ref{app:inference_token_analysis}.

\noindent\textbf{On-demand utilization: experience is invoked when local reasoning becomes difficult.}
To evaluate the adaptiveity of \textsc{HippoSpark}, we analyzed performance relative to the frequency of experience calls. As shown in Figure~\ref{fig:exp1}, retrieval follows a highly non-uniform distribution: most problems require minimal intervention, while a small subset of difficult cases necessitates frequent calls. Notably, instances with higher retrieval counts correspond to those where the baseline \textsc{Ours (w/o Exp)} exhibits lower accuracy and longer reasoning paths, confirming that \textsc{HippoSpark} successfully concentrates its resources on the most challenging reasoning bottlenecks. 
Our analysis also reveals that the routing mechanism provides utility even when retrieval is not triggered. In the zero-retrieval subset, \textsc{HippoSpark} outperforms the baseline with fewer reasoning steps. This suggests that the routing decision serves as a critical state-assessment step; by evaluating whether its current state is sufficient to proceed, the model clarifies its internal reasoning condition before committing to a move.



However, the benefit of on-demand retrieval depends on whether the model can identify a clear local obstacle.
In low-retrieval regions, experience often resolves a concrete bottleneck and shortens the reasoning trajectory.
In high-retrieval regions, gains become smaller or even negative, and reasoning steps increase sharply.
Our trace analysis suggests that excessive retrieval often reflects an unclear state assessment: when the model mischaracterizes the current state or subgoal, retrieval is guided by the wrong formulation and may reinforce an unproductive direction. See Appendix~\ref{app:on_demand} for further analysis and case studies.
Thus, on-demand retrieval is most effective when the model can formulate what it is stuck on at the transition level.

\begin{figure}[t]
    \centering
    \includegraphics[width=\textwidth]{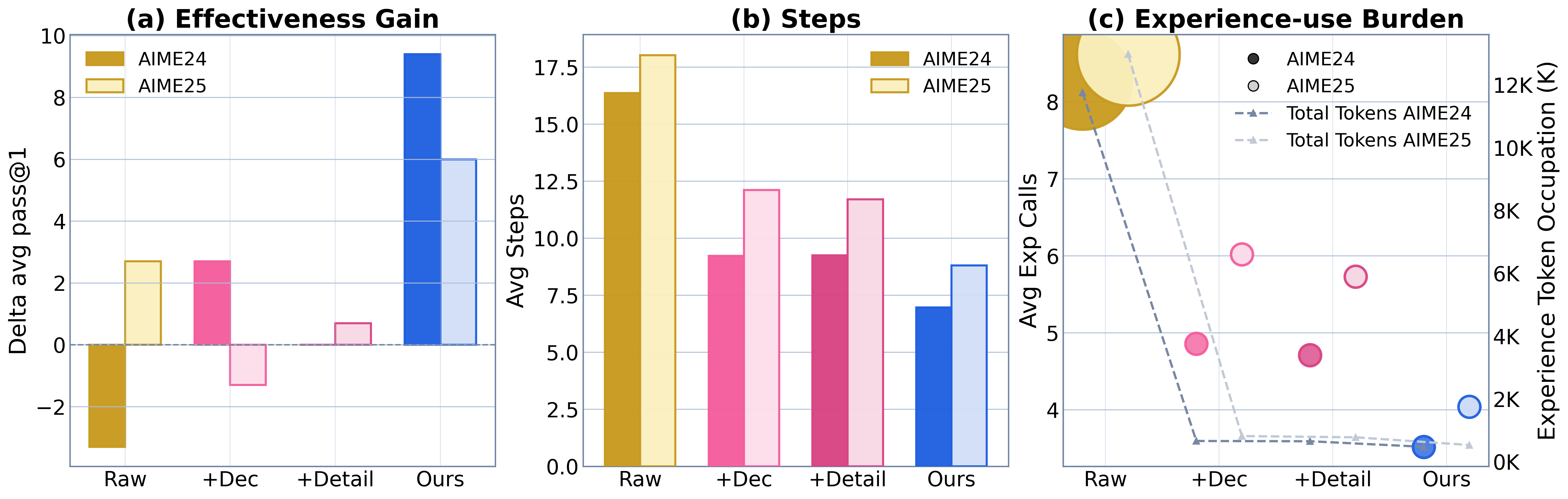}
    \caption{Effects of experience integration and content design. Raw directly inserts retrieved experience into the solving context; Ours uses next-move guidance. +Dec and +Detail keep only decision cues or execution details. Panels show (a) gain over Ours w/o Exp, (b) solving steps, and (c) retrieval calls with injected-token burden.}
    \label{fig:exp2}
\end{figure}

\noindent\textbf{Next-move-guided integration: retrieved experience should be grounded before entering the trajectory.}
To study how retrieved experience should be integrated, we compare \textsc{HippoSpark} with a raw-integration variant in Figure~\ref{fig:exp2}.
Raw integration directly inserts retrieved experience into the solving context, whereas \textsc{HippoSpark} uses it only to generate a next move for the current state.
Only the generated move is verified and committed to the trajectory.
Raw integration performs worse, requires more reasoning steps and experience calls, and injects substantially more experience tokens into the main reasoning context.
The key issue is that raw experience is not yet grounded in the current transition.
Although it do come from a similar past situation, the solver must still decide how to map it to the current variables, constraints, and subgoal.
This extra interpretation burden can lead the model to imitate a past solution pattern without properly adapting it, producing underspecified or misaligned moves.
In contrast, \textsc{HippoSpark} converts retrieved experience into a concrete next-move proposal before commitment, so experience acts as targeted transition-level guidance rather than persistent context. See Appendix~\ref{app:next_move_guiding} for further analysis and case studies.

\noindent\textbf{Bottleneck-focused and actionable construction: experience should record difficult states and support their next moves.}
According to Table~\ref{tab:main_results}, among prior experience-based baselines, methods with a more bottleneck-focused form of experience generally outperform trajectory-level experience reuse.
This suggests that useful experience should be constructed around difficult reasoning states that are likely to recur, rather than around complete past trajectories (cf. Appendix~\ref{app:bottleneck-focused}).
\textsc{HippoSpark} follows this principle and further makes each experience unit actionable for the next move. 

We further evaluate the synergy between components by comparing \textsc{HippoSpark} with two partial-experience variants (Figure~\ref{fig:exp2}):
1) \textsc{+Dec}, which keeps only decision-level cues such as the local difficulty, intended direction, or common traps;
and 2) \textsc{+Detail}, which keeps only execution-level support such as formulas, transformations, implementation details, or API usage.
Both variants underperform \textsc{HippoSpark} and require more reasoning steps and experience calls, showing that neither component is sufficient alone.
The two variants reveal complementary limitations.
Decision-only experience captures the intended direction of a transition, but lacks the operational details required to instantiate that direction in the current state, such as the relevant relation, transformation, or intermediate quantity.
Detail-only experience provides executable knowledge, but without a decision rationale, its relevance to the current reasoning state is under-specified, making it easier to apply the knowledge mechanically or pursue unnecessary subproblems.
\textsc{HippoSpark} addresses both limitations by coupling decision cues with execution support:
decision cues ground the experience in the current bottleneck, while execution support turns the selected direction into an executable next move. See Appendix~\ref{app:actionabel} for further analysis and case studies.

\section{Conclusion}
We advocate for state-level experience that provides actionable guidance at reasoning bottlenecks rather than static, task-level context. Our framework, \textsc{HippoSpark}, implements this through on-demand retrieval and bottleneck-centered construction, delivering both strategy cues and execution support at critical reasoning states. These results suggest that experience systems should be organized around discrete reasoning bottlenecks rather than whole-task reuse. More broadly, our work provides a preliminary exploration into the formalization of reasoning experience. While the optimal structure for representing and leveraging such experience remains an open question, we call for further research across diverse tasks and domains. By empirically uncovering additional properties of experience, the community can move toward a unified theory and identify the most effective architectures for experience-augmented reasoning.


\bibliographystyle{plainnat}
\bibliography{main}
\clearpage
\appendix

\section{Limitations}
\label{app:limitations}

\paragraph{Inference efficiency.}
\textsc{HippoSpark} improves reasoning by invoking experience during the reasoning process, but this design also introduces additional computational overhead.
Unlike task-level reuse methods that typically retrieve experience once before generation, \textsc{HippoSpark} may repeatedly assess the current reasoning state, retrieve relevant experience, generate an experience-guided next move, and verify the candidate move.
This leads to higher latency and a larger input-token burden, especially for harder problems that trigger more experience calls.
We provide a detailed inference-time token analysis in Appendix~\ref{app:inference_token_analysis} and analyze the token consumption of experience construction in Appendix~\ref{app:experience_construction_analysis}.

\paragraph{Experience management.}
This work focuses on when experience should be invoked and what information should be distilled into an actionable experience unit.
However, a complete experience system should also support autonomous memory management, such as updating outdated experience, deleting redundant or misleading experience, and reorganizing experience as the problem distribution changes.
\textsc{HippoSpark} does not explicitly model these management operations.
We leave dynamic experience maintenance and lifecycle management as important directions for future work.

\section{Broader Impacts}
\label{app:broader_impacts}

This work is a methodological study on experience-guided LLM reasoning.
By showing how past reasoning trajectories can be distilled and reused at local bottlenecks, \textsc{HippoSpark} provides a perspective that may be useful for broader research on LLM agents, skill acquisition, and self-evolving systems.
A potential positive impact is to help future systems improve from prior problem-solving experience rather than treating each task as an isolated episode.

At the same time, experience-based systems may also raise broader concerns when integrated into more autonomous agents.
The ability to accumulate and reuse experience may increase the capability and persistence of such agents, which could make their behavior harder to audit, constrain, or reset.
If applied to open-ended or user-facing environments, experience memories may also introduce risks related to privacy, data retention, and unintended reuse of past interactions.
Our work is evaluated only on public reasoning benchmarks and does not deploy such systems in real-world autonomous settings.

\section{LLM Usage}
\label{app:llm_usage}

LLMs are central components of our method and experiments.
All evaluated methods use LLMs as backbone reasoners, and \textsc{HippoSpark} further uses LLMs for bottleneck assessment, experience-guided next-move generation, candidate-move verification, and experience construction from historical reasoning trajectories.
Details of the model versions, prompting templates, decoding settings, and implementation protocols are provided in Appendix~\ref{app:implementation} and Appendix~\ref{app:method_details}.

LLMs were also used to assist with paper writing, including language polishing, grammar correction, and formatting suggestions.
They were not used to formulate the core research idea, design the method, conduct experiments, generate experimental results, or make scientific conclusions.
All technical claims, experimental analyses, and reported results were checked and finalized by the authors.

\section{Compute Resources}
\label{app:compute_resources}

For open-source backbone models, including Qwen3-14B and Qwen3-32B, we conduct inference on a local server with four NVIDIA A800 80GB GPUs.
The models are deployed for batched inference, and all reported results are obtained without further training or fine-tuning.
For closed-source backbone models, we use the corresponding API interface and report the model name, decoding configuration, and evaluation protocol in Appendix~\ref{app:implementation}.
The additional overhead of \textsc{HippoSpark} mainly comes from repeated bottleneck assessment, experience retrieval, experience-guided next-move generation, and verification during solving.
We provide token-level inference analysis in Appendix~\ref{app:inference_token_analysis} and experience construction cost analysis in Appendix~\ref{app:experience_construction_analysis}.

\section{HippoSpark Implementation Details}
\label{app:method_details}


\subsection{Overall Pipeline}

Our framework consists of two tightly coupled stages: \emph{experience construction} and \emph{on-demand experience utilization}. The construction stage runs offline and distills reusable transition-level solving patterns from historical trajectories into a structured experience library. The utilization stage runs online and invokes this library only when the current solver encounters a genuine transition gap. The two stages are aligned by the same design principle: experiences are constructed around pivotal transitions and later used to guide the next move at similarly structured transitions.

In the offline stage, we start from a set of training problems and collect multiple trajectories for each instance. These trajectories may include both successful and unsuccessful attempts, since both provide useful supervision: successful traces reveal effective pivots and reusable action patterns, while failed traces expose common traps, invalid transitions, and misleading heuristics. Based on these trajectories, the system follows the construction pipeline detailed in Appendix~\ref{app:Experience_Construction}. It first analyzes pivotal transitions to identify the decisive bottleneck and the strategy that resolved it. It then recalls the supporting knowledge needed to make that strategy executable, and synthesizes the result into structured experience cards. Each experience card has a two-layer form: an upper decision layer that describes the relevant \emph{situation} and \emph{strategic goal}, and a lower support layer that stores the execution-critical knowledge attached to that decision. Finally, these experiences are inserted into a dual-layer graph memory with normalization, merging, and deduplication.

In the online stage, the model does not consult the full experience library by default. Instead, it follows the utilization loop detailed in Appendix~\ref{app:On-Demand Experience Utilization}. At each step, a solver first assesses the current transition and identifies whether the dominant gap is strategic, computational, mechanical, or already resolved. If the next move is already clear, the solver proceeds directly. If a strategic gap remains, the solver formulates a structured query that describes the current blockage in terms of its abstract \emph{situation} and \emph{goal} or \emph{quest}. The system then retrieves a small set of relevant experience nodes from the graph, synthesizes them into context-aware local guidance, and feeds the adapted result back into the move-generation prompt. The resulting move is then checked by a verifier before it is committed to update the reasoning state.


This organization gives the framework two advantages. First, the expensive process of extracting reusable knowledge is amortized offline. Second, online utilization remains sparse and targeted: the solver accesses memory only when the current transition truly requires external support, and the retrieved experience is used only to guide the immediate next move rather than to globally condition the entire reasoning trajectory. As a result, the system preserves the flexibility of the base model while still benefiting from structured prior experience when local progress becomes difficult.

\subsection{Bottleneck-focused Experience Construction}
\label{app:Experience_Construction}

The goal of experience construction is to transform raw trajectories into compact, reusable decision knowledge rather than storing full solutions verbatim. In our framework, this process is organized as a staged pipeline. We first collect diverse trajectories for each training problem. We then analyze the pivotal transitions that determine success or failure, recall the supporting knowledge needed to make those transitions executable, and synthesize them into structured experience cards. Finally, the resulting cards are merged into a graph-structured memory with deduplication. At a high level, the construction process can be summarized as follows:

\begin{algorithm}[H]
\caption{Bottleneck-focused Experience Construction}
\label{alg:exp_construction}
\KwIn{training problems $\mathcal{D}$; sampled trajectories $\mathcal{T}$}
\KwOut{dual-layer experience graph $G$}

Initialize graph $G$\;

\ForEach{problem $p \in \mathcal{D}$}{
    Collect trajectories $\mathcal{T}_p$\;
    Analyze pivotal transitions $\{\alpha_i\}_{i=1}^{M} \leftarrow L_{\mathrm{analyze}}(\mathcal{T}_p, G_p)$\;
    Extract supporting topics $\{H_i\}_{i=1}^{M}$ from the transition analyses\;
    Obtain supporting knowledge $K_i \leftarrow \{L_{\mathrm{support}}(h): h \in H_i\}$\;
    Synthesize experience cards $\{e_i\}_{i=1}^{M} \leftarrow L_{\mathrm{card}}(\{\alpha_i\}_{i=1}^{M}, \{K_i\}_{i=1}^{M})$\;
    Insert cards $e_i=((c_i,g_i),d_i,K_i)$ and linked support nodes into $G$\;
    Merge near-duplicate experience nodes using situation--goal similarity and kernel similarity\;
}

\Return{$G$}
\end{algorithm}

\subsubsection{Trajectory Generation}

For each training problem, we sample multiple solution attempts with the base model. We keep both successful and unsuccessful trajectories, because they provide complementary supervision. Successful trajectories expose effective pivots and reusable solution patterns, while unsuccessful ones reveal common traps, invalid transitions, and misleading heuristics. The trajectory format is benchmark-dependent: for MATH and GPQA, it is primarily a reasoning trace; for BigCodeBench, it additionally includes implementation attempts and verification or execution feedback. In this stage, our goal is not to preserve the full trajectory as memory, but to prepare diverse raw material for later bottleneck-centered analysis.

\subsubsection{Analyze Pivotal Transitions}

This stage identifies the key turning points in a trajectory. Rather than summarizing the whole solution into a short sequence of steps, we isolate the pivotal transitions at which the correct strategy becomes identifiable. We also extract the supporting knowledge attached to these pivots, such as theorems, scientific principles, discriminative criteria, implementation methods, or debugging patterns. 

\paragraph{Prompt for pivot analysis.}
We use the prompt template shown in Figure~\ref{fig:app_pivot_analysis_prompt} to analyze each trajectory and mark its pivotal transitions.

\paragraph{Input and output.}
The input to this prompt is a single trajectory together with the original problem text, optional auxiliary context, and the trajectory outcome. The output has two parts: an \texttt{ANCHOR\_BLOCK} that lists the key domain anchors, and an \texttt{analysis} block that marks each decisive turning point with \texttt{[PIVOT]} and explains the barrier, the resolving anchor, and the role of the pivot.

\begin{figure}[htbp]
    \centering
    \includegraphics[width=0.92\linewidth]{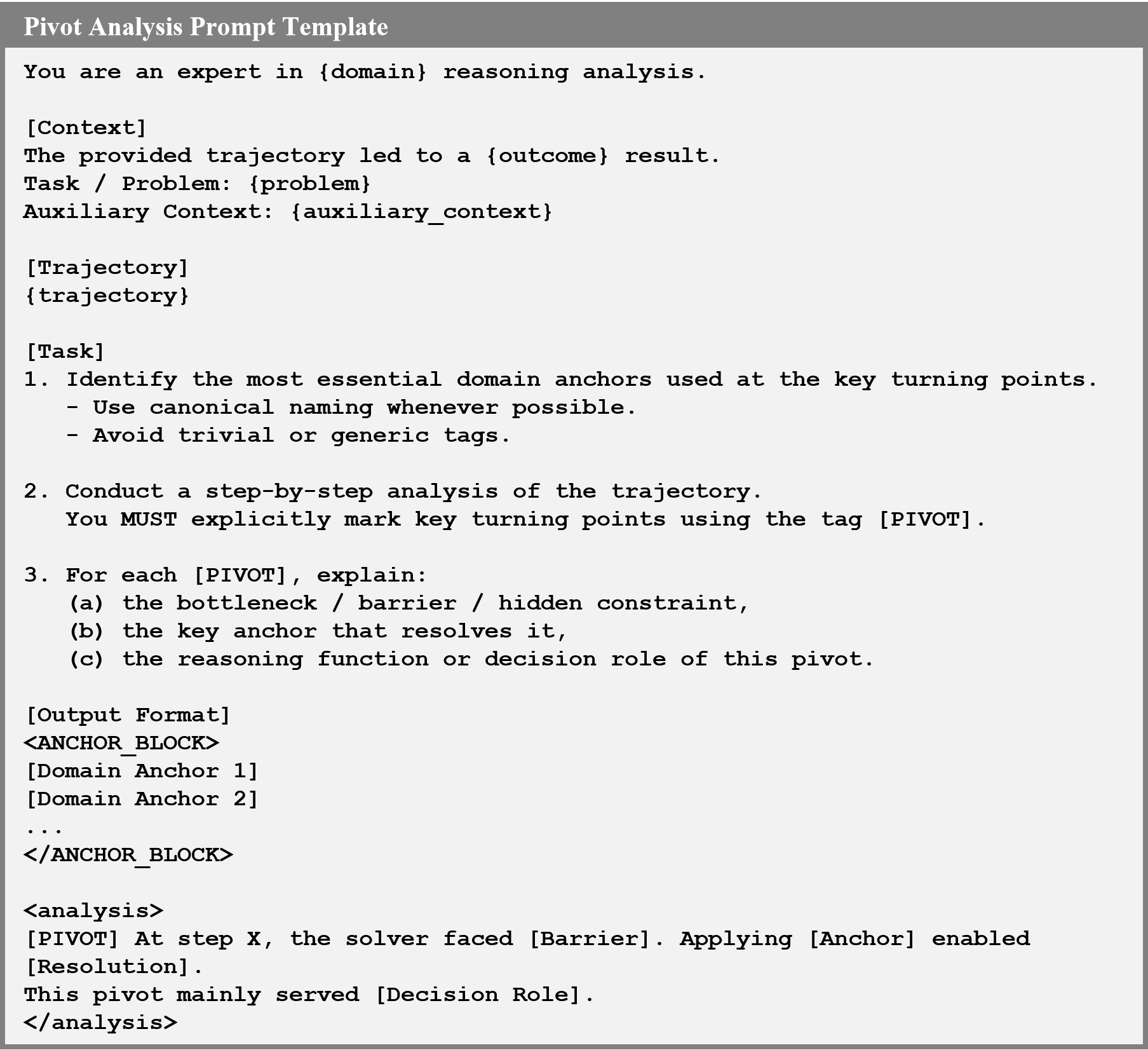}
    \caption{
    Pivot analysis prompt template. The prompt identifies domain anchors and marks decisive trajectory transitions with \texttt{[PIVOT]}.
    }
    \label{fig:app_pivot_analysis_prompt}
\end{figure}

\paragraph{Benchmark-specific specialization.}
All three benchmarks use the same analysis backbone, but the notion of an anchor is specialized by domain.

For MATH, the anchors are theoretical tools, lemmas, structural observations, or algorithmic principles. The prompt is tuned to emphasize mathematically meaningful barriers and the formal tool that resolves them.

For GPQA, the anchors are scientific principles, mechanisms, or discriminative criteria. The prompt further emphasizes the \emph{decision role} of each pivot, for example whether it identifies the governing principle, eliminates distractors, or corrects a misconception.

For BigCodeBench, the anchors are implementation methods, hidden constraints, API usage patterns, failure causes, or debugging cues. The prompt is grounded more directly in execution feedback and incorrect behavior, so that only genuinely decisive implementation signals are preserved as anchors.


\subsubsection{Recall Supporting Knowledge}

After the pivotal transitions have been identified, we explicitly recall the knowledge that should accompany them. The purpose of this stage is to convert ``known but not activated'' knowledge into callable support. This is important because a pivot alone is often not enough: the model may know which strategy is needed, but still fail to execute it without the relevant formula, criterion, check, or implementation pattern. In our framework, this recalled content forms the lower support layer attached to an experience card. The prompt template is shown in Figure~\ref{fig:app_supporting_knowledge_prompt}.

\paragraph{Input and output.}
The input to this prompt is the raw anchor set extracted from one or more pivotal transitions. The output is a cleaned and normalized set of reusable support objects. It includes a clustering map, a filtering log, and a final set of structured knowledge cards. In our implementation, these cards become the lower-layer support nodes later linked to experience cards in the graph.

\begin{figure}[htbp]
    \centering
    \includegraphics[width=0.92\linewidth]{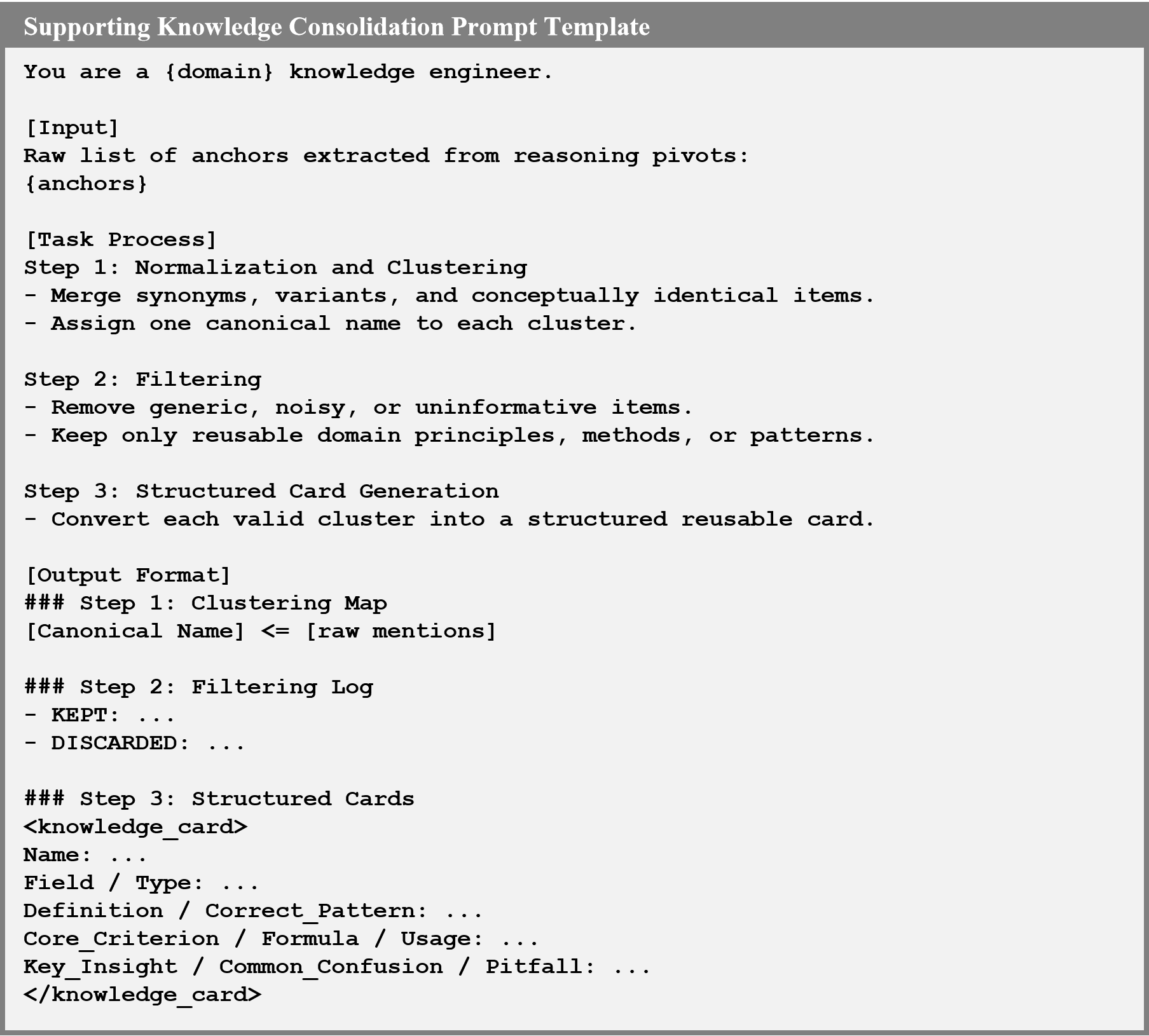}
    \caption{
    Supporting knowledge consolidation prompt template. The prompt normalizes raw anchors, filters noisy items, and converts valid clusters into structured support cards.
    }
    \label{fig:app_supporting_knowledge_prompt}
\end{figure}

\paragraph{Benchmark-specific specialization.}

For MATH, recall mainly returns formulas, derivation templates, and theorem-level reminders that can be directly used after a strategic pivot is recognized.

For GPQA, recall returns scientific principles, mechanism summaries, and discriminative criteria that help the solver compare options under the correct conceptual frame.

For BigCodeBench, recall returns implementation templates, API usage patterns, failure signatures, and lightweight debugging checks. This makes the recalled support directly executable during later code synthesis or repair.


\subsubsection{Synthesize Experience Cards}

Once we obtain both pivot analysis and supporting knowledge, we synthesize them into reusable experience cards. Each card is designed to capture not the full historical solution, but a reusable state-transition pattern: under what situation the card should be triggered, what the true strategic goal is, what action kernel should be applied, and how the supporting knowledge should be adapted. This directly follows the representation used in the method section, where experience separates a decision layer from an execution-support layer.

\paragraph{Prompt for experience-card synthesis.}
We use the prompt template shown in Figure~\ref{fig:app_experience_card_synthesis_prompt}.

\paragraph{Input and output.}
The input to this prompt consists of two parts: the expert analysis of the trajectory and the recalled knowledge cards or method cards. The output is a set of reusable experience units. Each unit specifies a \texttt{situation\_signature}, a \texttt{strategic\_goal}, an execution or action kernel, and explicit links to the cited support cards. These fields become the upper decision layer and the lower support references of the final dual-layer memory.

\begin{figure}[htbp]
    \centering
    \includegraphics[width=0.92\linewidth]{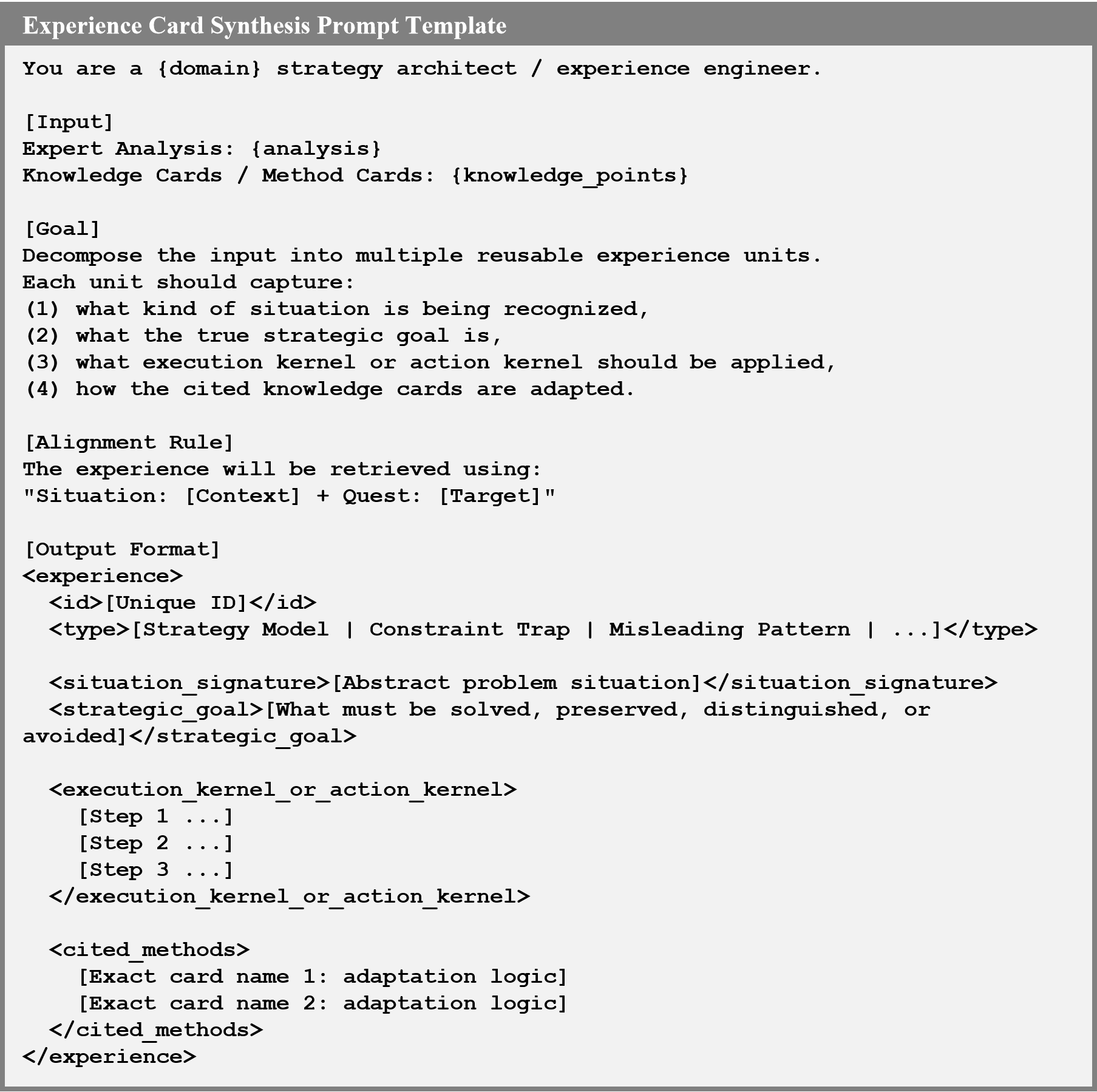}
    \caption{
    Experience card synthesis prompt template. The prompt converts pivot analysis and supporting knowledge into reusable experience units with a decision layer and linked execution-support layer.
    }
    \label{fig:app_experience_card_synthesis_prompt}
\end{figure}

\paragraph{Benchmark-specific specialization.}
For MATH, the synthesized experience card is a strategy card for an abstract mathematical situation. It emphasizes when a theorem, transformation, or structural method should be activated, and how it supports the next reduction step.

For GPQA, the synthesized experience card is closer to a scientific discrimination template. It focuses on identifying the governing principle, extracting the decisive criterion, and using that criterion to eliminate superficially plausible but mechanistically incorrect options.

For BigCodeBench, the synthesized experience card has a clearer two-layer form. The upper layer selects an implementation strategy based on task signals, constraints, and failure symptoms, while the lower layer cites method cards that provide concrete coding patterns or API usage details.


\subsubsection{Graph-based Integration and Deduplication}

All synthesized experiences are inserted into a graph-structured memory. In our implementation, this memory is explicitly organized as a two-layer graph. The first layer contains experience nodes, which store the reusable decision pattern itself, including the situation signature, the strategic goal, and the execution kernel. The second layer contains supporting knowledge nodes, such as math methods, scientific principles, or coding methods. Each problem first creates a problem hub node. The hub is connected to the experience nodes extracted from that problem, and each experience node is then connected to the lower-layer support nodes that it cites. In this way, the graph keeps both the decision pattern and the support needed to execute it.

The support-layer node type is benchmark-dependent. In MATH, the lower-layer nodes are \texttt{MathMethod} nodes. In GPQA, they are \texttt{ScientificPrinciple} nodes. In BigCodeBench, they are \texttt{CodingMethod} nodes. These support nodes are first normalized by canonical name, and repeated occurrences increment their count instead of creating a new node. This keeps the lower layer relatively stable and reusable across many problems.

Experience-node deduplication is more selective. For each newly synthesized experience, we build two embeddings. The first is a trigger embedding derived from the retrieval-facing fields, namely the \texttt{Situation} and \texttt{Quest/Goal}. The second is a content embedding derived from the execution kernel or action kernel. We then compare the new experience only against previously stored experience nodes of the same general family, rather than against all nodes in the graph. The implementation first retrieves up to the top three existing experience candidates by trigger similarity, and then checks whether both the trigger similarity and the kernel similarity exceed fixed thresholds.

For MATH and GPQA, an experience is merged into an existing node only when the trigger similarity is above \texttt{0.95} and the execution-kernel similarity is above \texttt{0.80}. If both conditions hold, the system reuses the existing node and increments its count. Otherwise, it creates a fresh experience node. For BigCodeBench, the same two-stage rule is used, but with a slightly looser trigger threshold and a slightly stricter kernel threshold: the default implementation uses \texttt{0.92} for trigger merging and \texttt{0.84} for kernel-level deduplication. This reflects the fact that coding tasks often share a broader implementation context while still requiring tighter agreement on the concrete action pattern.

This two-stage criterion is important. Using only trigger similarity would over-merge experiences that arise in similar situations but require different concrete actions. Using only kernel similarity would merge experiences that happen to share similar low-level procedures but are activated under different strategic conditions. Requiring both lets the graph preserve reusable abstractions without collapsing distinct decision patterns.

The edge structure also preserves how support is used. Problem-to-experience edges record that an experience was extracted from a given problem. Experience-to-support edges record citation and adaptation logic, that is, how a lower-layer method or principle is used to instantiate the upper-layer strategy. As a result, the stored memory is not just a bag of independent cards. It is a linked dual-layer structure that supports both retrieval and later interpretation.

\subsection{On-Demand Experience Utilization}
\label{app:On-Demand Experience Utilization}

At test time, our framework uses experience as a local transition aid rather than as a static prompt prefix. The solver does not consult memory at every step. Instead, it runs an iterative loop that first assesses the current condition and the next sub-goal, identifies whether the step can proceed directly or requires experience support, generates the next move, and verifies the proposed move before committing it. This appendix describes that loop in more detail than the main paper. The main text presents the core mechanism and its formal interface, while here we make the prompt-level controls and benchmark-specific routing details explicit.

\begin{algorithm}[H]
\caption{On-Demand Experience Utilization}
\label{alg:utilization}
\KwIn{Input problem $x$, final goal $G$, experience bank $\mathcal{E}$}
\KwOut{Reasoning trajectory $\tau$}

Initialize $s_0 \leftarrow \mathrm{Init}(x)$ and $\tau \leftarrow (s_0)$\;

\For{$t=0,1,\ldots$ until $G$ is reached}{
    \Repeat{$v_t=\texttt{accept}$}{
        Assess current condition and next sub-goal:
        $(c_t,g_t)\leftarrow L_{\mathrm{assess}}(s_t,G)$\;

        Identify the gap and route:
        $\Delta_t\leftarrow L_{\mathrm{gap}}(c_t,g_t)$,
        $r_t\leftarrow L_{\mathrm{route}}(\Delta_t)$\;

        \eIf{$r_t=\texttt{direct}$}{
            Generate a candidate move:
            $\tilde{m}_t\leftarrow L_{\mathrm{dir}}(c_t,g_t)$\;
        }{
            Retrieve transition-level experience:
            $E_t\leftarrow \operatorname{TopK}\big(\operatorname{Retrieve}((c_t,g_t),\mathcal{E})\big)$\;

            Generate a candidate move:
            $\tilde{m}_t\leftarrow L_{\mathrm{exp}}(c_t,g_t,E_t)$\;
        }

        Verify the candidate move:
        $v_t\leftarrow L_{\mathrm{verify}}(c_t,g_t,\tilde{m}_t)$\;
    }

    Commit $\tilde{m}_t$:
    $s_{t+1}\leftarrow \mathrm{Apply}(s_t,\tilde{m}_t)$\;

    Append $(\tilde{m}_t,s_{t+1})$ to $\tau$\;
}

\Return $\tau$\;
\end{algorithm}

\subsubsection{Integration into the Reasoning Loop}

The full online loop implements the transition-centric process defined in Section~3.1. At each turn, the system first summarizes the current condition and the next sub-goal:
\[
    (c_t,g_t)=L_{\mathrm{assess}}(s_t,G),
\]
where \(c_t\) describes what is known or has been derived so far, and \(g_t\) specifies the next sub-goal toward the final goal. The system then determines whether there is a clear operation for moving from \(c_t\) to \(g_t\):
\[
    \Delta_t=L_{\mathrm{gap}}(c_t,g_t), \qquad
    r_t=L_{\mathrm{route}}(\Delta_t)\in\{\texttt{direct},\texttt{experience}\}.
\]
If \(r_t=\texttt{direct}\), the candidate move is generated directly from the assessed condition and sub-goal:
\[
    \tilde{m}_t=L_{\mathrm{dir}}(c_t,g_t).
\]
If \(r_t=\texttt{experience}\), the assessed situation--goal pair is used as the retrieval key:
\[
    E_t=\operatorname{TopK}\big(\operatorname{Retrieve}((c_t,g_t),\mathcal{E})\big),
    \qquad
    \tilde{m}_t=L_{\mathrm{exp}}(c_t,g_t,E_t).
\]
Only the generated candidate move \(\tilde{m}_t\), not the raw retrieved experience \(E_t\), is considered for commitment. The verifier checks whether the move is grounded in the current condition and productive for the next sub-goal:
\[
    v_t=L_{\mathrm{verify}}(c_t,g_t,\tilde{m}_t),
    \qquad
    v_t\in\{\texttt{accept},\texttt{reject}\}.
\]
If \(v_t=\texttt{accept}\), the move is committed and the state is updated by
\[
    s_{t+1}=\mathrm{Apply}(s_t,\tilde{m}_t).
\]
If the move is rejected, the verifier returns feedback and the system re-enters transition assessment to generate a new candidate move.

This formulation is shared across benchmarks, while the concrete realization of the direct route is domain-specific. In AIME, direct moves often correspond to routine simplification, substitution, direct formula application, or numerical computation. In GPQA, direct moves correspond to local scientific comparisons, eliminations, one-step derivations, or numerical computation when needed. In BigCodeBench, direct moves are usually coding plans or localized implementation steps. Strategic gaps that require a theorem, transformation, domain criterion, API pattern, or repair strategy are routed to experience retrieval.

\subsubsection{Transition Assessment and Identify the Gap}

This stage corresponds to the first decision made in the online loop: what is the current transition problem, and does it already admit a direct next move? In practice, we implement transition assessment and gap identification in a single solver prompt. The solver first summarizes the current frame, then identifies the immediate subgoal, then decides what kind of gap remains. This combined design matches the way our code uses the solver output: the system does not separately classify the state and then open a second prompt to detect the gap. Both decisions are produced together and immediately routed to an action.

\paragraph{Input and output.}
The input is the current problem together with the latest verified state, tool output, execution feedback, or verifier response. The output contains a short diagnosis of the current state, the target next state, and the transition gap, followed by exactly one routed action. We use three abstract action types at the paper level: \texttt{<memory>} for strategic gaps, \texttt{<move>} for executable transitions, and \texttt{<answer>} when the final result is ready. Benchmark-specific operations such as calculation, option comparison, coding plans, or debugging steps are treated as concrete implementations of \texttt{<move>}.

\paragraph{Prompt template.}
The solver prompt template is shown in Figure~\ref{fig:app_transition_assessment_prompt}.

\begin{figure}[htbp]
    \centering
    \includegraphics[width=0.9\linewidth]{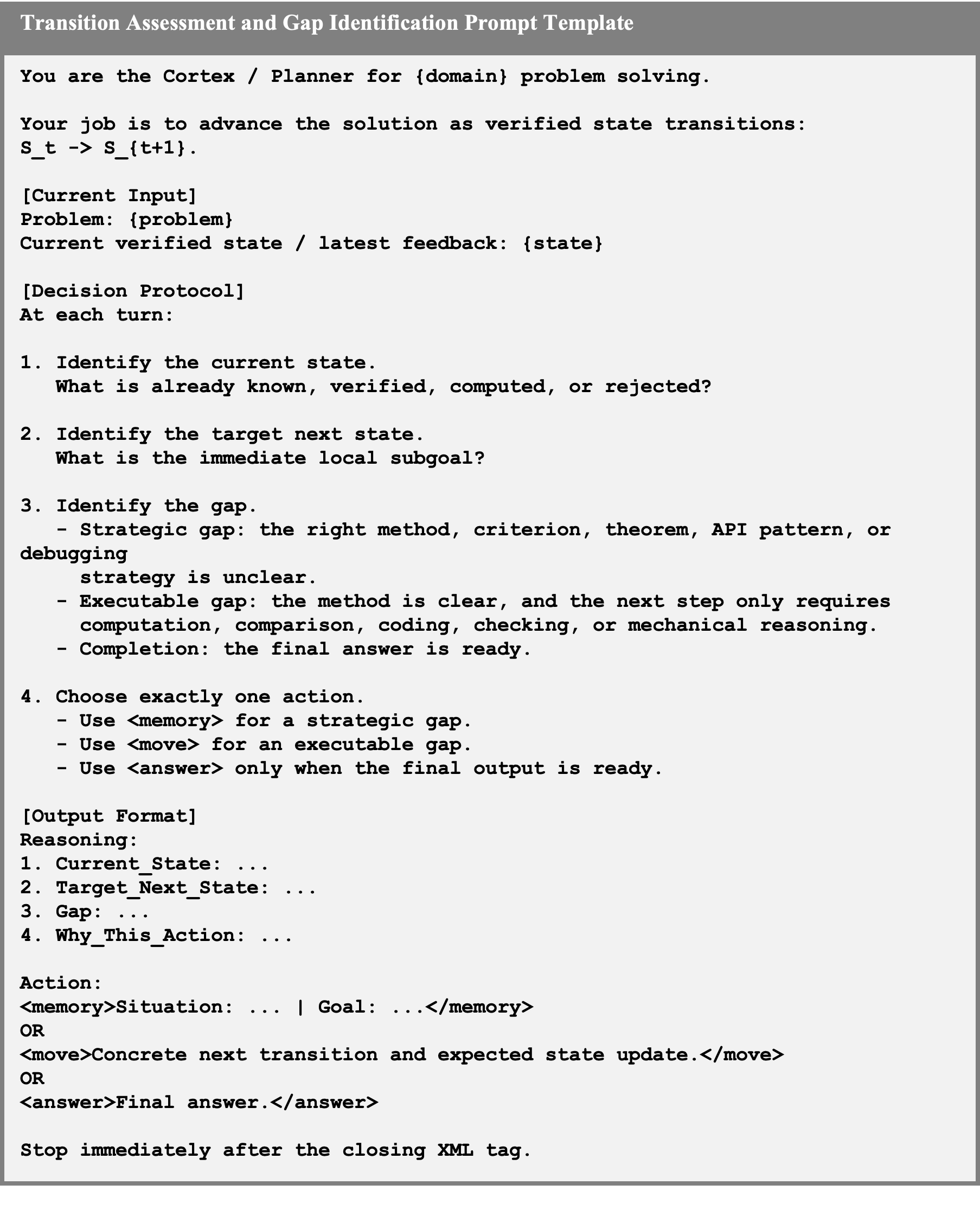}
    \caption{
    Transition assessment and gap identification prompt template. The solver diagnoses the current reasoning state, identifies the immediate gap, and routes the system to exactly one next action.
    }
    \label{fig:app_transition_assessment_prompt}
\end{figure}

\paragraph{Benchmark-specific specialization.}
The solver follows the same abstract routing rule across benchmarks: strategic gaps trigger experience retrieval, executable gaps trigger a concrete move, and completion triggers the final answer. The benchmark-specific prompts only specialize what counts as an executable move and what information the move should contain.

For MATH, executable moves are usually numerical or symbolic calculations. The implementation may therefore realize \texttt{<move>} as a calculation request, with the goal, known parameters, formula, and execution strategy specified explicitly. Strategic gaps, such as choosing the right theorem, transformation, or structural decomposition, still trigger \texttt{<memory>}. This makes the solver's decision closely tied to mathematical state decomposition: what is known, what target state is needed next, and whether the bridge is computational or strategic.

For GPQA, executable moves correspond to local scientific reasoning steps, such as comparison, elimination, one-step derivation, or checking a discriminator between plausible options. When the missing bridge is instead a scientific principle, domain criterion, or methodological pattern, the solver retrieves experience through \texttt{<memory>}. In our implementation, the GPQA solver is run with bounded control parameters, including up to \(16\) turns, at most \(2\) memory calls, and at most \(3\) calculator calls, reflecting that GPQA usually requires a small number of decisive interventions rather than long iterative search.

For BigCodeBench, executable moves are implementation plans. When the solver already knows the relevant API pattern, edge-case handling, and repair strategy, the \texttt{<move>} is instantiated as a coding plan for the implementation expert. When the solver is uncertain about APIs, hidden constraints, edge cases, or a specific failure mode, it retrieves experience first. In our implementation, this solver is typically run with a limit of \(12\) turns, so repeated failed planning does not expand indefinitely.

Thus, the solver prompt jointly performs state assessment and gap classification across all benchmarks, while the concrete realization of \texttt{<move>} is domain-specific: calculation for MATH, local option discrimination for GPQA, and coding plans for BigCodeBench.


\subsubsection{Move Generation}

Once the current transition has been assessed, the system generates the next move. If no strategic gap remains, the next move is generated directly from the current state. If a strategic gap remains, the system first constructs \(\mathrm{exp}_t\) through structured query formulation, retrieval, and synthesis, and then feeds the resulting guidance back into the same move-generation function. This is the sense in which the main-text expression \(f(s_t,\mathrm{exp}_t)\) should be interpreted: the prompt that generates the move is the same solver-level generator, but its effective context changes depending on whether experience support is empty or has been constructed by the memory pipeline.

\paragraph{Input and output.}
At the move-generation stage, the direct input is the current state plus, when needed, the adapted experience support for the current blockage. The output is one local next-step artifact: a mathematical move, a scientific comparison or elimination step, a coding plan, or the final answer when the task is complete. When retrieval is triggered, there are also intermediate outputs: a structured query, retrieved experience snippets, and a synthesized suggestion that is fed back into the move generator.

\paragraph{Structured query formulation prompt.}
When retrieval is needed, we first formulate a structured query using the prompt shown in Figure~\ref{fig:app_memory_query_prompt}.

\begin{figure}[htbp]
    \centering
    \includegraphics[width=0.92\linewidth]{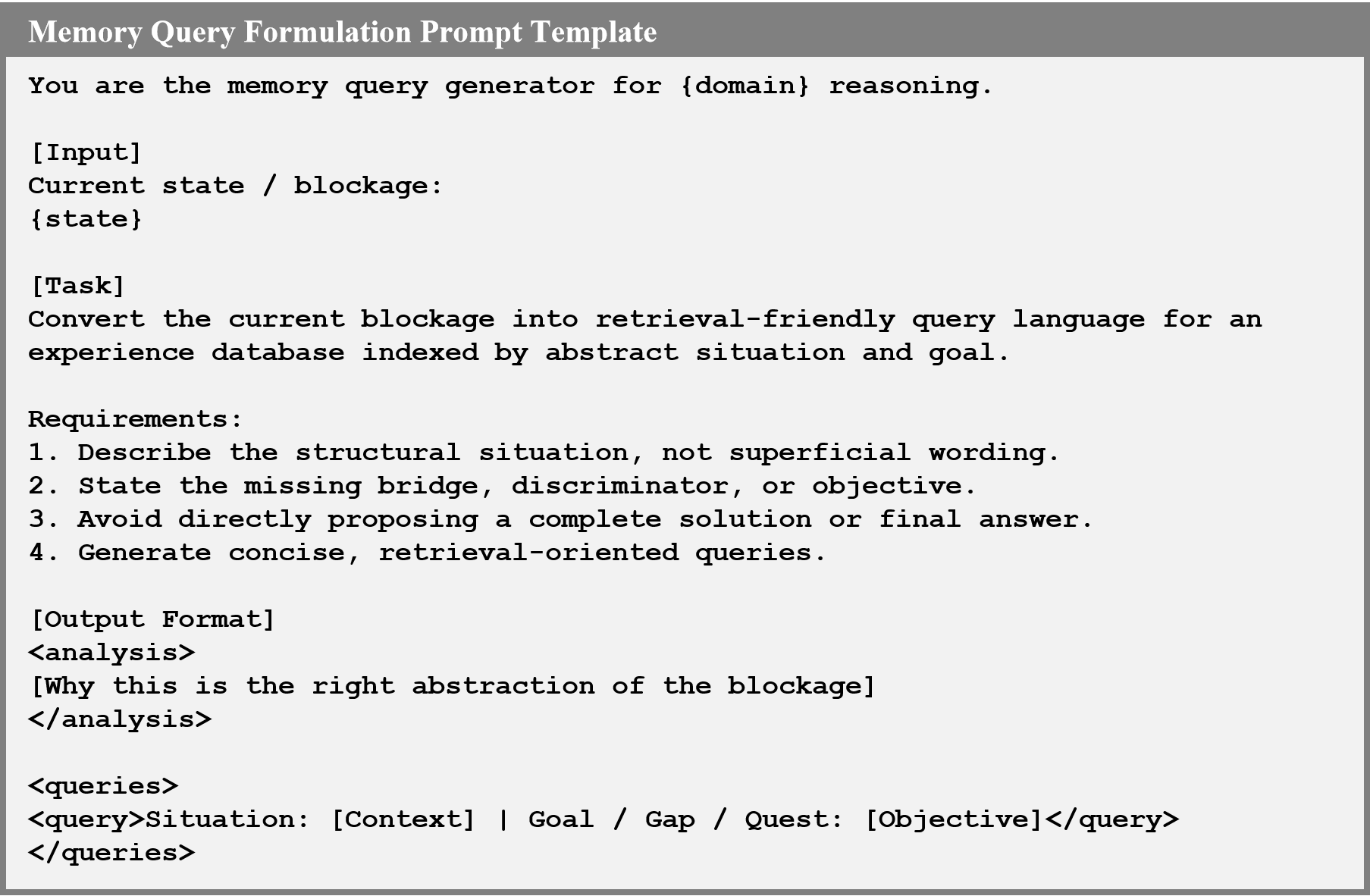}
    \caption{
    Memory query formulation prompt template. The prompt converts the current blockage into retrieval-oriented situation-and-goal queries for the experience memory.
    }
    \label{fig:app_memory_query_prompt}
\end{figure}

\paragraph{Memory-grounded next-step synthesis prompt.}
After retrieval, we synthesize the retrieved experience into local guidance using the prompt template shown in Figure~\ref{fig:app_memory_grounded_synthesis_prompt}.

\begin{figure}[htbp]
    \centering
    \includegraphics[width=0.92\linewidth]{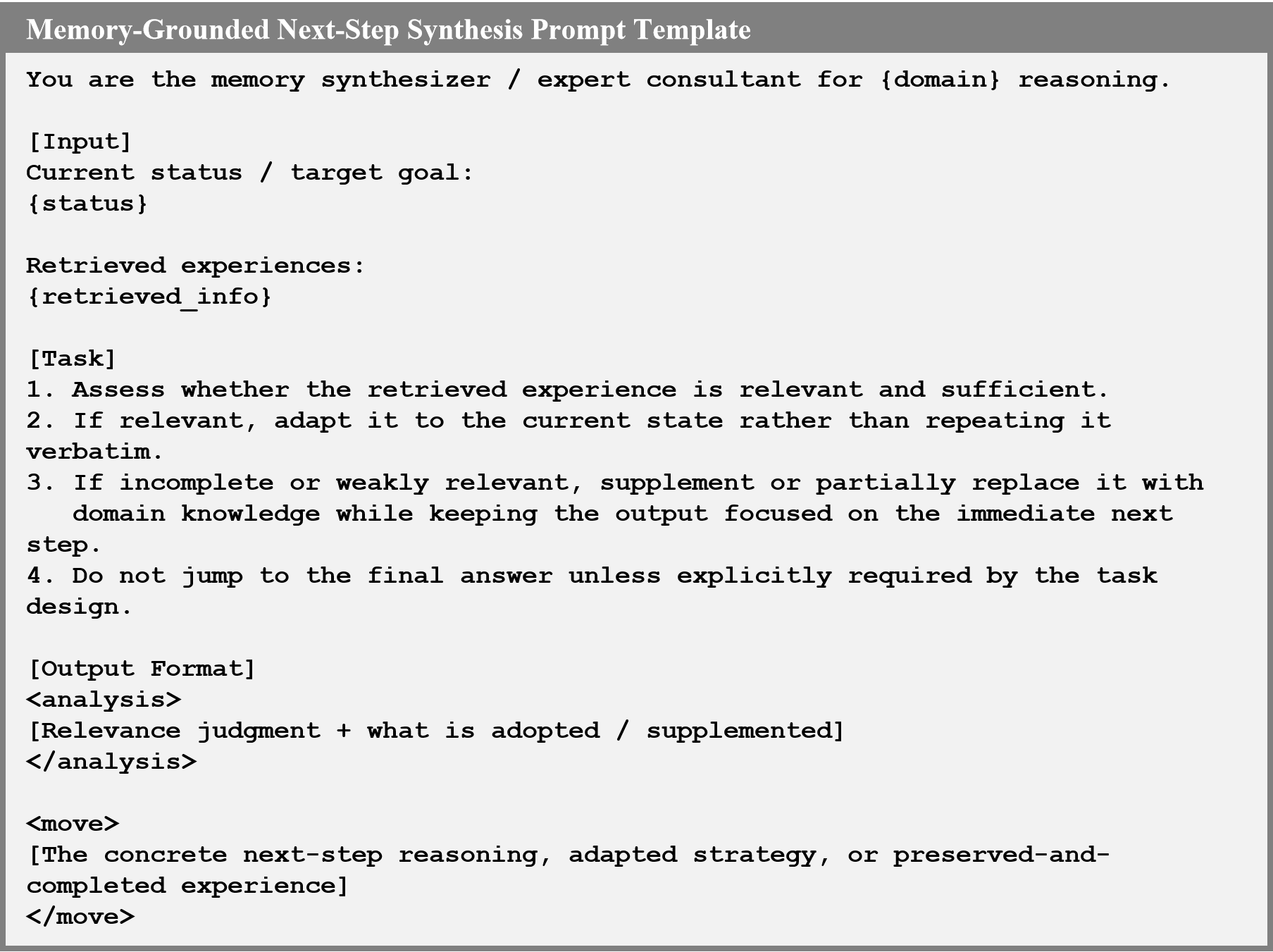}
    \caption{
    Memory-grounded next-step synthesis prompt template. The prompt judges the relevance of retrieved experience, adapts or supplements it, and produces focused guidance for the immediate next step.
    }
    \label{fig:app_memory_grounded_synthesis_prompt}
\end{figure}

\paragraph{Retrieval and integration details.}
In our implementation, retrieval is intentionally constrained by two mechanisms. First, we retrieve only the top-\(1\) core experience for each query, so the solver receives a single targeted memory item rather than a large set of loosely related examples. Second, we apply a similarity threshold to filter weak matches. The default threshold is \(0.7\) for most settings, including MATH and BigCodeBench. For GPQA-Biology, however, the available experience pool is much smaller, so we lower the threshold to \(0.6\) to avoid suppressing retrieval entirely when only moderately similar Biology experience is available.

For MATH, the query is normalized toward a \texttt{Situation}--\texttt{Quest} form, and the system retrieves from both the original need and the normalized query before synthesis. For GPQA, retrieval also uses top-\(1\) selection, but the solver is encouraged to issue the raw scientific need directly rather than over-rewriting it. For BigCodeBench, retrieval uses the same top-\(1\) principle with the default threshold of \(0.7\), since the goal is to recover one strong implementation strategy rather than aggregate many loosely related experiences.

After retrieval, the system does not paste the raw memory text into the main reasoning chain. Instead, it passes the retrieved snippets to the synthesis prompt in Figure~\ref{fig:app_memory_grounded_synthesis_prompt}. This prompt first judges whether the retrieved experience is relevant and sufficient, then adapts or supplements it into a local \texttt{<move>}. This is the key integration step that turns retrieved experience into \(\mathrm{exp}_t\) in \(f(s_t,\mathrm{exp}_t)\): the move generator continues from the current state plus this adapted suggestion, rather than from unprocessed retrieved text.

\paragraph{Benchmark-specific specialization.}
For MATH, the synthesized \texttt{<move>} usually preserves the retrieved strategy and supplements missing formulas, constraints, or intermediate mathematical steps. The goal is not to answer from memory, but to bridge the current verified state to the next valid state.

For GPQA, the synthesized \texttt{<move>} becomes a discriminative scientific next step, such as a concrete comparison, elimination, or criterion check. The synthesizer should not jump directly to the final option unless the task is already resolved; it should instead produce the local reasoning needed to advance the state.

For BigCodeBench, the synthesized \texttt{<move>} becomes localized implementation guidance. Retrieved programming experience is adapted to the task's concrete variable names, APIs, failure modes, and edge-case constraints. When the coding path is clear, this guidance can be instantiated as a coding plan for the implementation expert.


\subsubsection{Verification}

Before a proposed move is committed, the framework applies verification. The purpose of this stage is to block invalid local transitions before they distort the downstream trajectory. Verification is therefore local and pre-commitment oriented. It does not re-solve the full task from scratch. Instead, it checks whether the proposed move is executable, aligned with the current goal, and safe under the current state.

\paragraph{Input and output.}
The input to the verifier is the current state or reasoning trace together with the proposed action, move, or artifact. The output is a small control decision: approval, rejection with a concrete fix, or a targeted revision request. In some implementations, the verifier can also recommend that the solver return to memory retrieval instead of continuing with the current move.

\paragraph{Prompt template.}
The verification prompt template is shown in Figure~\ref{fig:app_verification_prompt}.

\begin{figure}[htbp]
    \centering
    \includegraphics[width=0.92\linewidth]{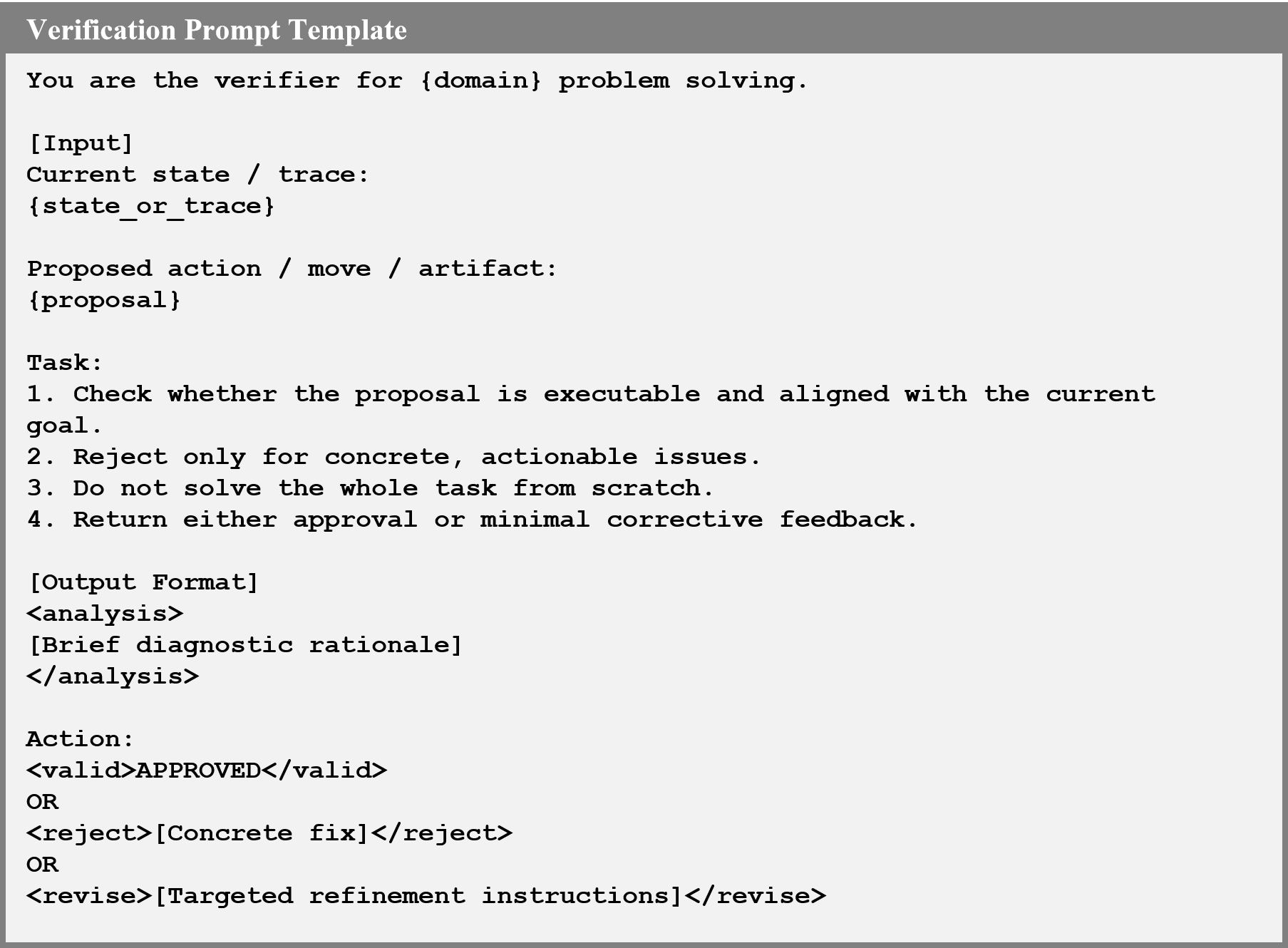}
    \caption{
    Verification prompt template. The verifier checks whether the proposed move is executable and goal-aligned, and returns approval or concise corrective feedback.
    }
    \label{fig:app_verification_prompt}
\end{figure}

\paragraph{Benchmark-specific specialization.}

For AIME, verification is specialized as state-transition checking.
The verifier checks whether the proposed action is coherent with the verified state, whether variables are defined, whether the action type matches the actual gap, and whether the operation is mathematically safe to execute.
This prevents the solver from forcing computation when a strategic retrieval step is still required.
In addition, AIME uses Python as a numerical calculation tool: when a move requires arithmetic computation, equation solving, or numerical checking, the solver generates a corresponding Python block within the \texttt{<move>} and uses its output as execution support for the next reasoning state.

For GPQA, verification is split into two levels.
The move verifier checks whether a proposed \texttt{<move>} is an actually executable scientific next step rather than a vague plan.
It requires concrete checks and at least one usable intermediate result.
When numerical calculation is needed, GPQA also allows the move to include a Python block as a calculation tool, so that intermediate quantities can be computed or checked before being incorporated into the reasoning trace.
A second verifier then judges whether the accumulated reasoning trace sufficiently supports the current answer attempt.
If not, it may request revision or explicitly recommend a new memory retrieval step.
In the code, this verifier stack is coupled with bounded revision control, including a configurable limit on verifier-driven refinements.

For BigCodeBench, verification is specialized into lightweight static code checking. The verifier checks the required interface, explicit constraints, missing imports, obvious logical slips, and edge-case omissions, but it does not execute the code. This static filter also changes the control flow: if static verification fails repeatedly, the system is routed back toward the memory route rather than being allowed to emit the same flawed executable move indefinitely. In implementation terms, BigCodeBench realizes the abstract \texttt{<move>} route as a coding plan for the implementation expert, and realizes the abstract \texttt{<memory>} route as a programming-experience query. In our default implementation, the memory fallback is triggered after \(3\) consecutive verifier rejections.

\section{Implementation Details}
\label{app:implementation}

\subsection{Benchmark Setup and Evaluation Protocol}

To comprehensively evaluate the proposed framework across heterogeneous reasoning scenarios, we conduct experiments on three representative benchmarks covering mathematical reasoning, scientific question answering, and code generation. For each benchmark, we explicitly separate the data used for experience construction from the data used for final evaluation, and report benchmark-specific metrics together with an overall score for cross-benchmark comparison.


\subsubsection{Benchmarks and Data Splits}
\label{app:bench_data}

We evaluate our framework on three benchmarks spanning \textbf{mathematical reasoning}, \textbf{scientific question answering}, and \textbf{code generation}: \textbf{AIME\citep{di_zhang_2025, huggingfaceh4_aime_2024, huggingfaceh4_aime_2025}}, \textbf{GPQA-Diamond\citep{rein2024gpqa}}, and \textbf{BigCodeBench\citep{zhuo2025bigcodebench}}.

\noindent\textbf{Experience Construction and Evaluation Split}
\label{app:experience_split}

We separate experience construction from final evaluation to avoid order-dependent evaluation.
Many experience-based methods~\citep{suzgun2026dynamic,ouyang2025reasoningbank} use online accumulation, where the $n$-th test problem can use experience collected from the previous $n-1$ problems.
This makes the result sensitive to the order of test examples.
Instead, we first construct an experience pool, freeze it, and then evaluate on a disjoint set of problems.

For AIME, historical problems provide a natural source of matched experience, so we construct experience from earlier AIME problems and evaluate on AIME 2024 and AIME 2025.
For GPQA-Diamond and BigCodeBench-Hard, no comparable historical pool with the same format and difficulty is available.
We therefore randomly split each benchmark into disjoint construction and evaluation subsets.
No evaluation example is used during experience construction.
This protocol preserves the benchmark distribution while ensuring that experience use is evaluated on unseen problems.

\textbf{AIME.}[License/Terms: contest materials; used for research evaluation and not redistributed]
AIME (American Invitational Mathematics Examination) is a widely used benchmark for evaluating advanced mathematical reasoning abilities of large language models. The problems typically require multi-step symbolic manipulation, numerical computation, and precise logical deduction, with each question having a unique integer answer between 0 and 999. In our experiments, we randomly sample 100 problems from \textbf{AIME 1983-2023}\citep{di_zhang_2025} to construct the experience memory, and use \textbf{AIME 2024}\citep{huggingfaceh4_aime_2024} and AIME 2025\citep{huggingfaceh4_aime_2025} as test sets to evaluate the effectiveness of our framework. This setting naturally separates memory construction from final evaluation and helps reduce potential overlap between historical experience data and test problems.

\textbf{GPQA-Diamond.}\citep{rein2024gpqa}[License: MIT]
GPQA is a challenging benchmark designed to assess expert-level reasoning in the domains of physics, chemistry, and biology. We adopt GPQA-Diamond, a high-quality subset consisting of 198 multiple-choice questions. Following our experimental design, we randomly select approximately 30\% of the dataset (57 out of 198 questions) to construct experience, with the selection performed proportionally across the three domains to preserve the original distribution of physics, chemistry, and biology in GPQA-Diamond. The remaining 141 questions are used for evaluation, forming a held-out split between experience construction and testing that enables us to study whether the proposed framework can transfer useful experience within scientific reasoning tasks, consistent with standard held-out evaluation practices in language model benchmarking~\citep{liang2023holistic}. This benchmark allows us to study whether the proposed framework can transfer useful experience within scientific reasoning tasks.

\textbf{BigCodeBench.}\citep{zhuo2025bigcodebench}[License: Apache 2.0]
BigCodeBench is an execution-based code generation benchmark for realistic programming tasks, evaluating models based on functional correctness rather than surface-level text matching. It provides two prompt settings, \textbf{Complete} and \textbf{Instruct}. The Complete setting focuses on code completion from relatively detailed docstrings, whereas the Instruct setting requires models to generate code directly from natural language instructions and thus demands stronger task understanding and reasoning ability. Since our work focuses on leveraging historical experience under natural language task descriptions, we adopt the \textbf{Instruct} setting of \textbf{BigCodeBench-Hard}. To better highlight the impact of the proposed experience mechanism, we use this more challenging subset, which contains 148 tasks in total. Among them, approximately 30\% of the tasks (40 out of 148) are used for experience construction, while the remaining 70\% (108 tasks) are reserved for evaluation, forming a held-out evaluation setup between experience memory construction and downstream code generation testing, consistent with standard language model benchmarking protocols~\citep{liang2023holistic}.

\subsubsection{Evaluation Metrics}




\paragraph{Metric definitions.}
For AIME, $\hat{y}_i^{(r)} = y_i$ means the predicted answer is mathematically equivalent to the gold answer. Our AIME answer extraction and equivalence checking follow the evaluation style of MATH-500, whose reference implementation is released with the MATH benchmark evaluation code~\citep{hendrycks2021measuring}. For GPQA, $\hat{y}_i^{(r)} = y_i$ means the selected option matches the gold choice. The per-run pass@1 is
\begin{equation}
\mathrm{pass@1}^{(r)} = \frac{1}{N}\sum_{i=1}^{N}\mathbb{I}(\hat{y}_i^{(r)} = y_i),
\end{equation}
where $\mathbb{I}(\cdot)$ is the indicator function and $r$ indexes the run. For AIME, $\hat{y}_i^{(r)} = y_i$ means the predicted answer is mathematically equivalent to the gold answer; for GPQA, it means the selected option matches the gold choice. If a benchmark is run for $R$ independent runs, its avg pass@1 is defined as
\begin{equation}
\mathrm{avg\ pass@1} = \frac{1}{R}\sum_{r=1}^{R}\mathrm{pass@1}^{(r)}.
\end{equation}
For AIME, we additionally report best-of-5:
\begin{equation}
\mathrm{best\text{-}of\text{-}5} = \frac{1}{N}\sum_{i=1}^{N}\mathbb{I}\Big(\max_{1 \leq r \leq 5}\mathbb{I}(\hat{y}_i^{(r)} = y_i)=1\Big).
\end{equation}

For BigCodeBench, let $c_{i,j}^{(r)}$ denote the $j$-th generated candidate for task $i$ in run $r$, and let $\mathrm{Pass}(c_{i,j}^{(r)}) \in \{0,1\}$ indicate whether that candidate passes the benchmark test suite. The per-run pass@1 is
\begin{equation}
\mathrm{pass@1}^{(r)} = \frac{1}{N}\sum_{i=1}^{N}\mathrm{Pass}(c_{i,1}^{(r)}),
\end{equation}
and BigCodeBench avg pass@1 is
\begin{equation}
\mathrm{avg\ pass@1} = \frac{1}{R}\sum_{r=1}^{R}\mathrm{pass@1}^{(r)}.
\end{equation}
Its pass@3 is defined as
\begin{equation}
\mathrm{pass@3} = \frac{1}{N}\sum_{i=1}^{N}\mathbb{I}\Big(\max_{1 \leq j \leq 3}\mathrm{Pass}(c_{i,j}) = 1\Big),
\end{equation}
where $c_{i,j}$ denotes the $j$-th candidate in a three-sample evaluation for task $i$.

\paragraph{Overall score.}
Let $S_{\text{math}}$ denote AIME avg pass@1, $S_{\text{gpqa}}$ denote GPQA avg pass@1, and $S_{\text{bcb}}$ denote BigCodeBench avg pass@1. We define the overall score as the unweighted mean of these three benchmark-level scores:
\begin{equation}
S_{\text{overall}} = \frac{1}{3}\left(S_{\text{math}} + S_{\text{gpqa}} + S_{\text{bcb}}\right).
\end{equation}
All reported scores are ratios in $[0,1]$.

\subsection{Baseline Implementation Details}

We compare our method with several representative baselines. These baselines cover direct prompting, explicit reasoning, self-refine, raw trajectory reuse, and memory-augmented experience reuse.

\begin{itemize}
    \item \textbf{Vanilla.}
    Vanilla is the most basic direct-generation setting. The model receives only the original task description. It does not use external experience, historical memory, or additional demonstrations. This baseline measures the base model's performance without experience augmentation.

    \item \textbf{CoT (Chain-of-Thought).}\citep{wei2022chain}
    CoT explicitly prompts the model to produce intermediate reasoning steps before giving the final answer. It uses only the current task context and does not store or retrieve experience across tasks. This baseline tests whether explicit reasoning alone can provide gains comparable to experience-based methods.

    \item \textbf{Self-Refine.}\citep{madaan2023selfrefine}
    Self-Refine improves an answer through a generate-feedback-revise loop. The feedback is produced for the current output and is used to revise the same task. It does not build a cross-task experience memory. This baseline represents single-task self-improvement at test time.

    \item \textbf{Trajectory Experience.}
    This baseline stores raw solution trajectories from historical tasks as experience units. It does not further abstract or structure the trajectories. At evaluation time, it retrieves a related historical task and prepends the full trajectory to the prompt. This baseline helps test whether gains come from seeing past trajectories directly or from extracting more compact reusable experience.

    \item \textbf{ReasoningBank.}\citep{ouyang2025reasoningbank}
    ReasoningBank extracts reusable reasoning memories from previous problem-solving trajectories. Instead of storing the raw trajectory, it summarizes useful reasoning patterns, strategies, or lessons into experience entries. During evaluation, it retrieves relevant experience and adds it to the prompt. This method is an important comparison because it also studies experience reuse, but its experience is still retrieved before solving starts.

    \item \textbf{Dynamic Cheatsheet (DC).}\citep{suzgun2026dynamic}
    Dynamic Cheatsheet builds and updates a cheatsheet as tasks are solved, and then uses the cheatsheet for later tasks. We evaluate two official variants:
    \begin{itemize}
        \item \textbf{DC-Cu (Cumulative):} updates the cheatsheet after each task and accumulates experience over time. For a new task, the current cheatsheet is used as static background knowledge.
        \item \textbf{DC-RS (Retrieval + Synthesis):} retrieves relevant historical task descriptions and solution traces before solving the current task. It then updates the cheatsheet based on the retrieved information and uses the updated cheatsheet in the prompt.
    \end{itemize}
    These two variants represent different ways to accumulate and reuse external memory.

\end{itemize}



\subsection{Experiment Setup Details}

We conduct experiments on three backbone models: 
\textbf{Qwen3-14B}~\citep{yang2025qwen3}, 
\textbf{Qwen3-32B}~\citep{yang2025qwen3}, and 
\textbf{GPT-5.4-mini}~\citep{openai_gpt54mini}. 
Both Qwen3 models are used under the Apache 2.0 License and are run in no-thinking mode. 
This setting helps isolate the effect of external experience construction and utilization, rather than relying on the models' built-in deliberation mode. 
GPT-5.4-mini is accessed through its official API under the corresponding provider terms of use.

To ensure a fair comparison, all methods use the same backbone and evaluation protocol within each benchmark. Unless otherwise specified, generation uses temperature = 0.7, top\_p = 0.8, and max\_tokens = 8k. Retrieval-based methods use top-1 retrieval. We only change these settings when a baseline requires a different procedure.

\begin{itemize}
    \item \textbf{Vanilla.}
    We directly prompt the model with the original question.

    \item \textbf{CoT.}
    We use the same prompt input as Vanilla, but ask the model to produce step-by-step reasoning. We increase max\_tokens to 16k to avoid truncating longer reasoning chains.

    \item \textbf{Self-Refine.}
    We use a three-stage process. The model first generates an initial answer, then produces feedback on that answer, and finally generates a revised answer conditioned on the original question and the feedback.

    \item \textbf{Trajectory Experiment.}
    We first generate solution trajectories on the experience split of each benchmark. Only correctly solved problems and their trajectories are kept in the trajectory memory. During evaluation, we retrieve the most semantically similar historical problem and prepend its trajectory to the current prompt.

    \item \textbf{ReasoningBank.}
    We first generate trajectories on the experience split. We then extract reusable experience entries from each problem-trajectory pair to build an experience memory. At evaluation time, the most relevant experience entry is retrieved and prepended to the prompt. Experience extraction uses max\_tokens = 2k.

    \item \textbf{Dynamic Cheatsheet.}
    We evaluate both DC-Cu and DC-RS. In both settings, the cheatsheet is updated only on the experience split and kept fixed during evaluation. DC-Cu accumulates the cheatsheet during experience-set problem solving and then directly uses it on the evaluation split. DC-RS additionally retrieves related examples from the experience split before updating the cheatsheet. Following the official implementation, both answer generation and cheatsheet generation use temperature = 0.


\end{itemize}

\section{Additional Results}
\label{app:additional_results}

\subsection{Analysis on GPQA-Biology: When Experience Coverage Becomes the Bottleneck}
\label{app:gpqa_biology_analysis}





We analyze the performance gap on the GPQA-Biology subset. In our setup, the Biology experience pool is significantly smaller than that of other domains, resulting in limited coverage of relevant reasoning patterns. Specifically, the GPQA experience split contains only 5 Biology problems, compared with 25 Physics and 27 Chemistry problems. After experience construction, this yields only 20 Biology experience items for Qwen3-32B, 28 for Qwen3-14B, and 20 for GPT-5.4-mini, while the corresponding numbers are 78/124/94 for Physics and 112/152/124 for Chemistry. Therefore, the Biology memory is much smaller than the other two domains from the start.

As a consequence, retrieval is less likely to return experience aligned with the current situation and local goal. We further verify this by analyzing the similarity between queries and retrieved experience. On Qwen3-32B, the average top-1 retrieval score on Biology is only 0.676, with a maximum of 0.760. This is lower than Physics (average 0.706, maximum 0.788) and clearly lower than Chemistry (average 0.751, maximum 0.879). A similar pattern also appears on the other backbones: the average top-1 Biology retrieval score is 0.683 on Qwen3-14B, and for GPT-5.4-mini the few Biology cases that trigger retrieval reach only 0.689 on average. These numbers indicate that even the highest-ranked retrieved item is often only weakly matched to the actual need in Biology.

Figure~\ref{fig:gpqa_biology_qpcr_case} provides a representative example of this coverage bottleneck. In this qPCR calibration problem, the model formulates a query that correctly targets the local reasoning need: interpreting what a slope of $-3.3$ means under ten-fold dilution. However, the retrieved experience remains only weakly matched and drifts toward a generic parameter-transfer pattern, rather than providing the domain-specific qPCR rule needed to answer the question. This illustrates that the main issue in Biology is not necessarily failure to trigger retrieval or formulate the local query, but the lack of sufficiently close experience in the memory pool.

\begin{figure}[htbp]
    \centering
    \includegraphics[width=0.92\linewidth, height=0.8\textheight, keepaspectratio]{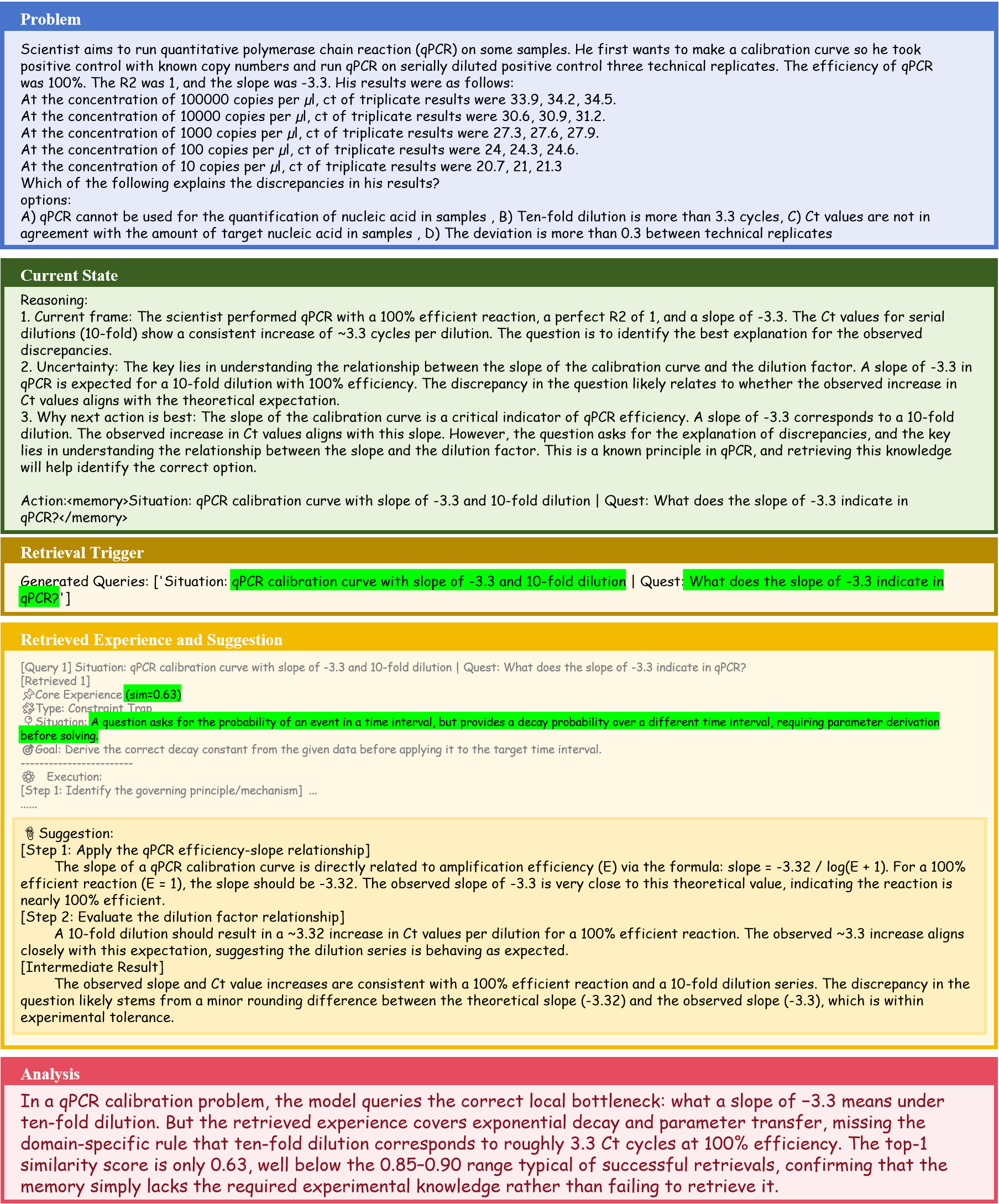}
    \caption{
    Representative GPQA-Biology case showing the coverage bottleneck. The query correctly identifies the local need, but the retrieved experience is only loosely related and fails to provide the required qPCR-specific knowledge.
    }
    \label{fig:gpqa_biology_qpcr_case}
\end{figure}
\FloatBarrier

This reduced coverage directly weakens the effectiveness of experience-based guidance, and explains why HippoSpark does not consistently outperform \textit{Ours (w/o Exp)} on this subset. These observations highlight a key limitation of our framework: when experience coverage is insufficient, retrieval quality degrades, and the benefit of experience correspondingly diminishes.

\subsection{Detailed Analysis of HippoSpark}
\label{app:detailed_analysis}

To complement the analysis in the main paper, we provide a more detailed examination of HippoSpark in this appendix. In addition to expanded quantitative results, we include supplementary observations and representative case analyses to better understand where the gains come from and how the experience mechanism functions during problem solving.

We analyze HippoSpark from two perspectives. From the utilization perspective, we study how experience is retrieved on demand during reasoning and integrated in a next-move-guiding manner. From the construction perspective, we examine how bottleneck-focused experience organization provides actionable support at critical reasoning states. 

Unless otherwise specified, all analyses are based on Qwen3-32B on AIME24 and AIME25, covering 300 problem-solving instances in total.

\subsubsection{Difficulty-triggered retrieval.}
\label{app:on_demand}

\textbf{Useful experience typically becomes identifiable only after reasoning reaches a concrete bottleneck.}

Prior experience-based methods often fail to provide stable improvements over Vanilla (Table~\ref{tab:main_results}), and in some settings even underperform it. \textbf{A key reason is that they retrieve experience before reasoning begins, when the actual bottleneck has not yet emerged.} In most cases, such retrieval is based on task-level semantic similarity. However, task-level similarity does not guarantee alignment with the actual solution path or the specific difficulty that later arises. As a result, the retrieved experience may be irrelevant to the current need and can even act as distracting context.


Figure~\ref{fig:difficulty_task_level_mismatch} illustrates this failure mode. The current problem and the retrieved problem are both about geometric configurations with equal lengths, so task-level retrieval considers them similar. However, the actual bottleneck is different: the current problem requires combinatorial matching under cyclic symmetry in a regular 24-gon, whereas the retrieved experience follows a triangle-similarity construction with parallel equal-length segments. The retrieved path is therefore semantically related but operationally misaligned, and it does not provide useful guidance for the needed next move.

\begin{figure}[htbp]
    \centering
    \includegraphics[width=0.92\linewidth]{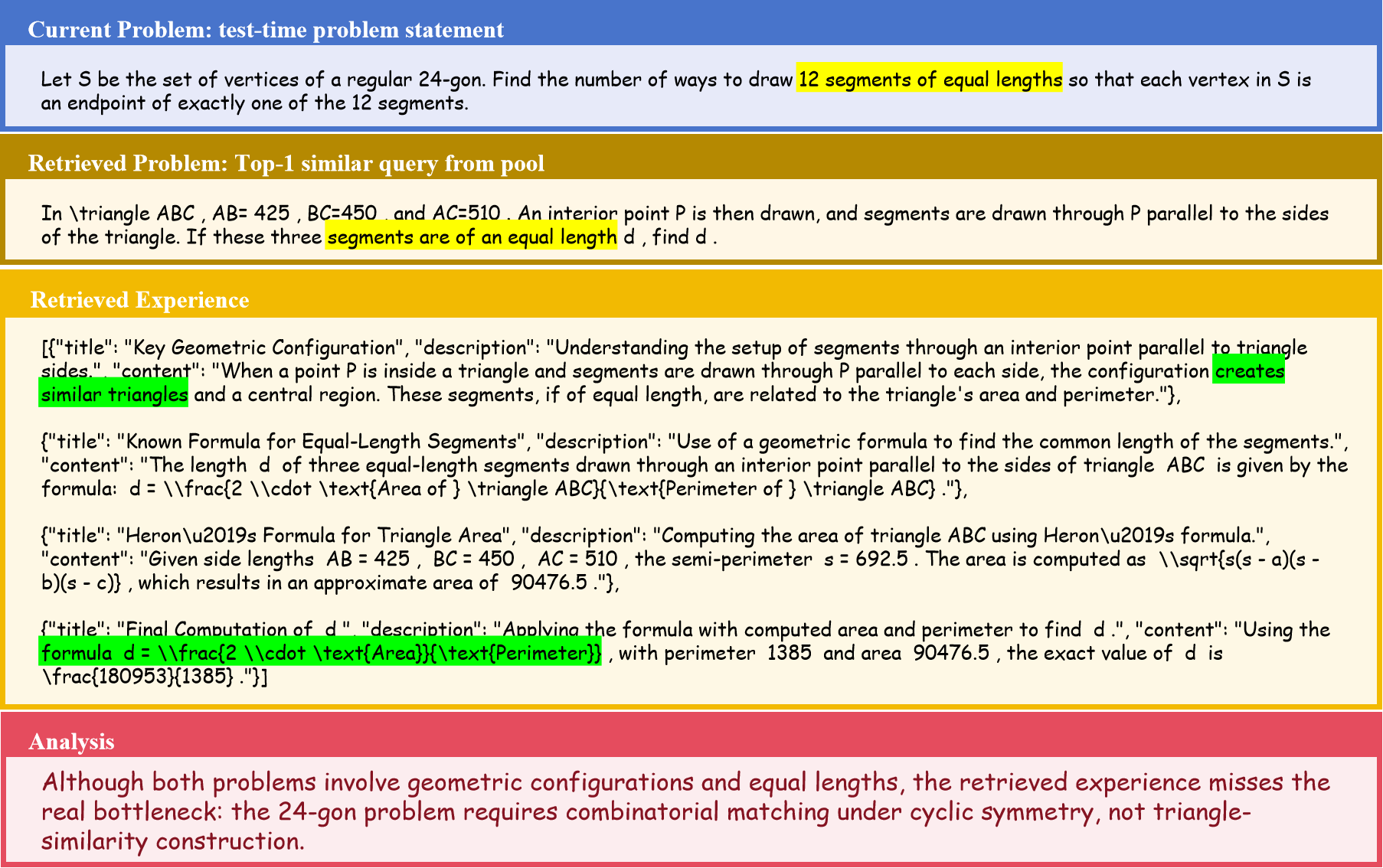}
    \caption{
    A task-level retrieval failure case. The retrieved problem is superficially similar because both involve geometry and equal lengths, but its solution path does not match the combinatorial bottleneck of the current regular-polygon matching problem.
    }
    \label{fig:difficulty_task_level_mismatch}
\end{figure}

\textbf{By contrast, HippoSpark achieves more reliable gains because retrieval is conditioned on the current reasoning state.} Specifically, retrieval is triggered from the current \textit{situation} and \textit{local goal}, and experience items are indexed in the same form. This allows the model to retrieve experience that is better aligned with the current bottleneck and therefore more useful for deciding the next move.


Figure~\ref{fig:difficulty_local_bottleneck_match} shows a representative successful retrieval. In a rotated-parabola problem, retrieval is triggered exactly when the model needs to derive the equation after a $60^\circ$ rotation. The retrieved experience is not merely about rotation in general; it identifies the precise local operation needed at this step: express the original coordinates in terms of the transformed coordinates and substitute them back into the original equation. This directly changes the next state from recognizing a hard transformation to executing the correct algebraic setup.

\begin{figure}[htbp]
    \centering
    \includegraphics[width=0.92\linewidth, height=0.8\textheight, keepaspectratio]{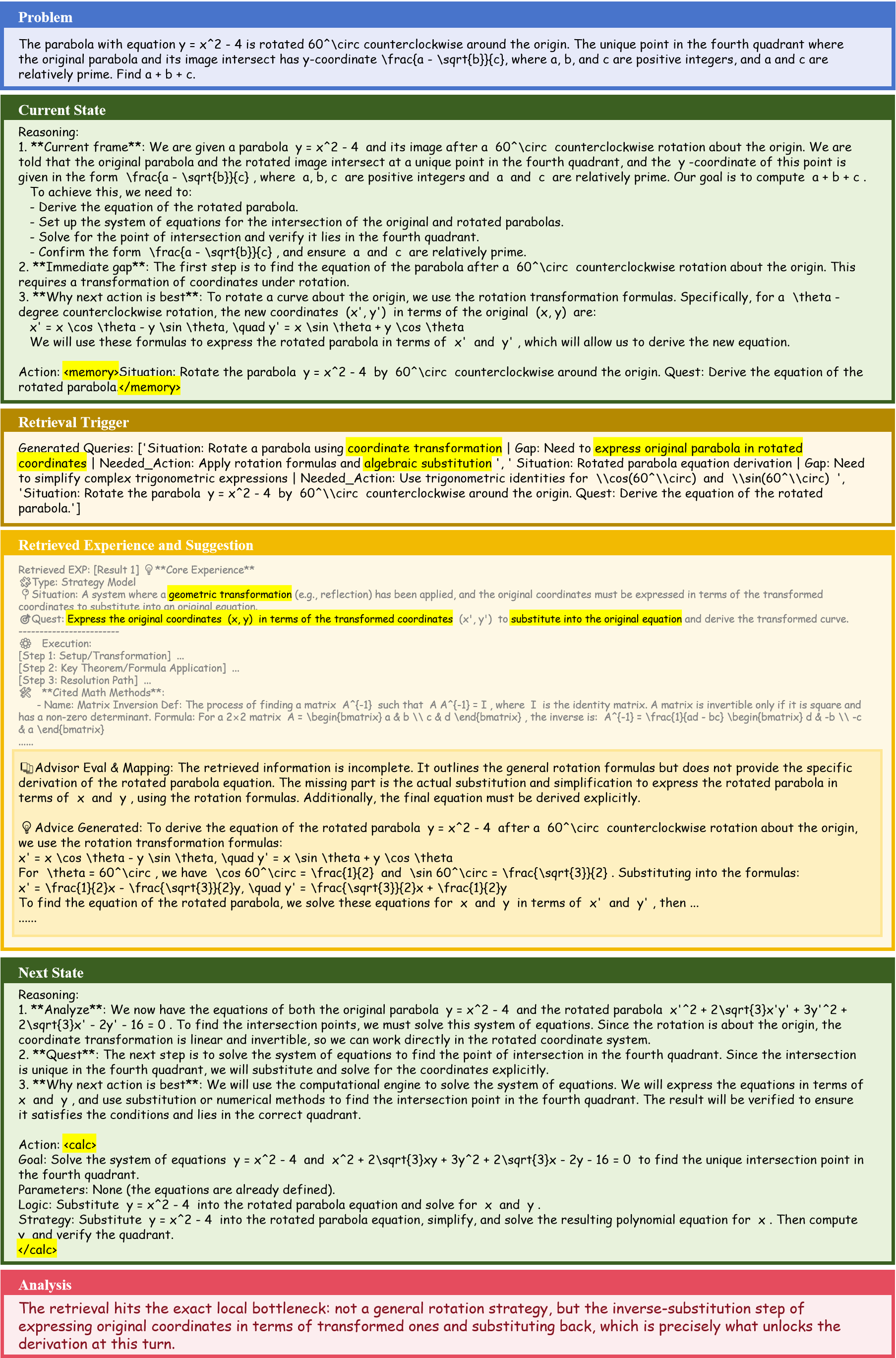}
    \caption{
    A successful difficulty-triggered retrieval case. The query is formed from the current reasoning state, and the retrieved experience matches the local bottleneck: inverse coordinate substitution after rotation.
    }
    \label{fig:difficulty_local_bottleneck_match}
\end{figure}

\textbf{We further verify that retrieval in HippoSpark indeed follows a difficulty-triggered pattern.} 
Specifically, we examine 300 problem-solving instances produced by Qwen3-32B on AIME24 and AIME25, and analyze both the distribution of retrieval frequency and the gains brought by experience under different numbers of retrieval calls. The detailed statistics are shown in Figure~\ref{fig:exp1}.

Among the 300 cases, 10.67\% involve no retrieval and already achieve 71.88\% accuracy under \textit{Ours (w/o EXP)}, suggesting that retrieval is often unnecessary for problems the model can already solve. For the remaining cases where retrieval is triggered, Figure~\ref{fig:exp1}(a) shows a clear long-tail distribution: most problems trigger only a few retrievals, while only a small fraction require many. Figure~\ref{fig:exp1}(b) further shows that as retrieval frequency increases, the accuracy of \textit{Ours (w/o EXP)} consistently decreases. \textbf{Together, these results suggest that retrieval is not invoked indiscriminately, but is more likely to be triggered on problems that are intrinsically harder for the model.}

\textbf{However, more retrieval does not necessarily lead to larger gains from experience.} As shown in Figure~\ref{fig:exp1}(c), HippoSpark yields clear improvements when only a few retrievals are needed, but the benefit becomes unstable when retrieval is triggered many times (e.g., more than five times). This is likely because such cases are not only harder, but may also involve accumulated errors in the model's interpretation of the current situation or local goal. Once this state estimation has drifted, subsequently retrieved experience may no longer align well with the actual bottleneck, and therefore becomes less effective at improving the final outcome.


Figure~\ref{fig:difficulty_state_drift_many_retrievals} provides a hard case illustrating this limitation. The problem again requires combinatorial matching on a regular 24-gon. Early retrieval correctly targets the segment-length and pairing bottleneck, but by a later turn the model has drifted into triangle-and-tangent geometry. Subsequent retrieval is then coherent with this wrong state, retrieving experience about tangents and chords rather than cyclic matching. This shows why repeated retrieval alone cannot guarantee recovery: once the current situation is mischaracterized, retrieval becomes aligned with the mistaken trajectory instead of the original problem.

\begin{figure}[htbp]
    \centering
    \includegraphics[width=0.92\linewidth]{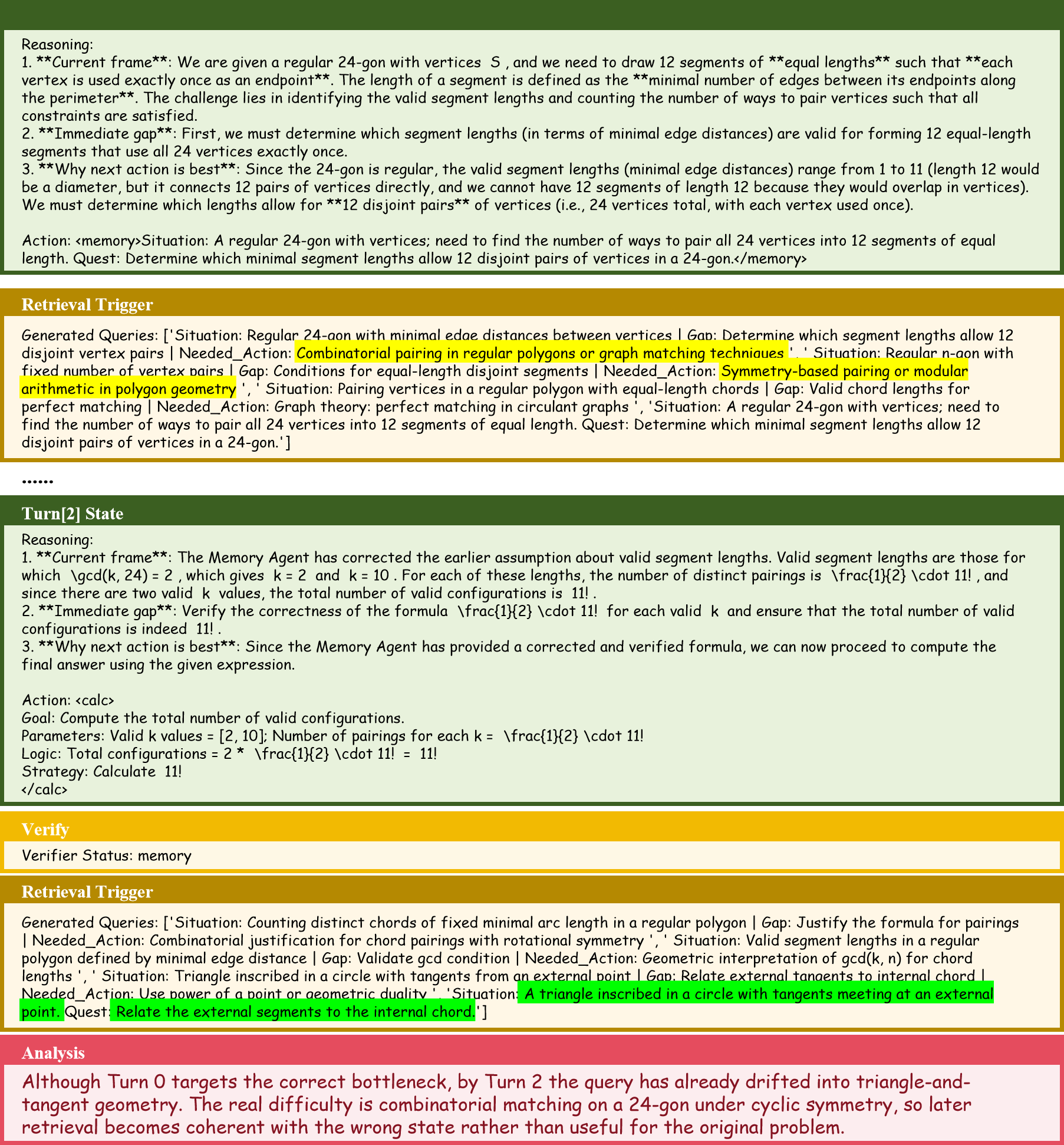}
    \caption{
    A hard case with repeated retrieval after state drift. The initial query targets the correct combinatorial bottleneck, but the later state shifts to irrelevant triangle-and-tangent geometry, so subsequent retrieval no longer supports the original problem.
    }
    \label{fig:difficulty_state_drift_many_retrievals}
\end{figure}
\FloatBarrier

\subsubsection{Next-move-guiding integration.}
\label{app:next_move_guiding}

\textbf{Experience should support the current next move, rather than remain in the main reasoning context after the bottleneck has been crossed.}

In HippoSpark, retrieved experience is used only to guide next-move generation. Once the current bottleneck is resolved, the experience itself is not retained in the main reasoning context. This design prevents previously retrieved experience from becoming irrelevant information that may interfere with subsequent reasoning.

\textbf{To verify this design, we compare HippoSpark with a global integration variant that directly inserts the retrieved experience text into the main problem-solving context.} The analysis results are shown in Figure~\ref{fig:exp2}, where \textit{Global} denotes directly appending retrieved experience to the main reasoning context, and \textit{Ours} denotes our next-move-guiding integration strategy. Figure~\ref{fig:exp2}(a) reports the performance changes brought by different experience integration methods relative to the same transition-centric framework without experience (\textit{Ours (w/o EXP)}). Figures~\ref{fig:exp2}(b) and \ref{fig:exp2}(c) further compare their effects on reasoning efficiency. Specifically, Figure~\ref{fig:exp2}(b) reports the number of reasoning steps, measured by the number of intermediate states in our state--move formulation. Figure~\ref{fig:exp2}(c) reports the additional reasoning burden introduced by different integration strategies, including both the number of experience calls and the total amount of experience tokens inserted into the main reasoning trajectory. The y-axis denotes the number of experience calls, and the bubble size represents the average number of tokens introduced into the main reasoning context per call.

\textbf{Directly inserting raw experience into the main reasoning context does not provide stable gains, but substantially increases reasoning cost.} As shown in Figure~\ref{fig:exp2}(a), the \textit{Global} variant fails to consistently improve over \textit{Ours (w/o EXP)}. At the same time, Figures~\ref{fig:exp2}(b) and \ref{fig:exp2}(c) show that it significantly increases both reasoning length and retrieval overhead: compared with our method, the number of reasoning steps and experience calls roughly doubles, while the total number of experience tokens introduced into the main reasoning trajectory increases by about an order of magnitude.

\textbf{These results suggest that unintegrated experience often does not effectively help the model advance reasoning, and instead becomes additional context that dilutes attention.} By contrast, when experience is used only to guide the current next move, its role is explicit and bounded: it helps the model cross the current bottleneck without polluting the subsequent reasoning process with no-longer-relevant information. This observation is also consistent with the design principle adopted in many agentic systems, where the main task-solving trajectory is maintained separately to avoid distraction from excessive auxiliary context.


Figures~\ref{fig:app_next_move_ours_case} and \ref{fig:app_next_move_global_case} show a representative comparison on the same nested absolute-value and trigonometric intersection problem. Under our next-move-guiding integration, retrieval is used to sharpen the current state into a concrete next operation: derive the breakpoints and piecewise behavior of the nested absolute-value functions. The retrieved experience is therefore converted into a bounded transition that advances the solver from broad problem understanding to an executable subproblem.

\begin{figure}[htbp]
    \centering
    \includegraphics[width=0.92\linewidth, height=0.8\textheight, keepaspectratio]{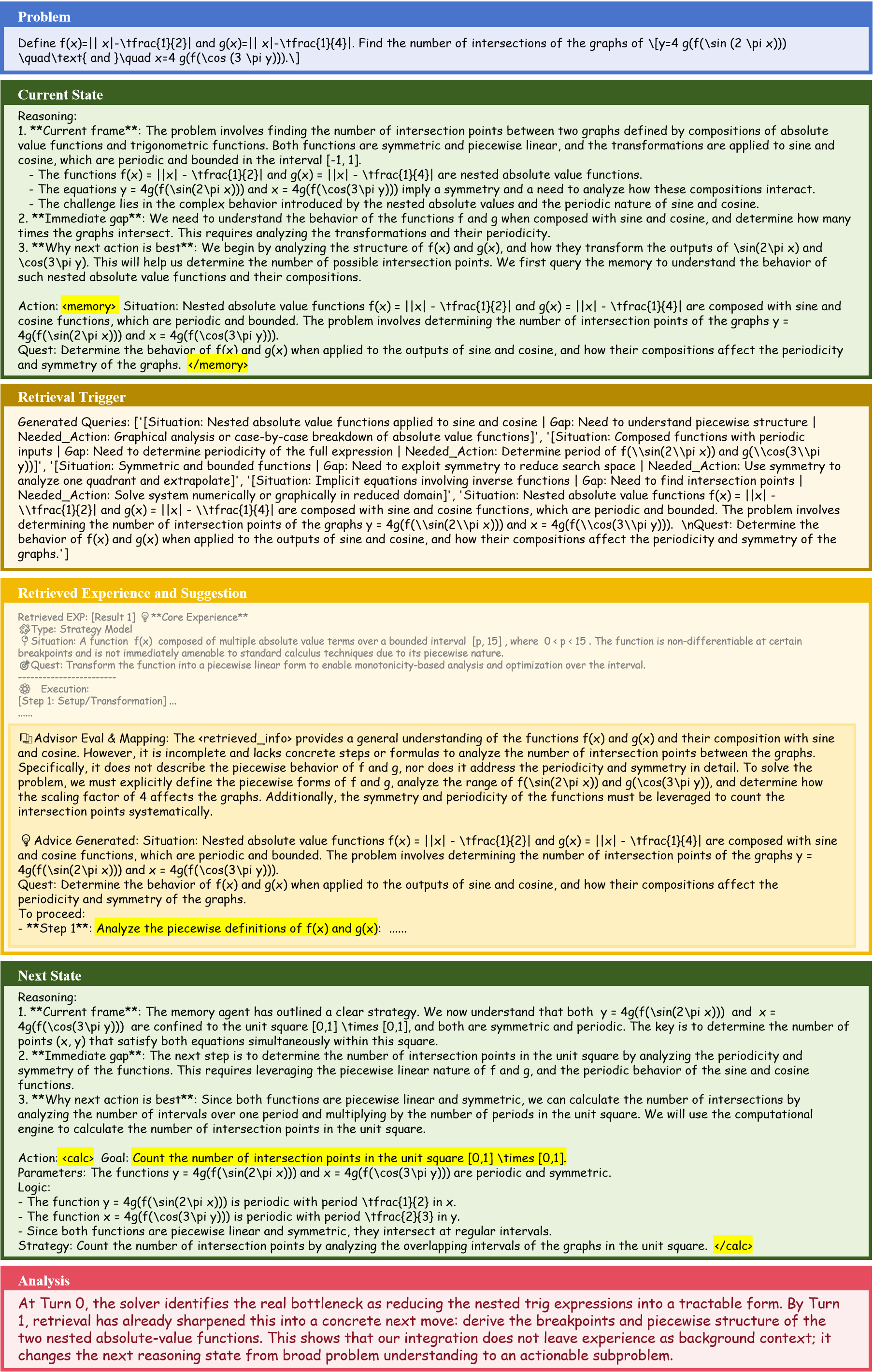}
    \caption{
    Representative case for next-move-guiding integration. The retrieved experience is integrated as local guidance for the current transition, turning a broad difficulty into the concrete next move of deriving breakpoints and piecewise structure.
    }
    \label{fig:app_next_move_ours_case}
\end{figure}

By contrast, when the retrieved experience is kept as raw context, it remains at the level of topic similarity: nested absolute values, symmetry, and periodicity are mentioned, but the reasoning state is not pushed toward a clearly executable next step. The model continues to circle around the same difficulty, showing how unbounded experience integration can add relevant-looking context without actually changing the next move.

\begin{figure}[htbp]
    \centering
    \includegraphics[width=0.92\linewidth, height=0.8\textheight, keepaspectratio]{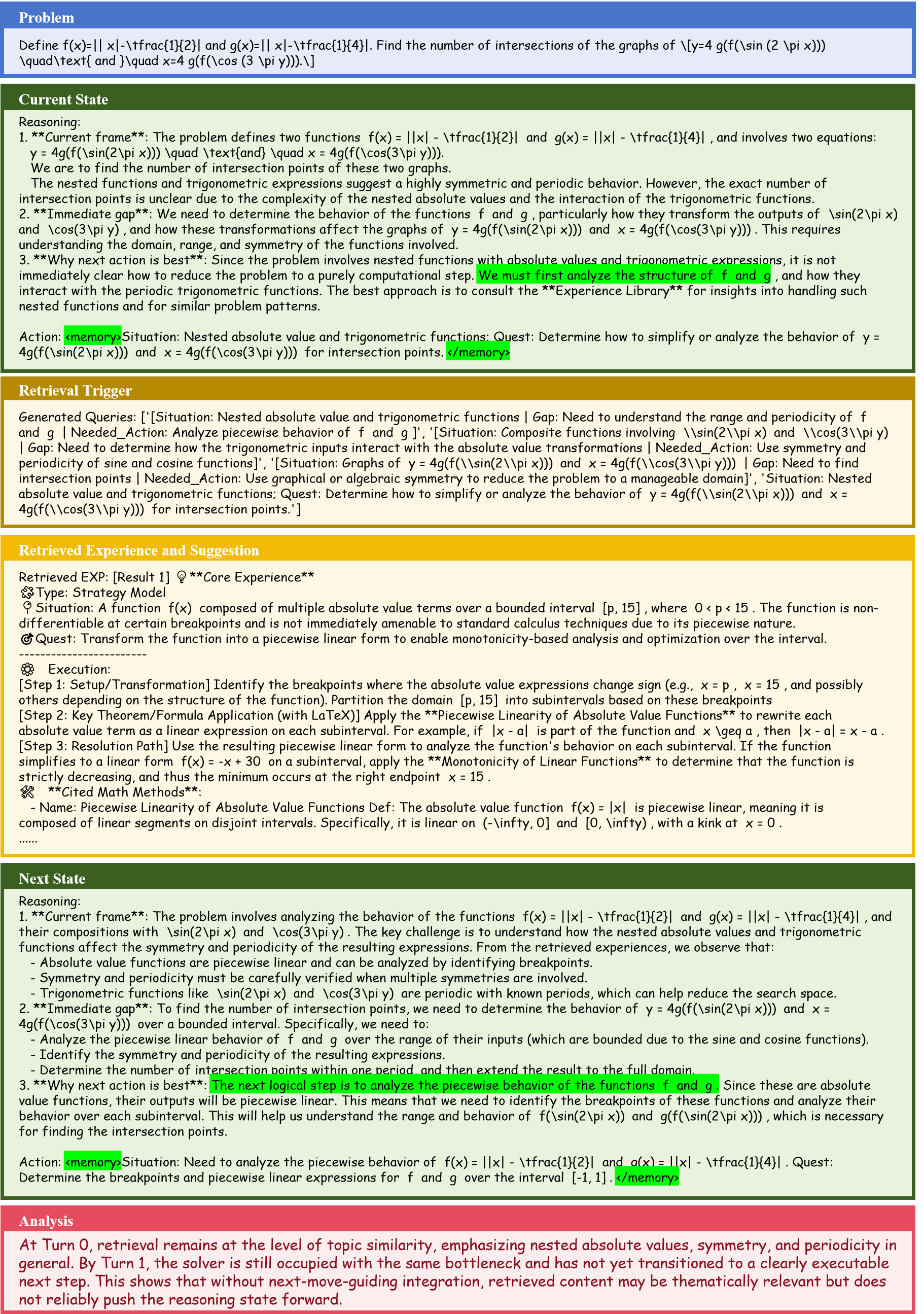}
    \caption{
    Representative case for global experience integration. The retrieved content is thematically relevant, but remains as broad background context and does not reliably move the solver to the next executable subproblem.
    }
    \label{fig:app_next_move_global_case}
\end{figure}
\FloatBarrier

\subsubsection{Bottleneck-focused construction.}
\label{app:bottleneck-focused}

\textbf{For experience to be reusable across tasks, it should be constructed around bottlenecks rather than full trajectories.}

Experience organized around decisive bottlenecks is more likely to transfer across problems, because bottlenecks often recur at the level of local reasoning difficulty even when the full task trajectories differ. By contrast, storing complete trajectories tends to mix useful support with task-specific context, which reduces reusability and introduces more irrelevant information.

For HippoSpark, this principle is also naturally aligned with our utilization framework. Since retrieval is triggered when the model encounters a concrete difficulty, the experience memory must be organized around such difficulties in order to provide effective support. We examine the constructed experience on Qwen3-32B across different benchmarks and find that it is indeed centered on bottlenecks. Specifically, from \(100\) AIME problems we construct \(406\) experience items, from \(57\) GPQA-Diamond problems we construct \(210\) items, and from \(40\) BigCodeBench problems we construct \(118\) items. These experience items are extracted around pivotal difficulties rather than around full task trajectories.


Figure~\ref{fig:app_bottleneck_rotated_parabola_case} gives a representative math example. The original problem involves a rotated parabola and asks for an arithmetic expression derived from its intersection point. The constructed experience does not summarize the entire solution process. Instead, it isolates the actual transferable bottleneck: after a geometric rotation, the original coordinates must be recovered in terms of transformed coordinates and substituted back into the original curve equation. Once this step is handled, the remaining work is routine algebra and checking, so storing the whole derivation would add little reusable value.

\begin{figure}[htbp]
    \centering
    \includegraphics[width=0.92\linewidth]{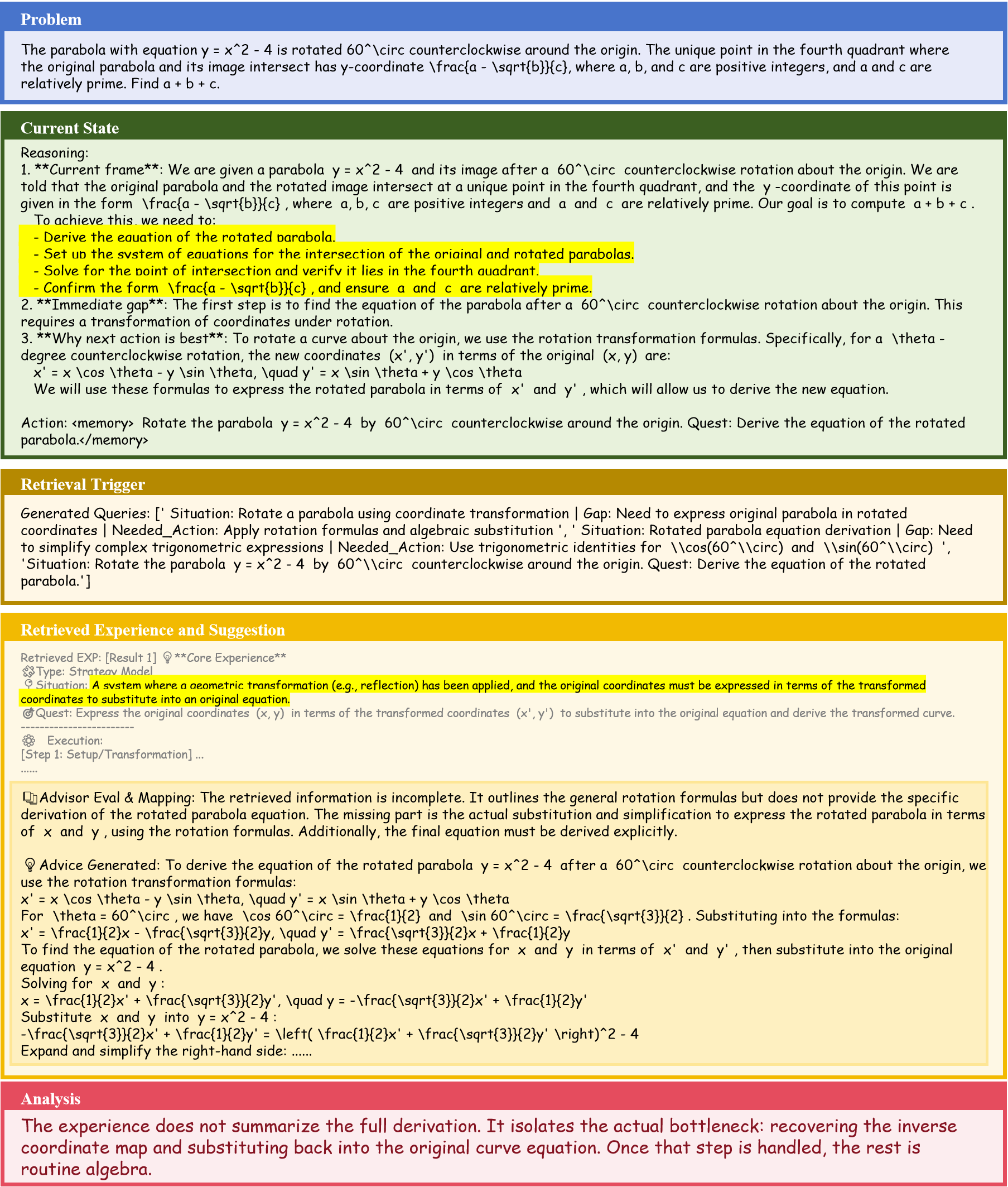}
    \caption{
    Representative bottleneck-focused experience construction. The stored experience does not preserve the full rotated-parabola derivation; it isolates the reusable bottleneck of inverse coordinate substitution after rotation.
    }
    \label{fig:app_bottleneck_rotated_parabola_case}
\end{figure}

\textbf{More broadly, bottleneck-focused construction is beneficial beyond our own utilization framework.} As discussed in Section~\ref{sec:related_work}, prior experience-based methods differ not only in how experience is used, but also in how it is constructed. Trajectory-based methods store full task-solving traces as experience, whereas methods such as ReasoningBank and DC aim to extract more focused experience centered on key difficulties or critical reasoning points. The overall results in Table~\ref{tab:main_results} suggest that, in most settings, experience constructed around key bottlenecks is more effective than storing full trajectories. A likely reason is that bottleneck-focused experience is more reusable across tasks and introduces less irrelevant information.

However, this advantage is realized only when the bottleneck is genuinely isolated, rather than when the original trajectory is merely broken into smaller pieces. This helps explain one exception in our main results: on Qwen3 series models, ReasoningBank performs worse than trajectory-based experience. Although ReasoningBank is intended to organize experience around key reasoning points, we find that under Qwen3-32B and Qwen3-14B it often produces experience that resembles a step-by-step decomposition of the original solution, treating many local steps as independent bottlenecks. As a result, its content remains close to trajectory-style experience and fails to realize the full advantage of true bottleneck-focused construction. By contrast, under GPT-5.4-mini, the constructed experience is more clearly centered on decisive bottlenecks, which is more consistent with the intended design and also more effective in practice.


Figure~\ref{fig:app_reasoningbank_bottleneck_comparison} shows this contrast on a recursive-function problem. The Qwen3-32B version splits the solution into procedural items such as understanding the definition, computing values near the base case, identifying a pattern, and applying it. This mirrors the original solving trajectory. The GPT-5.4-mini version instead centers on the decisive failure mode: naive upward propagation is not justified, and the nested recursive branch must be reconciled algebraically with the explicit branch. In this case, one construction fragments the trajectory into steps, while the other isolates what actually makes the problem difficult.

\begin{figure}[htbp]
    \centering
    \includegraphics[width=0.92\linewidth]{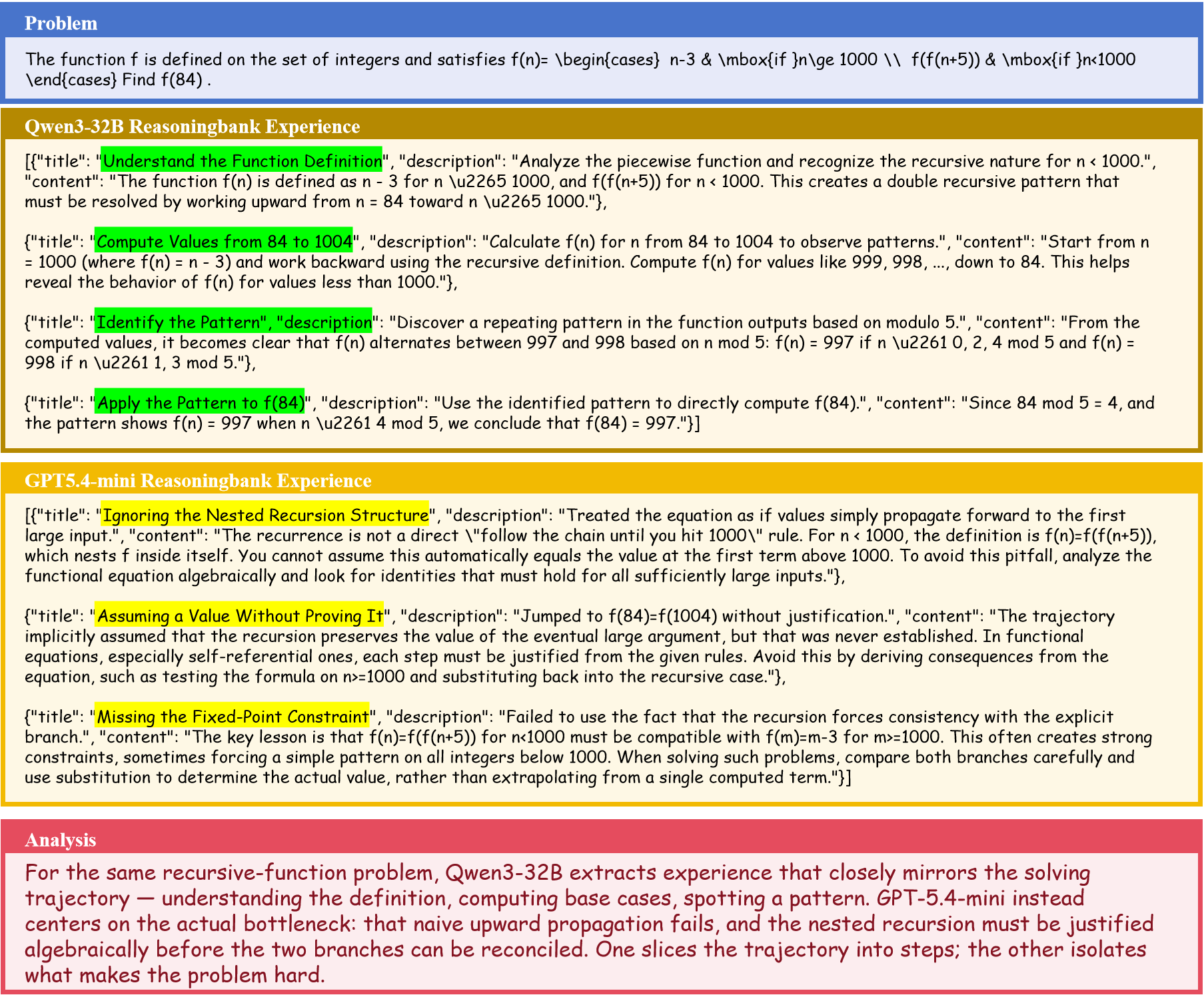}
    \caption{
    Comparison of ReasoningBank-style experience construction across backbones. Qwen3-32B over-fragments the original recursive-function trajectory into procedural steps, while GPT-5.4-mini more clearly isolates the decisive bottleneck.
    }
    \label{fig:app_reasoningbank_bottleneck_comparison}
\end{figure}
\FloatBarrier

\subsubsection{Actionable experience design.}
\label{app:actionabel}

\textbf{Experience must be actionable: to help resolve a bottleneck, it should provide both decision support and execution details.}

For a concrete reasoning bottleneck, it is not enough for experience to only indicate what strategy should be used, nor is it sufficient to only provide isolated execution details. Effective support requires both: decision support to determine the appropriate next move, and detail support to make that move executable in the current context.

To verify this, we conduct an additional ablation in which the retrieved experience used for next-move guidance is restricted to only one component: either \textit{Decision Support} or \textit{Detail Support}. The results are also summarized in Figure~\ref{fig:exp2}, where \textit{+Dec} and \textit{+Detail} denote these two variants.

\textbf{Neither decision support alone nor detail support alone provides stable gains over the base framework.} As shown in Figure~\ref{fig:exp2}(a), when used in isolation, both \textit{+Dec} and \textit{+Detail} fail to consistently improve over \textit{Ours (w/o EXP)}. This suggests that neither component is sufficient by itself: decision support without executable details may leave the model knowing what direction to take but not how to carry it out, while detail support without decision guidance may provide useful fragments without clarifying how they should be used to resolve the current bottleneck.

\textbf{Insufficiently actionable experience also increases reasoning cost.} Figures~\ref{fig:exp2}(b) and \ref{fig:exp2}(c) show that when experience content is incomplete, the model tends to produce longer reasoning trajectories and trigger retrieval more frequently. A plausible explanation is that when the retrieved experience does not fully resolve the current bottleneck, the model must continue exploring, asking for more support, or revisiting similar difficulties, which leads to more reasoning steps and more retrieval calls.

\textbf{These results support the need for actionable experience design.} To effectively help the model cross a bottleneck, experience should contain both the decision needed to choose the next move and the details required to execute that move. Only when these two components are provided together can experience serve as compact and sufficient support for local reasoning progression.

\subsubsection{Summary of analysis.}

\textbf{Effective experience for complex problem solving must be retrieved on demand, applied locally, and constructed for direct execution.}

Our analysis shows that HippoSpark satisfies these requirements along three dimensions. It retrieves experience only when a bottleneck emerges (\textbf{difficulty-triggered retrieval}), uses it only to guide the current transition (\textbf{next-move-guiding integration}), and constructs it around decisive difficulties with executable support (\textbf{bottleneck-focused and actionable construction}). These properties jointly ensure that experience remains relevant, targeted, and effective throughout the reasoning process.

\subsection{Ablation Study}
\label{app:ablation_study}

\begin{table*}[t]
\centering
\caption{
Ablation study on math reasoning.
The first block evaluates progressive framework components, while the second compares different forms of experience support on top of verification.
}
\label{tab:math_ablation}
\footnotesize
\setlength{\tabcolsep}{2.2pt}
\renewcommand{\arraystretch}{1.06}
\begin{adjustbox}{width=0.92\textwidth}
\begin{tabular}{>{\raggedright\arraybackslash}p{2.0cm}*{8}{r@{\hspace{2pt}}l}}
\toprule
\multirow{3}{*}{\textbf{Method}}
& \multicolumn{8}{c}{\textbf{Qwen3-32B}}
& \multicolumn{8}{c}{\textbf{GPT-5.4 mini}} \\
\cmidrule(lr){2-9} \cmidrule(lr){10-17}
& \multicolumn{4}{c}{\textbf{AIME24}}
& \multicolumn{4}{c}{\textbf{AIME25}}
& \multicolumn{4}{c}{\textbf{AIME24}}
& \multicolumn{4}{c}{\textbf{AIME25}} \\
\cmidrule(lr){2-5} \cmidrule(lr){6-9}
\cmidrule(lr){10-13} \cmidrule(lr){14-17}
& \multicolumn{2}{c}{\small\textbf{Avg@1}$^{\uparrow}$}
& \multicolumn{2}{c}{\small\textbf{Best@5}$^{\uparrow}$}
& \multicolumn{2}{c}{\small\textbf{Avg@1}$^{\uparrow}$}
& \multicolumn{2}{c}{\small\textbf{Best@5}$^{\uparrow}$}
& \multicolumn{2}{c}{\small\textbf{Avg@1}$^{\uparrow}$}
& \multicolumn{2}{c}{\small\textbf{Best@5}$^{\uparrow}$}
& \multicolumn{2}{c}{\small\textbf{Avg@1}$^{\uparrow}$}
& \multicolumn{2}{c}{\small\textbf{Best@5}$^{\uparrow}$} \\
\midrule

\rowcolor{groupgray}
\multicolumn{17}{l}{\textit{\textbf{Framework ablations}}} \\

Vanilla
& 0.253 & & 0.267 & & 0.147 & & 0.200 &
& 0.273 & & 0.300 & & 0.253 & & 0.300 & \\

+ Transition
& 0.333 & & 0.367 & & 0.220 & & 0.267 &
& 0.347 & & 0.467 & & 0.327 & & 0.400 & \\

+ Verification
& 0.393 & & 0.400 & & 0.253 & & 0.267 &
& 0.387 & & 0.500 & & 0.347 & & 0.400 & \\

\rowcolor{groupgray}
\multicolumn{17}{l}{\textit{\textbf{Experience-integration}}} \\

Global Context 
& 0.360 & \pctloss{8}
& 0.400 & 
& 0.280 & \pctgain{5}
& 0.333 & \pctgain{25}
& 0.447 & \pctgain{16}
& 0.533 & \pctgain{7}
& 0.360 & \pctgain{4}
& 0.467 & \pctgain{17} \\

\rowcolor{groupgray}
\multicolumn{17}{l}{\textit{\textbf{Experience-content}}} \\

+ Decision Sup.
& 0.420 & \pctgain{7}
& 0.500 & \pctgain{25}
& 0.240 & \pctloss{5}
& 0.300 & \pctgain{12}
& 0.447 & \pctgain{16}
& 0.533 & \pctgain{7}
& 0.340 & \pctloss{2}
& 0.400 & \\

+ Detailed Sup.
& 0.393 &
& 0.433 & \pctgain{8}
& 0.260 & \pctgain{3}
& 0.300 & \pctgain{12}
& 0.413 & \pctgain{7}
& 0.467 & \pctloss{7}
& 0.373 & \pctgain{8}
& 0.467 & \pctgain{17} \\

\textbf{Ours}
& \textbf{0.487} & \pctgain{24}
& \textbf{0.533} & \pctgain{33}
& \textbf{0.313} & \pctgain{24}
& \textbf{0.367} & \pctgain{38}
& \textbf{0.507} & \pctgain{31}
& \textbf{0.600} & \pctgain{20}
& \textbf{0.413} & \pctgain{19}
& \textbf{0.500} & \pctgain{25} \\
\bottomrule
\end{tabular}
\end{adjustbox}
\end{table*}

We conduct ablations to isolate the contributions of three aspects of HippoSpark: 
(i) the transition-centric reasoning framework itself, 
(ii) the way experience is integrated into the reasoning process, and 
(iii) the two-part design of each experience unit. 
The results are reported in Table~\ref{tab:math_ablation}. 
The upper block evaluates the reasoning framework without external experience, including transition-based problem solving and verification before commitment. 
The middle block compares different experience integration strategies. 
The lower block studies the content of experience, where each experience unit consists of a \emph{decision cue} for selecting an appropriate strategy for the current transition and \emph{detailed support} for carrying out that strategy. 
For the experience-related ablations, we also report the relative improvement over the no-memory transition-based framework with verification.

\textbf{The transition-centric reasoning framework is itself beneficial, even without external experience.}
Compared with vanilla generation, explicitly organizing problem solving as a sequence of local transitions consistently improves performance across backbones and benchmarks. 
This supports our view that complex reasoning is better modeled as a stepwise search process over intermediate states, rather than as a direct mapping from problem to answer. 
Adding verification before commitment further improves results, indicating that locally proposed moves should be checked before execution, as errors at a single transition can propagate and distort downstream reasoning.

\textbf{How experience is integrated matters substantially.}
Injecting retrieved experience as \emph{global context} into the reasoning chain brings only limited and less stable gains compared with our full model. 
Although the experience is still retrieved on demand, directly appending it to the ongoing reasoning context forces the model to carry the retrieved content throughout subsequent generation. 
This differs from our next-move-guiding integration, where experience is used only to support the current transition and is discarded afterward. 
We conjecture that global-context integration weakens reasoning efficiency because the model must continuously attend to additional experience content that is no longer directly relevant, thereby diluting focus on the main problem-solving process~\citep{shi2023large,liu2024lost}.

\textbf{The two experience components are complementary.}
Decision experience helps the model determine how the current transition should proceed, while detailed support provides the concrete knowledge needed to execute that decision reliably. 
Removing either component consistently weakens overall performance, whereas the full model performs best across all settings. 
This shows that effective support at a reasoning bottleneck requires both identifying the right next move and grounding that move in actionable execution support. {The relative contribution of the two components varies across benchmarks.}
Prior work suggests that AIME24, as an older benchmark, is more likely to have been exposed to current language models than AIME25~\citep{valsaime2025,matharena2025}. 

Consistent with this, retaining only decision experience still yields substantial gains on AIME24, suggesting that for more exposed problems, strategy-level guidance is often sufficient. 
On AIME25, in contrast, retaining only decision experience leads to weaker or less stable improvements, while retaining detailed support remains more consistently helpful. 
This suggests that for less exposed and more genuinely challenging problems, execution-level support becomes more important.

\subsection{Experience Construction Analysis}
\label{app:experience_construction_analysis}
\begin{table}[t]
\centering
\caption{
Statistics of experience construction on 100 AIME problems using Qwen3-32B.
Input and output tokens are averaged per problem.
For DC-CU and DC-RS, token counts are summed over the generator and cheatsheet stages.
For Reasoning Bank, although 409 raw items are extracted in total, they are used as bundled task-level memory units during retrieval, so the effective number of retrievable units is reported as 100.
}
\label{tab:exp_construction_stats}
\footnotesize
\setlength{\tabcolsep}{3.5pt}
\renewcommand{\arraystretch}{1.08}
\begin{adjustbox}{width=0.8\columnwidth}
\begin{tabular}{>{\raggedright\arraybackslash}p{2.1cm}ccccc}
\toprule
\multirow{2}{*}{\textbf{Method}}
& \multicolumn{5}{c}{\textbf{Qwen3-32B on 100 AIME Problems}} \\
\cmidrule(lr){2-6}
& \textbf{Avg. Input}
& \textbf{Avg. Output}
& \textbf{Avg. Exp. Len.}
& \textbf{\# Units}
& \textbf{Remark} \\
\midrule

\rowcolor{groupgray}
\multicolumn{6}{l}{\textit{\textbf{Trajectory-/task-level experience}}} \\

Trajectory Exp
& 224.83
& 2493.64
& 2493.64
& 100
& trajectory-level \\

Reasoning Bank
& 3476.26
& 1967.59
& 430.27
& 100
& bundled per task \\

\rowcolor{groupgray}
\multicolumn{6}{l}{\textit{\textbf{Distilled experience}}} \\

DC-CU
& 11169.34
& 6198.06
& 215.32
& 32
& gen. + cheatsheet \\

DC-RS
& 13261.88
& 5499.10
& 174.85
& 41
& gen. + cheatsheet \\

\rowcolor{groupgray}
\multicolumn{6}{l}{\textit{\textbf{Bottleneck-level experience}}} \\

\textbf{Ours}
& \textbf{7344.26}
& \textbf{6993.82}
& \textbf{360.15}
& \textbf{406}
& actionable unit \\
\bottomrule
\end{tabular}
\end{adjustbox}
\end{table}

\begin{table}[t]
\centering
\caption{
Statistics of the constructed experience memory across three domains and three models.
Each experience unit contains one decision-level entry and its associated supporting details.
We report the total number of experience units, decision units, supporting details, and the average number of supporting details linked to each decision unit.
}
\label{tab:memory_stats_by_domain}
\footnotesize
\setlength{\tabcolsep}{3.8pt}
\renewcommand{\arraystretch}{1.08}
\begin{adjustbox}{width=0.72\columnwidth}
\begin{tabular}{>{\raggedright\arraybackslash}p{2.45cm}ccc}
\toprule
\multirow{2}{*}{\textbf{Statistic}}
& \multicolumn{3}{c}{\textbf{Experience Memory by Domain}} \\
\cmidrule(lr){2-4}
& \textbf{AIME (100)}
& \textbf{GPQA-Diamond (57)}
& \textbf{BigCodeBench (40)} \\
\midrule

\rowcolor{groupgray}
\multicolumn{4}{c}{\textbf{Qwen3-32B}} \\

\rowcolor{groupgray}
\multicolumn{4}{l}{\textit{\textbf{Memory scale}}} \\
Experience Cards
& 406
& 210
& 118 \\
Decision Sup.
& 406
& 210
& 118 \\
Details Sup.
& 352
& 262
& 170 \\

\rowcolor{groupgray}
\multicolumn{4}{l}{\textit{\textbf{Memory structure}}} \\
Avg. Detail Count
& 1.54
& 1.74
& 1.66 \\
\midrule

\rowcolor{groupgray}
\multicolumn{4}{c}{\textbf{Qwen3-14B}} \\

\rowcolor{groupgray}
\multicolumn{4}{l}{\textit{\textbf{Memory scale}}} \\
Experience Cards
& 478
& 304
& 115 \\
Decision Sup.
& 478
& 304
& 115 \\
Details Sup.
& 336
& 211
& 137 \\

\rowcolor{groupgray}
\multicolumn{4}{l}{\textit{\textbf{Memory structure}}} \\
Avg. Detail Count
& 1.34
& 1.31
& 1.69 \\
\midrule

\rowcolor{groupgray}
\multicolumn{4}{c}{\textbf{GPT-5.4-mini}} \\

\rowcolor{groupgray}
\multicolumn{4}{l}{\textit{\textbf{Memory scale}}} \\
Experience Cards
& 437
& 238
& 85 \\
Decision Sup.
& 437
& 238
& 85 \\
Details Sup.
& 469
& 316
& 223 \\

\rowcolor{groupgray}
\multicolumn{4}{l}{\textit{\textbf{Memory structure}}} \\
Avg. Detail Count
& 1.67
& 1.37
& 2.86 \\

\bottomrule
\end{tabular}
\end{adjustbox}
\end{table}

We compare how different methods construct experience, focusing on both \textit{quality} and \textit{cost}. Specifically, we report the statistics of the experience pools produced by our method and by representative baselines, including the number of experience items, the average length of each item, and the total token consumption during construction. For methods whose experience is still consumed at the task level during retrieval, we explicitly clarify the effective retrieval unit, even if multiple experience fragments are extracted from a single problem.

Beyond corpus-level statistics, we also present side-by-side examples of the experience constructed from the same problem by different methods. This comparison helps illustrate the key qualitative difference of our approach: the experience constructed by HippoSpark is centered on concrete bottlenecks and provides actionable support for subsequent reasoning, whereas prior methods tend to produce more global, trajectory-level, or less operational experience.


The different methods also differ substantially in what counts as one experience unit. As shown in Table~\ref{tab:exp_construction_stats}, Trajectory Exp stores one full solution trajectory per problem, so on the 100-problem AIME experience set it naturally produces 100 experience units. ReasoningBank may extract multiple distilled items from a single problem, but these items are still retrieved and used together at the task level. Therefore, we treat its effective retrieval unit as one bundled memory per problem, and report 100 effective units in the main comparison. For reference, the raw distilled content is still more fine-grained than this task-level count. DC-CU and DC-RS use a different construction pattern: the cheatsheet is updated cumulatively across problems, so we treat the final stabilized cheatsheet as the effective experience memory used at inference time. By contrast, HippoSpark constructs bottleneck-level experience cards, producing many more fine-grained reusable units from the same 100-problem set.


The average length of each experience unit also reflects different design choices. Trajectory Exp has the longest units because each item is an entire task-solving trace. ReasoningBank shortens this considerably by distilling trajectories into several compact pieces, but those pieces are still bundled back into a task-level unit during use. DC-CU and DC-RS produce even shorter average units because their final cheatsheets are organized as concise reusable entries. HippoSpark lies between these extremes: its experience cards are much shorter than full trajectories, but still long enough to contain both a decision-level cue and the execution details needed to make the next move actionable. This intermediate length is consistent with our design goal of constructing compact but directly usable bottleneck-level experience.


Construction cost varies widely across methods. As Table~\ref{tab:exp_construction_stats} shows, trajectory-based methods are expensive mainly because they preserve long generation traces, while DC-style methods incur substantial cost from repeatedly updating and maintaining the cheatsheet over time. For DC-CU and DC-RS, we count both generator-side and cheatsheet-side tokens in the construction budget. HippoSpark also requires a nontrivial construction budget, since each experience card is generated around a local bottleneck and includes structured support for later reuse. However, this cost is spent on producing more targeted and more reusable units rather than on preserving full trajectories or maintaining a growing global cheatsheet.

Table~\ref{tab:memory_stats_by_domain} further shows that this construction pattern is stable across domains and backbones. Unlike Table~\ref{tab:exp_construction_stats}, which compares different methods on the AIME experience set, Table~\ref{tab:memory_stats_by_domain} focuses only on HippoSpark and examines the internal structure of the constructed memory on AIME, GPQA-Diamond, and BigCodeBench for Qwen3-32B, Qwen3-14B, and GPT-5.4-mini. Across all settings, the number of experience cards is exactly the number of decision-level units, because each card is anchored by one bottleneck-centered decision cue. The number of supporting details varies with the benchmark and backbone: for example, Qwen3-32B constructs 406/210/118 cards on AIME, GPQA-Diamond, and BigCodeBench, while GPT-5.4-mini constructs 437/238/85 cards. The average detail count remains in a narrow range for AIME and GPQA, and becomes larger on BigCodeBench for GPT-5.4-mini. This suggests that the construction process repeatedly produces the same actionable memory structure: one decision-level cue linked to a small set of execution details, rather than isolated short hints or full task trajectories.



Figures~\ref{fig:app_construct_ours_case}--\ref{fig:app_construct_dcrs_case} compare experience constructed from the same combinatorial counting problem. The problem asks for four-digit numbers beginning with 1 and having exactly two identical digits. The decisive bottleneck is not arithmetic execution, but setting up disjoint cases so that the repeated digit is either the fixed leading digit or a different digit, without overcounting.

\begin{figure}[htbp]
    \centering
    \includegraphics[width=0.92\linewidth]{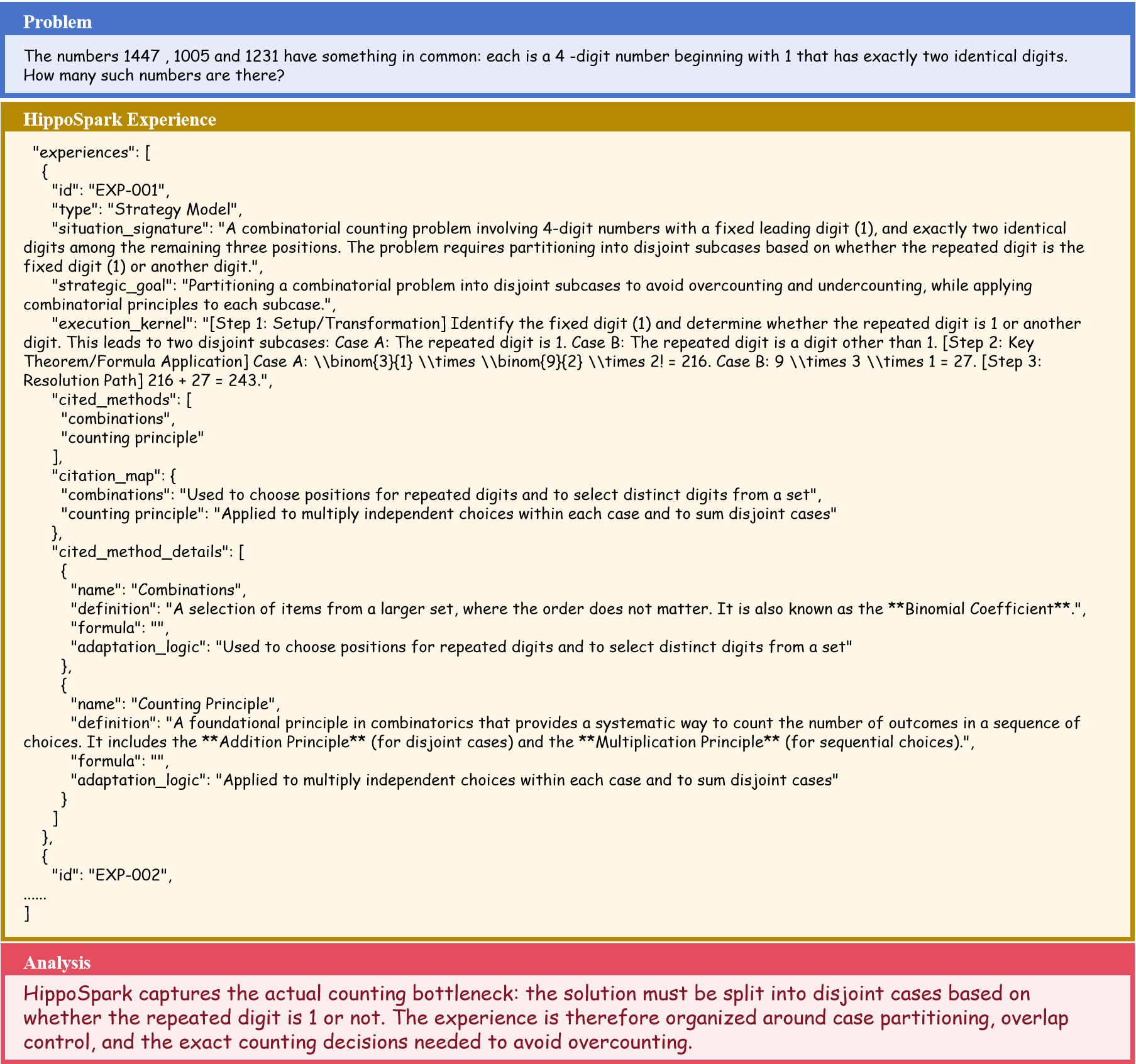}
    \caption{
    HippoSpark construction on the repeated-digit counting problem. The experience is organized around the true bottleneck: disjoint case partitioning and overlap control under the repeated-digit constraint.
    }
    \label{fig:app_construct_ours_case}
\end{figure}

HippoSpark captures this bottleneck directly. Its experience card records the situation signature, the strategic goal of avoiding overcounting, and the concrete execution kernel for separating the two cases and applying the counting principle. This makes the unit reusable for later problems with similar repetition constraints, rather than merely preserving this problem's final derivation.

\begin{figure}[htbp]
    \centering
    \includegraphics[width=0.92\linewidth]{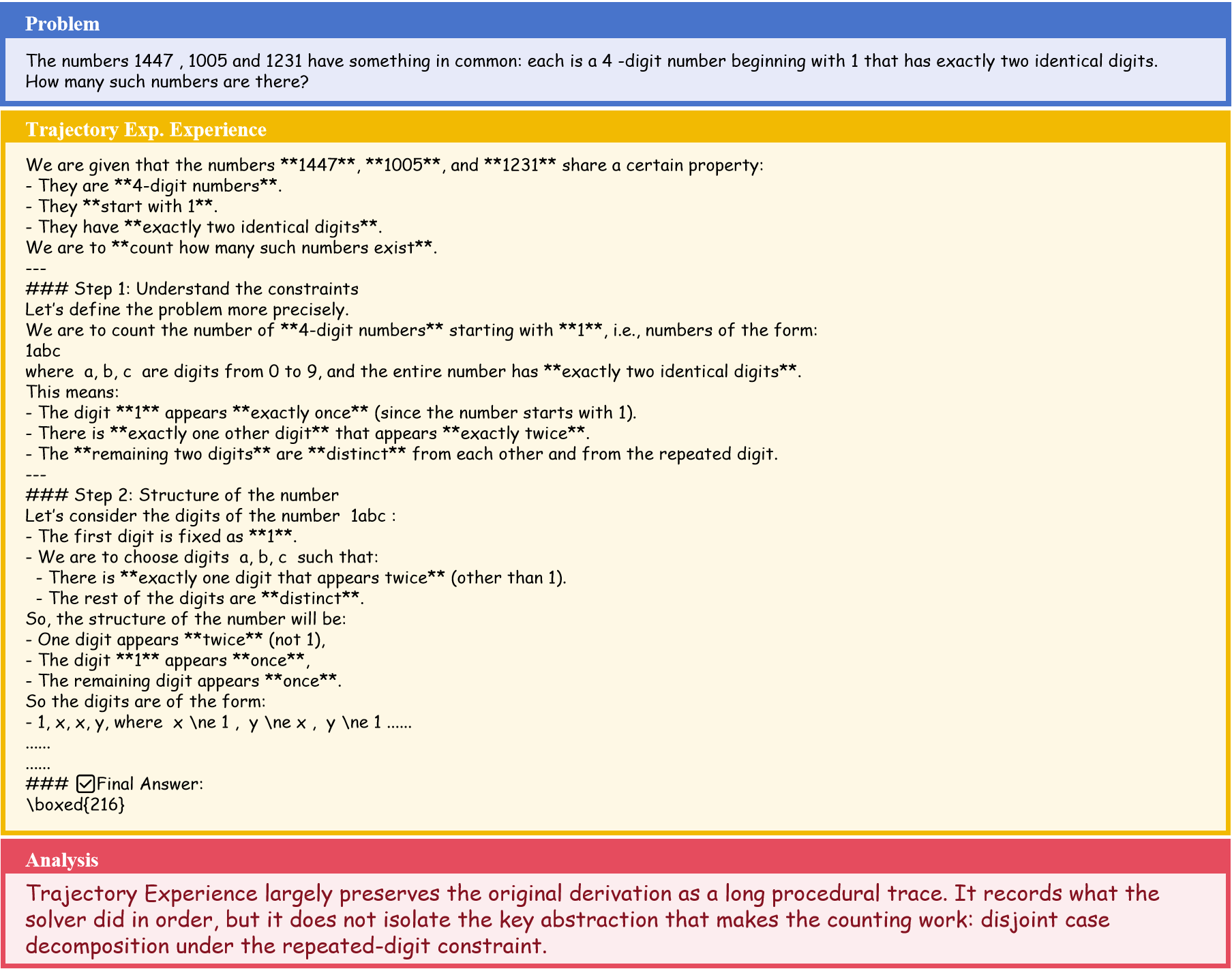}
    \caption{
    Trajectory Exp construction on the same problem. The experience largely preserves the original derivation as a long procedural trace, recording what the solver did rather than isolating why the counting strategy works.
    }
    \label{fig:app_construct_trajectory_case}
\end{figure}

By contrast, Trajectory Exp stores the solving process almost as written. It includes problem restatement, constraint interpretation, structure description, and the final answer. Although correct, this format leaves the reusable insight implicit: the reason the solution works is the disjoint decomposition of repeated-digit cases.

\begin{figure}[htbp]
    \centering
    \includegraphics[width=0.92\linewidth]{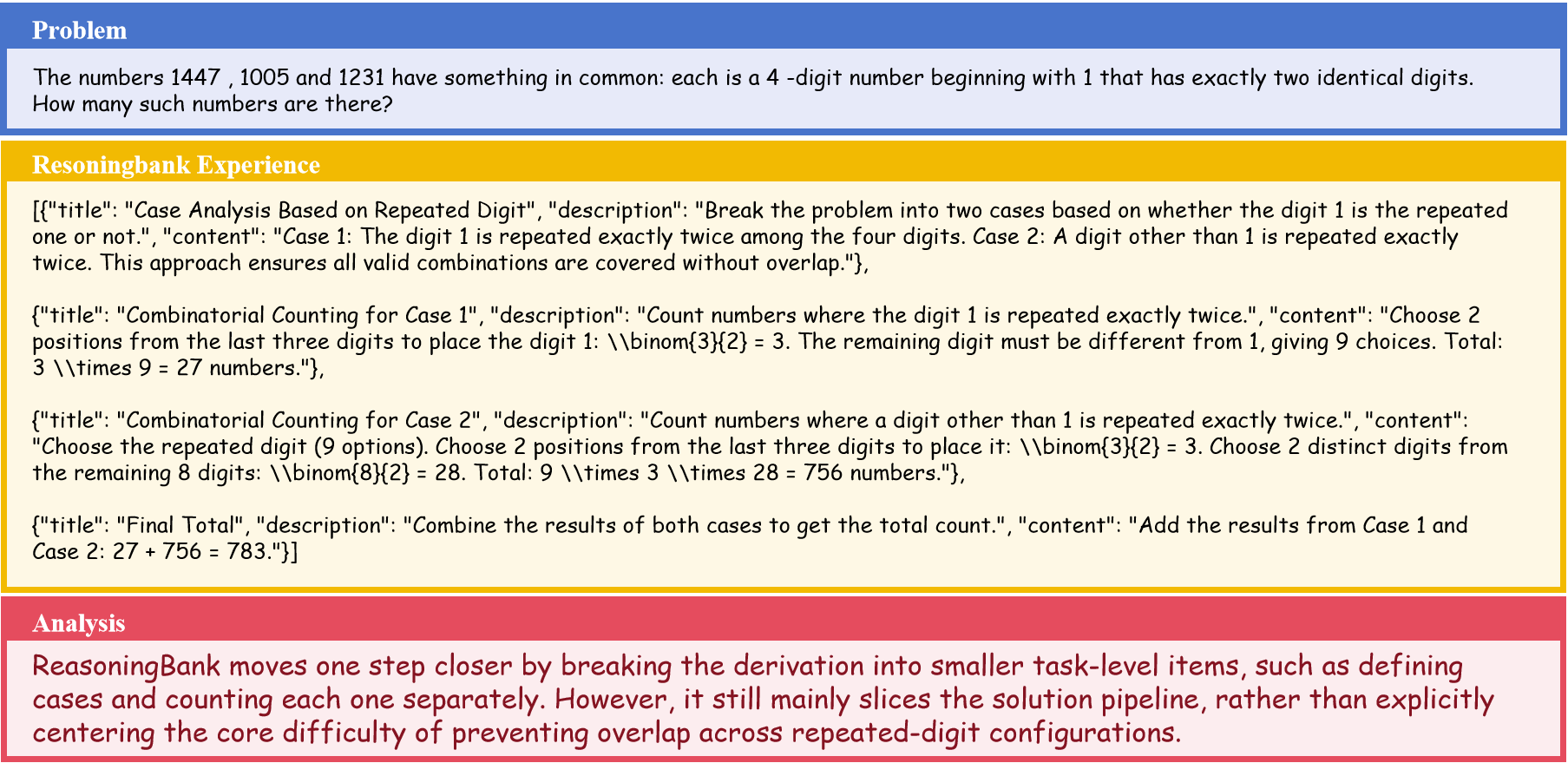}
    \caption{
    ReasoningBank construction on the same problem. It breaks the solution into smaller task-level items, but still mainly follows the procedural solution pipeline.
    }
    \label{fig:app_construct_reasoningbank_case}
\end{figure}

ReasoningBank moves closer to a structured explanation by splitting the derivation into case-analysis items. However, the resulting memory still mainly mirrors the solution pipeline: define the cases, count each case, and add them. It is more compact than a full trajectory, but it does not explicitly emphasize the underlying risk of overlap across repeated-digit configurations as the central reusable bottleneck.

\begin{figure}[htbp]
    \centering
    \includegraphics[width=0.92\linewidth]{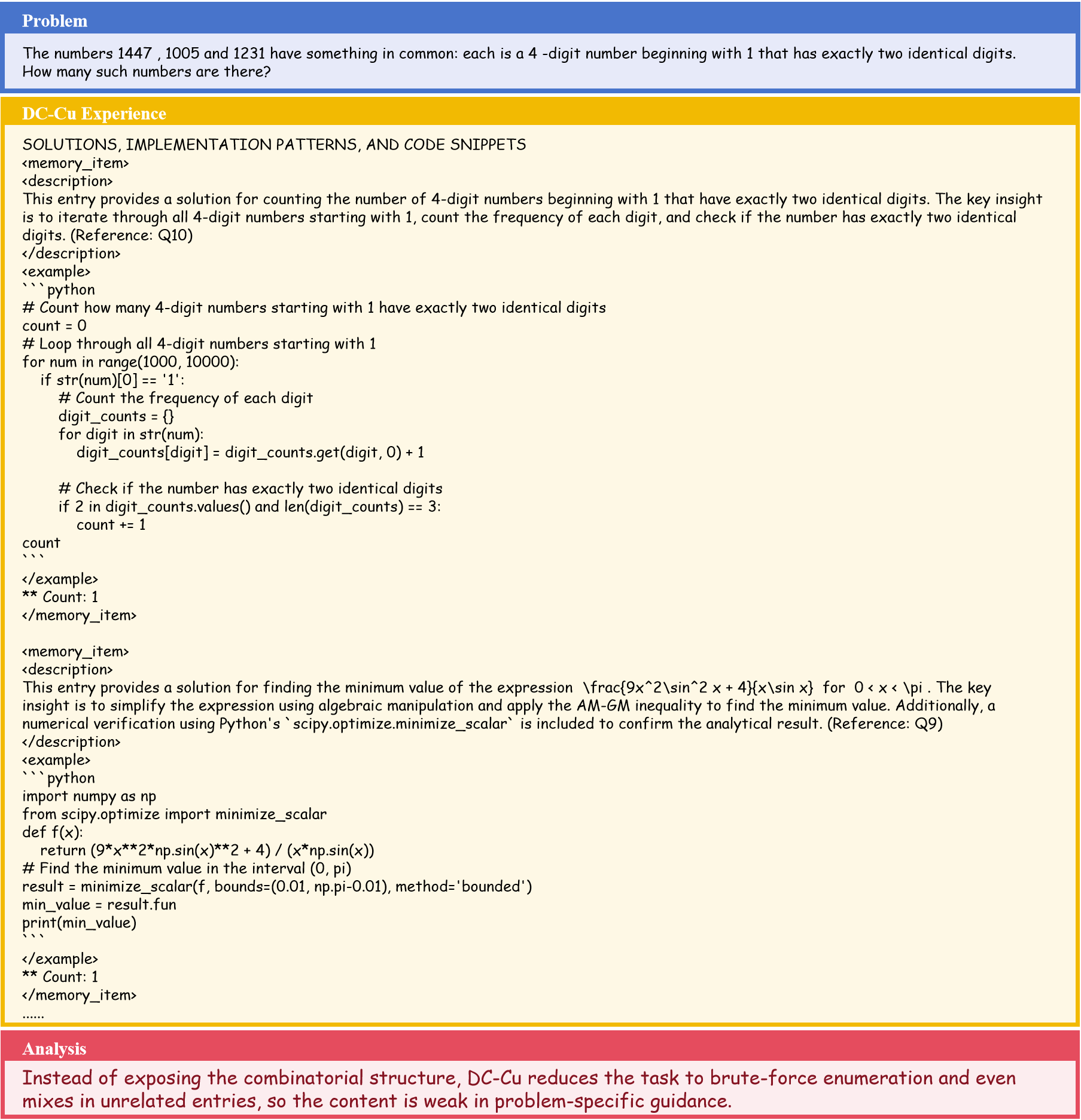}
    \caption{
    DC-CU construction on the same problem. The memory falls back to brute-force enumeration and includes unrelated entries, making the experience weak as problem-specific guidance.
    }
    \label{fig:app_construct_dccu_case}
\end{figure}

DC-CU is less targeted. Instead of exposing the combinatorial structure, it reduces the task to enumerating candidate numbers and checking digit frequencies. The same memory also contains unrelated entries, so the effective guidance is diluted and less useful for deciding a principled next move.

\begin{figure}[htbp]
    \centering
    \includegraphics[width=0.92\linewidth]{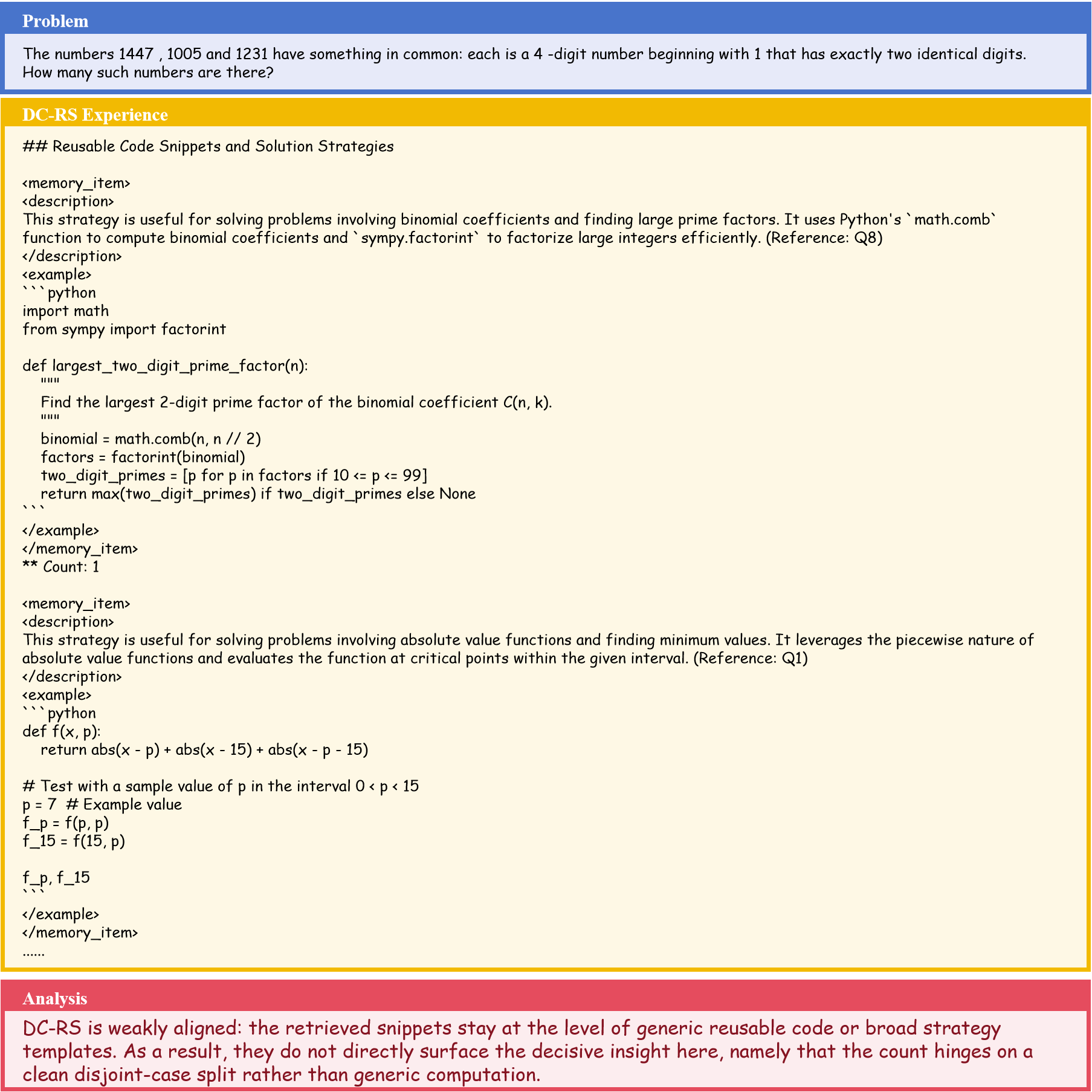}
    \caption{
    DC-RS construction on the same problem. The retrieved snippets remain generic and do not surface the decisive disjoint-case structure needed for this counting task.
    }
    \label{fig:app_construct_dcrs_case}
\end{figure}
\FloatBarrier

DC-RS is also weakly aligned with the actual bottleneck. Its snippets remain at the level of generic reusable code or broad strategy templates, such as binomial coefficients or absolute-value minimization. As a result, it does not directly surface the key insight that correctness hinges on a clean disjoint-case split rather than generic computation.

\subsection{Inference-Time Efficiency Analysis}
\label{app:inference_token_analysis}
\begin{table*}[t]
\centering
\caption{
Inference-time token consumption and Avg@1 accuracy on AIME24 and AIME25 for Qwen3-32B.
Input tokens count the full prompt budget used at test time, including any retrieved or maintained experience context.
Output tokens count the generated reasoning and answer tokens.
}
\label{tab:inference_main_qwen32b}
\footnotesize
\setlength{\tabcolsep}{3.8pt}
\renewcommand{\arraystretch}{1.08}
\begin{adjustbox}{width=0.72\textwidth}
\begin{tabular}{>{\raggedright\arraybackslash}p{2.4cm}cccccc}
\toprule
\multirow{2}{*}{\textbf{Method}}
& \multicolumn{3}{c}{\textbf{AIME24 (Qwen3-32B)}}
& \multicolumn{3}{c}{\textbf{AIME25 (Qwen3-32B)}} \\
\cmidrule(lr){2-4} \cmidrule(lr){5-7}
& \textbf{Avg. Input} & \textbf{Avg. Output} & \textbf{Avg@1}
& \textbf{Avg. Input} & \textbf{Avg. Output} & \textbf{Avg@1} \\
\midrule
Vanilla
& 211.77 & 2350.77 & 0.253
& 130.03 & 2937.85 & 0.147 \\
CoT
& 218.77 & 2779.67 & 0.280
& 137.03 & 3304.11 & 0.180 \\
Self-Refine
& 2958.07 & 3433.17 & 0.307
& 3506.03 & 4108.53 & 0.247 \\
Trajectory Exp.
& 2837.10 & 2812.36 & 0.267
& 3380.57 & 3237.13 & 0.193 \\
ReasoningBank
& 678.63 & 2042.03 & 0.260
& 650.43 & 3070.33 & 0.207 \\
DC-CU
& 13608.57 & 6788.57 & 0.307
& 13972.85 & 7280.12 & 0.227 \\
DC-RS
& 8585.71 & 4881.82 & 0.293
& 10118.58 & 5293.41 & 0.253 \\
\rowcolor{lightblueweak}
\textbf{Ours}
& 54029.41 & 4333.45 & 0.487
& 90450.44 & 5744.23 & 0.313 \\
\bottomrule
\end{tabular}
\end{adjustbox}
\end{table*}

\begin{table}[t]
\centering
\caption{
Input- and output-token breakdown of HippoSpark on Qwen3-32B.
\textit{Cortex} denotes the main reasoning component, \textit{Memory} denotes retrieval-conditioned experience guidance, and \textit{Verify} denotes move verification before commitment.
}
\label{tab:inference_breakdown_qwen32b}
\footnotesize
\setlength{\tabcolsep}{3.8pt}
\renewcommand{\arraystretch}{1.08}
\begin{adjustbox}{width=0.5\columnwidth}
\begin{tabular}{>{\raggedright\arraybackslash}p{1.35cm}rrrr}
\toprule
\textbf{Split}
& \textbf{Cortex}
& \textbf{Memory}
& \textbf{Verify}
& \textbf{Total} \\
\midrule

\rowcolor{groupgray}
\multicolumn{5}{l}{\textit{\textbf{Input Token Breakdown}}} \\
AIME24
& 36972.66 & 10930.71 & 6126.04 & 54029.41 \\
AIME25
& 70156.13 & 12462.20 & 7832.11 & 90450.44 \\

\midrule
\rowcolor{groupgray}
\multicolumn{5}{l}{\textit{\textbf{Output Token Breakdown}}} \\
AIME24
& 2811.79 & 1359.81 & 161.85 & 4333.45 \\
AIME25
& 3757.22 & 1787.79 & 199.22 & 5744.23 \\

\bottomrule
\end{tabular}
\end{adjustbox}
\end{table}

We further analyze inference-time token consumption to understand the efficiency--performance trade-off of different methods. Table~\ref{tab:inference_main_qwen32b} reports the overall input/output token consumption and final Avg@1 accuracy of HippoSpark and all baselines on Qwen3-32B. Table~\ref{tab:inference_breakdown_qwen32b} further decomposes HippoSpark's own token usage into \textit{cortex}, \textit{memory}, and \textit{verify} components, separately for input and output tokens.

We focus this token analysis on Qwen3-32B because it is our main open-weight backbone for studying the explicit inference-time cost of the external agentic loop. For Qwen3-32B and Qwen3-14B, we run the models without their built-in thinking mode. In preliminary runs, enabling the model's internal reasoning mode often made the model less compliant with the required action protocol: instead of following the external state--action loop, it tended to complete the problem internally or ignore the requested tool/action format. Disabling thinking mode therefore makes the solver--memory--verifier workflow explicit and measurable. GPT-5.4-mini, by contrast, is a reasoning model with its own internal reasoning-token accounting, so its token usage is not directly comparable to the explicit prompt/output tokens measured for Qwen3-32B. We therefore report the detailed module-level token analysis on Qwen3-32B.






Table~\ref{tab:inference_main_qwen32b} shows that HippoSpark is not the most token-efficient method in raw usage. It consumes substantially more input tokens than Vanilla, CoT, trajectory-based experience, ReasoningBank, and DC-style baselines. However, this additional cost is accompanied by stronger effectiveness: HippoSpark achieves the best Avg@1 accuracy on both AIME24 and AIME25. This indicates that the extra inference budget is not merely overhead, but is associated with meaningful gains in reasoning quality.

Figures~\ref{fig:app_inference_aime24_barline} and \ref{fig:app_inference_aime25_barline} visualize this trade-off. The bar plots show average input and output tokens, while the line plot shows Avg@1 accuracy. The main pattern is consistent across both splits: cheaper methods consume fewer tokens but provide weaker reasoning performance, whereas HippoSpark spends more inference budget to support a more reliable reasoning process.

\begin{figure}[htbp]
    \centering
    \includegraphics[width=0.82\linewidth]{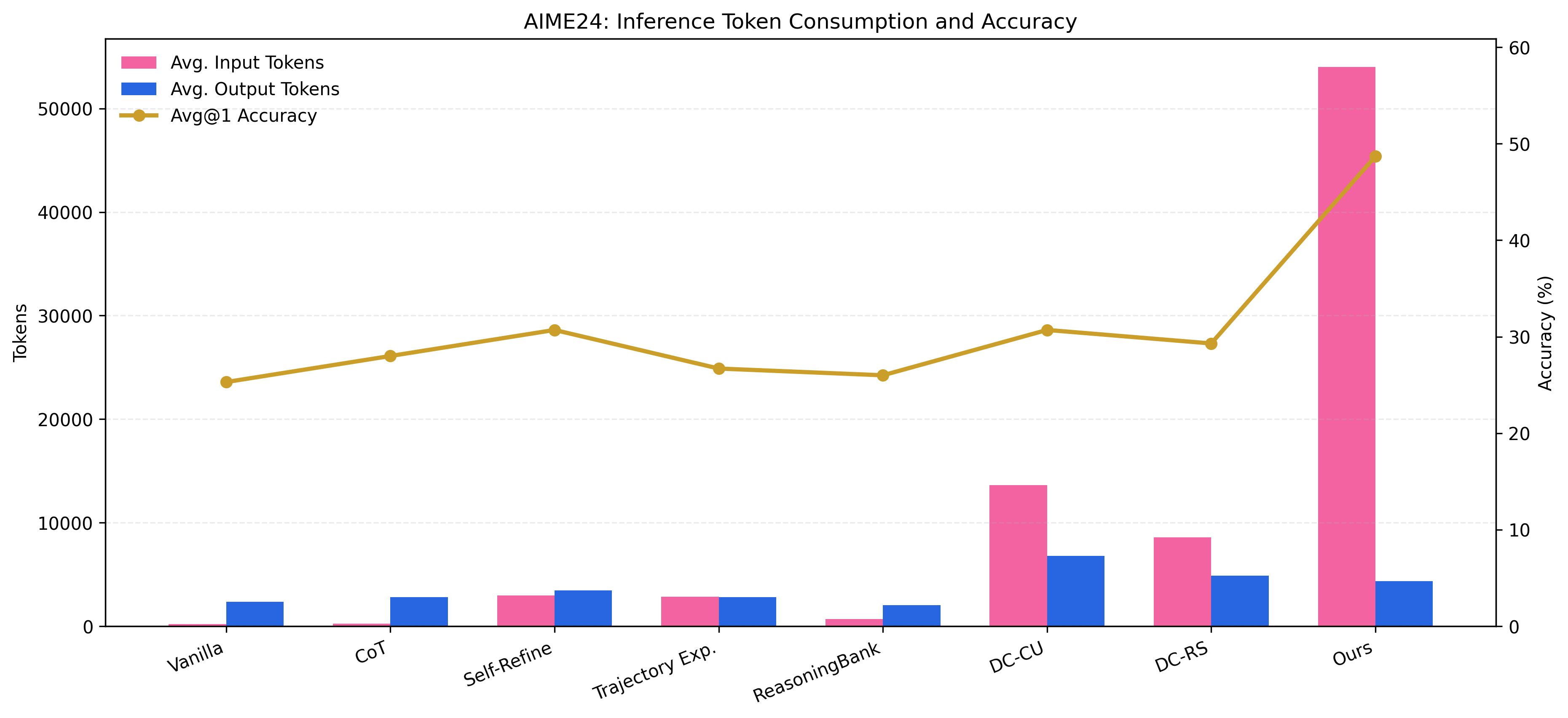}
    \caption{
    Inference-time token consumption and Avg@1 accuracy on AIME24 for Qwen3-32B. Bars show average input and output tokens, and the line shows final accuracy.
    }
    \label{fig:app_inference_aime24_barline}
\end{figure}

\begin{figure}[htbp]
    \centering
    \includegraphics[width=0.82\linewidth]{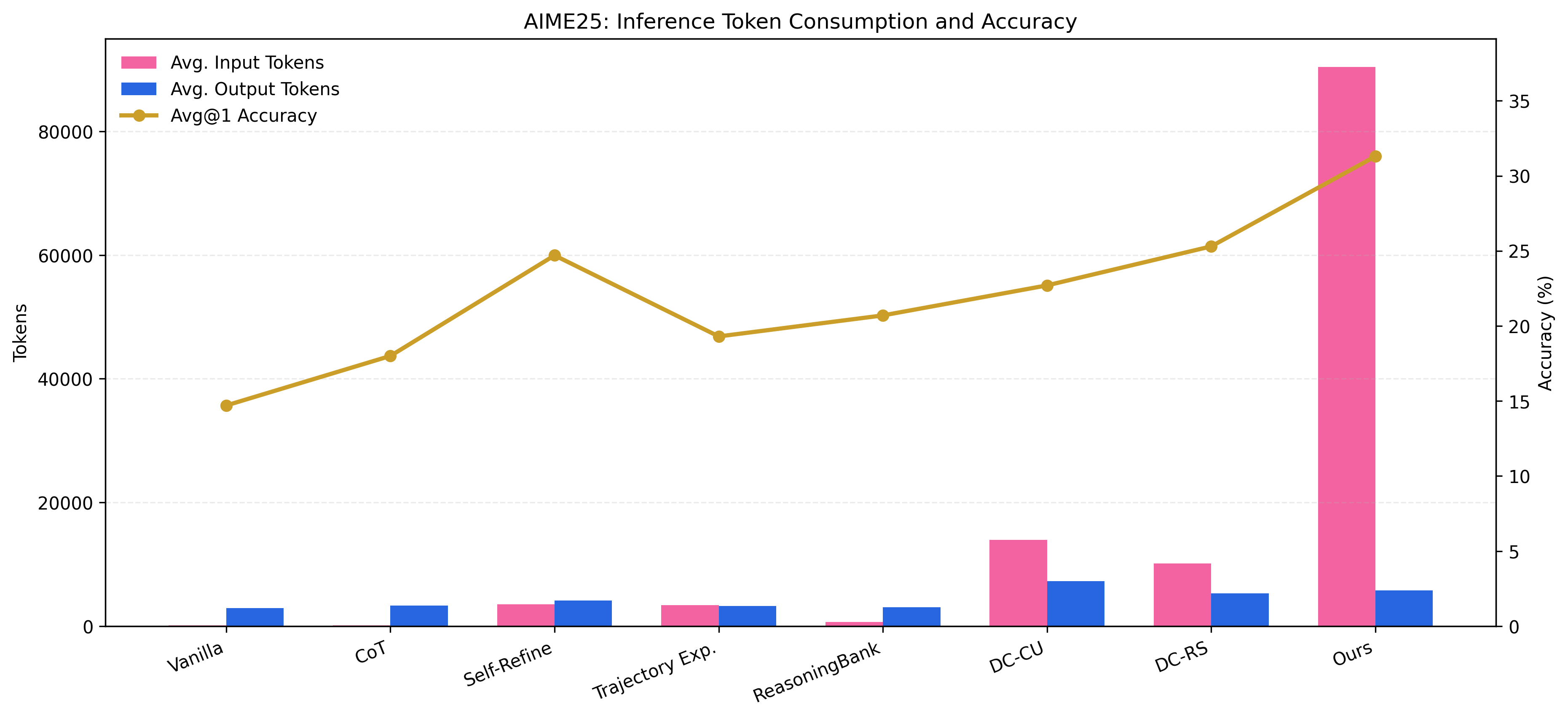}
    \caption{
    Inference-time token consumption and Avg@1 accuracy on AIME25 for Qwen3-32B. HippoSpark uses a larger input budget, but this cost is paired with the strongest overall performance.
    }
    \label{fig:app_inference_aime25_barline}
\end{figure}

Table~\ref{tab:inference_breakdown_qwen32b} breaks HippoSpark's inference budget into three components: \textit{cortex} tokens for main reasoning, \textit{memory} tokens for retrieval-conditioned experience guidance, and \textit{verify} tokens for checking proposed moves before commitment. This decomposition shows that the largest share of the input budget comes from the cortex component, which maintains and advances the main reasoning state. Memory is the second-largest input component, reflecting the cost of retrieving and adapting experience at difficult transitions. Verification is smaller but still nontrivial, because each proposed move must be checked locally before it is allowed to shape the downstream trajectory.

Figures~\ref{fig:app_inference_aime24_pies} and \ref{fig:app_inference_aime25_pies} present the same decomposition as input/output pie charts. On the input side, cortex dominates the budget, followed by memory and verification. On the output side, the same ordering remains: most generated tokens come from the main reasoning component, while memory synthesis and verification add smaller but targeted outputs. This cost structure is consistent with HippoSpark's design. The framework spends extra tokens not on preserving full retrieved trajectories, but on explicit support modules that help identify local bottlenecks, adapt retrieved experience, and block invalid moves before they contaminate later reasoning.

\begin{figure}[htbp]
    \centering
    \includegraphics[width=0.82\linewidth]{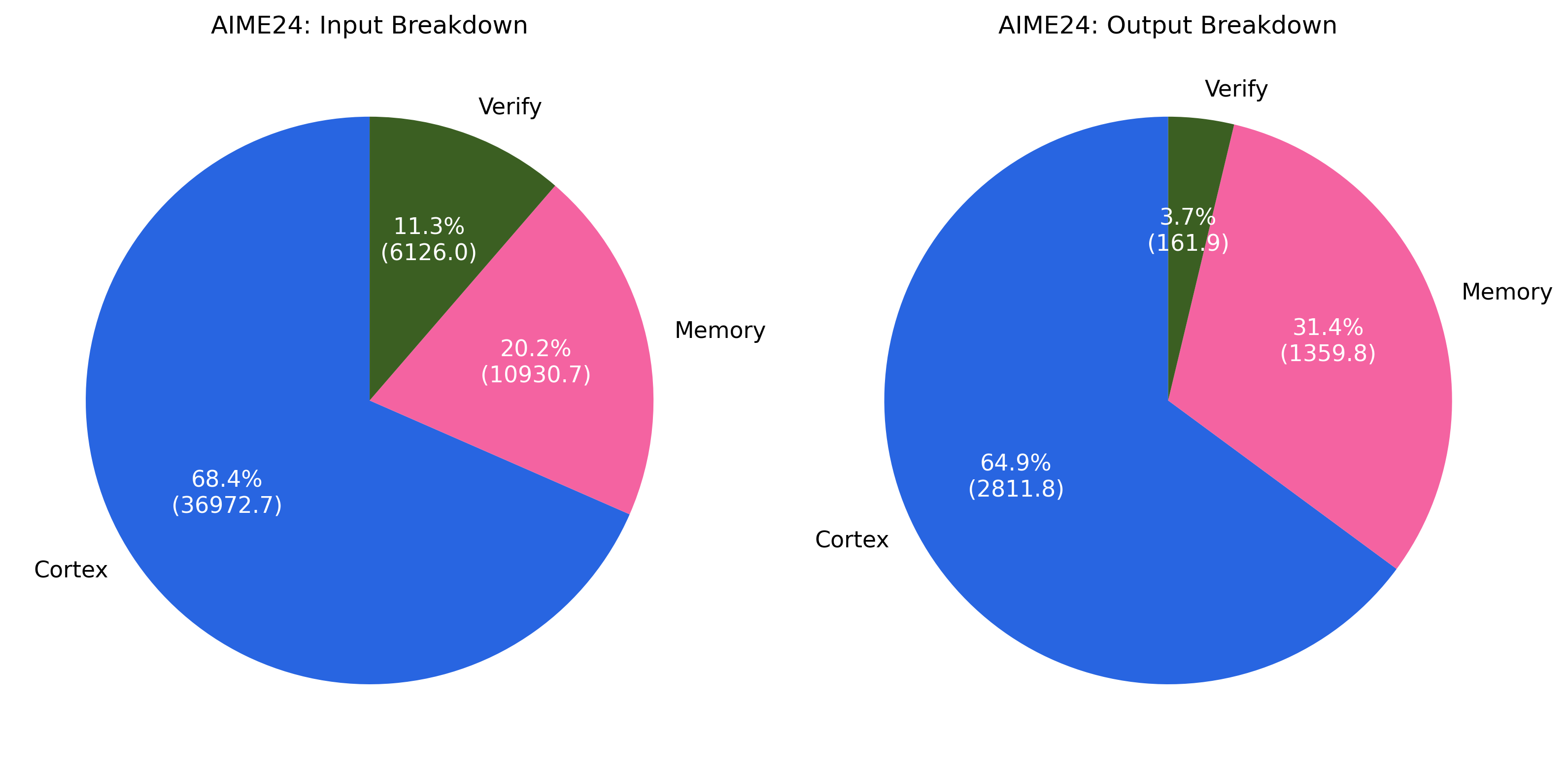}
    \caption{
    Input and output token breakdown of HippoSpark on AIME24. Cortex accounts for the largest share, while memory and verification provide targeted support for difficult transitions.
    }
    \label{fig:app_inference_aime24_pies}
\end{figure}

\begin{figure}[h]
    \centering
    \includegraphics[width=0.82\linewidth]{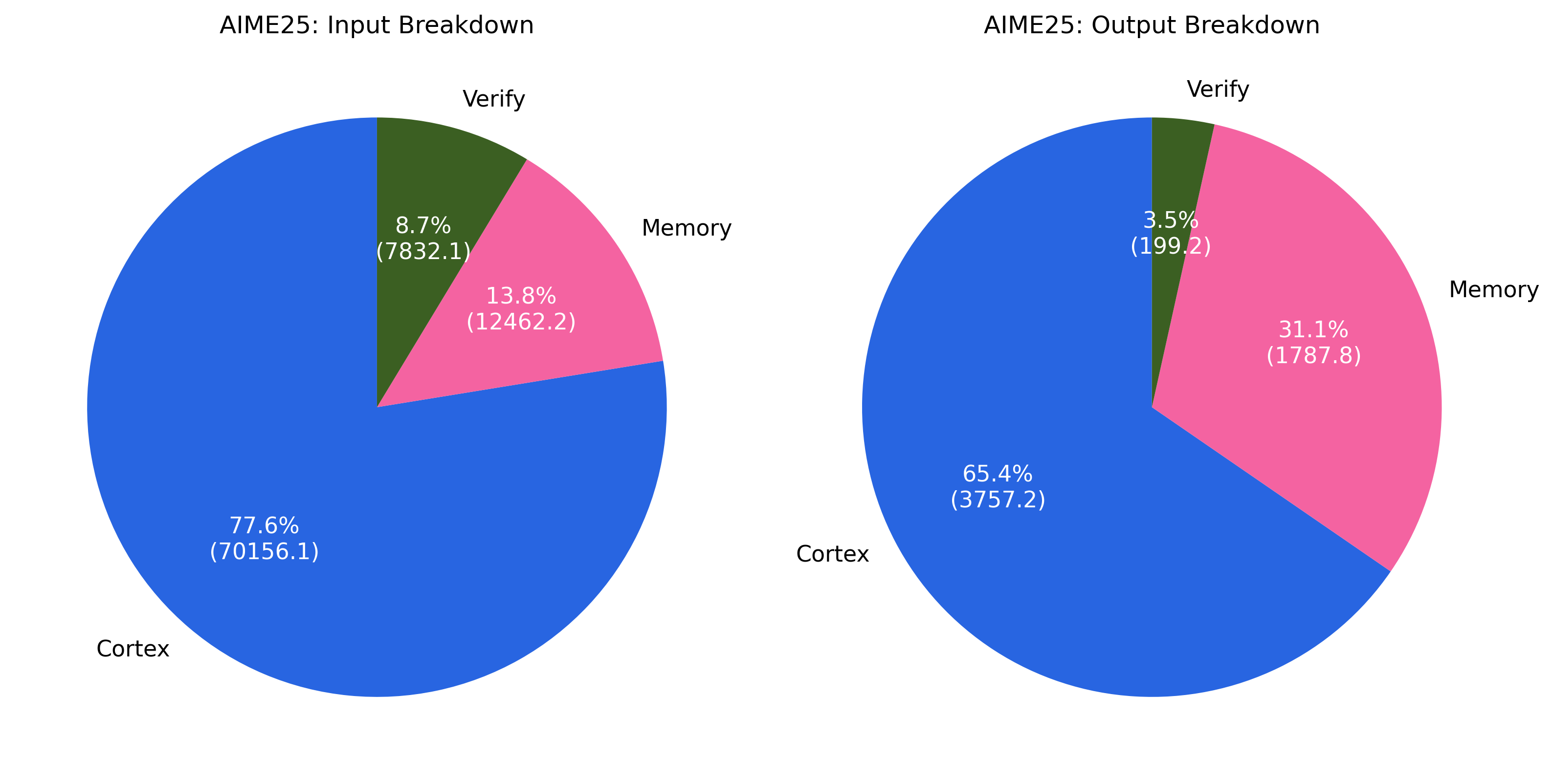}
    \caption{
    Input and output token breakdown of HippoSpark on AIME25. The additional budget is mainly spent on main reasoning and retrieval-conditioned guidance, with verification acting as a smaller pre-commitment check.
    }
    \label{fig:app_inference_aime25_pies}
\end{figure}
\FloatBarrier

\textbf{Overall, HippoSpark trades higher inference cost for stronger reasoning reliability.} Most of the additional budget is spent on maintaining the explicit reasoning state, adapting retrieved experience at difficult transitions, and verifying proposed moves before commitment. Thus, the extra tokens are not merely longer generations, but the cost of making the reasoning process controlled, inspectable, and more robust.

\newpage
\section*{NeurIPS Paper Checklist}

\begin{enumerate}

\item {\bf Claims}
    \item[] Question: Do the main claims made in the abstract and introduction accurately reflect the paper's contributions and scope?
    \item[] Answer: \answerYes{} 
    \item[] Justification: The abstract and introduction state that \textsc{HippoSpark} targets experience systems for LLM reasoning, with on-demand experience utilization and bottleneck-focused actionable experience construction. The claims are supported by experiments on mathematical, scientific, and code reasoning benchmarks and are scoped to these evaluated settings.
    \item[] Guidelines:
    \begin{itemize}
        \item The answer \answerNA{} means that the abstract and introduction do not include the claims made in the paper.
        \item The abstract and/or introduction should clearly state the claims made, including the contributions made in the paper and important assumptions and limitations. A \answerNo{} or \answerNA{} answer to this question will not be perceived well by the reviewers. 
        \item The claims made should match theoretical and experimental results, and reflect how much the results can be expected to generalize to other settings. 
        \item It is fine to include aspirational goals as motivation as long as it is clear that these goals are not attained by the paper. 
    \end{itemize}

\item {\bf Limitations}
    \item[] Question: Does the paper discuss the limitations of the work performed by the authors?
    \item[] Answer: \answerYes{} 
    \item[] Justification: We discuss the limitations of \textsc{HippoSpark} in Appendix~\ref{app:limitations}, including additional inference and construction overhead, as well as the lack of explicit lifecycle management for stored experience.
    \item[] Guidelines:
    \begin{itemize}
        \item The answer \answerNA{} means that the paper has no limitation while the answer \answerNo{} means that the paper has limitations, but those are not discussed in the paper. 
        \item The authors are encouraged to create a separate ``Limitations'' section in their paper.
        \item The paper should point out any strong assumptions and how robust the results are to violations of these assumptions (e.g., independence assumptions, noiseless settings, model well-specification, asymptotic approximations only holding locally). The authors should reflect on how these assumptions might be violated in practice and what the implications would be.
        \item The authors should reflect on the scope of the claims made, e.g., if the approach was only tested on a few datasets or with a few runs. In general, empirical results often depend on implicit assumptions, which should be articulated.
        \item The authors should reflect on the factors that influence the performance of the approach. For example, a facial recognition algorithm may perform poorly when image resolution is low or images are taken in low lighting. Or a speech-to-text system might not be used reliably to provide closed captions for online lectures because it fails to handle technical jargon.
        \item The authors should discuss the computational efficiency of the proposed algorithms and how they scale with dataset size.
        \item If applicable, the authors should discuss possible limitations of their approach to address problems of privacy and fairness.
        \item While the authors might fear that complete honesty about limitations might be used by reviewers as grounds for rejection, a worse outcome might be that reviewers discover limitations that aren't acknowledged in the paper. The authors should use their best judgment and recognize that individual actions in favor of transparency play an important role in developing norms that preserve the integrity of the community. Reviewers will be specifically instructed to not penalize honesty concerning limitations.
    \end{itemize}

\item {\bf Theory assumptions and proofs}
    \item[] Question: For each theoretical result, does the paper provide the full set of assumptions and a complete (and correct) proof?
    \item[] Answer: \answerNA{} 
    \item[] Justification: The paper does not present formal theoretical results, theorems, or proof-based claims. Its contributions are methodological and empirical, supported by experiments and analyses rather than theoretical proofs.
    \item[] Guidelines:
    \begin{itemize}
        \item The answer \answerNA{} means that the paper does not include theoretical results. 
        \item All the theorems, formulas, and proofs in the paper should be numbered and cross-referenced.
        \item All assumptions should be clearly stated or referenced in the statement of any theorems.
        \item The proofs can either appear in the main paper or the supplemental material, but if they appear in the supplemental material, the authors are encouraged to provide a short proof sketch to provide intuition. 
        \item Inversely, any informal proof provided in the core of the paper should be complemented by formal proofs provided in appendix or supplemental material.
        \item Theorems and Lemmas that the proof relies upon should be properly referenced. 
    \end{itemize}

    \item {\bf Experimental result reproducibility}
    \item[] Question: Does the paper fully disclose all the information needed to reproduce the main experimental results of the paper to the extent that it affects the main claims and/or conclusions of the paper (regardless of whether the code and data are provided or not)?
    \item[] Answer: \answerYes{} 
    \item[] Justification: We provide the experimental details and evaluation protocols in Appendices~\ref{app:method_details} and~\ref{app:implementation}, and release code at \url{https://anonymous.4open.science/r/HippoSpark/}.
    \item[] Guidelines: 
    \begin{itemize}
        \item The answer \answerNA{} means that the paper does not include experiments.
        \item If the paper includes experiments, a \answerNo{} answer to this question will not be perceived well by the reviewers: Making the paper reproducible is important, regardless of whether the code and data are provided or not.
        \item If the contribution is a dataset and\slash or model, the authors should describe the steps taken to make their results reproducible or verifiable. 
        \item Depending on the contribution, reproducibility can be accomplished in various ways. For example, if the contribution is a novel architecture, describing the architecture fully might suffice, or if the contribution is a specific model and empirical evaluation, it may be necessary to either make it possible for others to replicate the model with the same dataset, or provide access to the model. In general. releasing code and data is often one good way to accomplish this, but reproducibility can also be provided via detailed instructions for how to replicate the results, access to a hosted model (e.g., in the case of a large language model), releasing of a model checkpoint, or other means that are appropriate to the research performed.
        \item While NeurIPS does not require releasing code, the conference does require all submissions to provide some reasonable avenue for reproducibility, which may depend on the nature of the contribution. For example
        \begin{enumerate}
            \item If the contribution is primarily a new algorithm, the paper should make it clear how to reproduce that algorithm.
            \item If the contribution is primarily a new model architecture, the paper should describe the architecture clearly and fully.
            \item If the contribution is a new model (e.g., a large language model), then there should either be a way to access this model for reproducing the results or a way to reproduce the model (e.g., with an open-source dataset or instructions for how to construct the dataset).
            \item We recognize that reproducibility may be tricky in some cases, in which case authors are welcome to describe the particular way they provide for reproducibility. In the case of closed-source models, it may be that access to the model is limited in some way (e.g., to registered users), but it should be possible for other researchers to have some path to reproducing or verifying the results.
        \end{enumerate}
    \end{itemize}

\item {\bf Open access to data and code}
    \item[] Question: Does the paper provide open access to the data and code, with sufficient instructions to faithfully reproduce the main experimental results, as described in supplemental material?
    \item[] Answer: \answerYes{} 
    \item[] Justification: We release the code at \url{https://anonymous.4open.science/r/HippoSpark/} and use publicly available benchmarks. Instructions for reproducing the main experimental results are provided in the supplemental material.
    \item[] Guidelines:
    \begin{itemize}
        \item The answer \answerNA{} means that paper does not include experiments requiring code.
        \item Please see the NeurIPS code and data submission guidelines (\url{https://neurips.cc/public/guides/CodeSubmissionPolicy}) for more details.
        \item While we encourage the release of code and data, we understand that this might not be possible, so \answerNo{} is an acceptable answer. Papers cannot be rejected simply for not including code, unless this is central to the contribution (e.g., for a new open-source benchmark).
        \item The instructions should contain the exact command and environment needed to run to reproduce the results. See the NeurIPS code and data submission guidelines (\url{https://neurips.cc/public/guides/CodeSubmissionPolicy}) for more details.
        \item The authors should provide instructions on data access and preparation, including how to access the raw data, preprocessed data, intermediate data, and generated data, etc.
        \item The authors should provide scripts to reproduce all experimental results for the new proposed method and baselines. If only a subset of experiments are reproducible, they should state which ones are omitted from the script and why.
        \item At submission time, to preserve anonymity, the authors should release anonymized versions (if applicable).
        \item Providing as much information as possible in supplemental material (appended to the paper) is recommended, but including URLs to data and code is permitted.
    \end{itemize}

\item {\bf Experimental setting/details}
    \item[] Question: Does the paper specify all the training and test details (e.g., data splits, hyperparameters, how they were chosen, type of optimizer) necessary to understand the results?
    \item[] Answer: \answerYes{} 
    \item[] Justification: We do not train or fine-tune backbone models. We report the experimental setup, model settings, inference configurations, and evaluation protocols in Appendices~\ref{app:method_details} and~\ref{app:implementation}.
    \item[] Guidelines:
    \begin{itemize}
        \item The answer \answerNA{} means that the paper does not include experiments.
        \item The experimental setting should be presented in the core of the paper to a level of detail that is necessary to appreciate the results and make sense of them.
        \item The full details can be provided either with the code, in appendix, or as supplemental material.
    \end{itemize}

\item {\bf Experiment statistical significance}
    \item[] Question: Does the paper report error bars suitably and correctly defined or other appropriate information about the statistical significance of the experiments?
    \item[] Answer: \answerNo{} 
    \item[] Justification: We report averaged results across repeated runs where applicable, but do not include formal significance tests, confidence intervals, or error bars. Our main results are reported using the standard aggregate metrics of the evaluated LLM reasoning benchmarks.
    \item[] Guidelines:
    \begin{itemize}
        \item The answer \answerNA{} means that the paper does not include experiments.
        \item The authors should answer \answerYes{} if the results are accompanied by error bars, confidence intervals, or statistical significance tests, at least for the experiments that support the main claims of the paper.
        \item The factors of variability that the error bars are capturing should be clearly stated (for example, train/test split, initialization, random drawing of some parameter, or overall run with given experimental conditions).
        \item The method for calculating the error bars should be explained (closed form formula, call to a library function, bootstrap, etc.)
        \item The assumptions made should be given (e.g., Normally distributed errors).
        \item It should be clear whether the error bar is the standard deviation or the standard error of the mean.
        \item It is OK to report 1-sigma error bars, but one should state it. The authors should preferably report a 2-sigma error bar than state that they have a 96\% CI, if the hypothesis of Normality of errors is not verified.
        \item For asymmetric distributions, the authors should be careful not to show in tables or figures symmetric error bars that would yield results that are out of range (e.g., negative error rates).
        \item If error bars are reported in tables or plots, the authors should explain in the text how they were calculated and reference the corresponding figures or tables in the text.
    \end{itemize}

\item {\bf Experiments compute resources}
    \item[] Question: For each experiment, does the paper provide sufficient information on the computer resources (type of compute workers, memory, time of execution) needed to reproduce the experiments?
    \item[] Answer: \answerYes{} 
    \item[] Justification: We provide the compute resources used in our experiments in Appendix~\ref{app:compute_resources}.
    \item[] Guidelines:
    \begin{itemize}
        \item The answer \answerNA{} means that the paper does not include experiments.
        \item The paper should indicate the type of compute workers CPU or GPU, internal cluster, or cloud provider, including relevant memory and storage.
        \item The paper should provide the amount of compute required for each of the individual experimental runs as well as estimate the total compute. 
        \item The paper should disclose whether the full research project required more compute than the experiments reported in the paper (e.g., preliminary or failed experiments that didn't make it into the paper). 
    \end{itemize}
    
\item {\bf Code of ethics}
    \item[] Question: Does the research conducted in the paper conform, in every respect, with the NeurIPS Code of Ethics \url{https://neurips.cc/public/EthicsGuidelines}?
    \item[] Answer: \answerYes{} 
    \item[] Justification: The research uses public reasoning benchmarks, open-weight models, and closed-source models accessed through their official APIs. It does not involve private or sensitive data, human-subject studies, or deployment in high-stakes decision-making settings.
    \item[] Guidelines:
    \begin{itemize}
        \item The answer \answerNA{} means that the authors have not reviewed the NeurIPS Code of Ethics.
        \item If the authors answer \answerNo, they should explain the special circumstances that require a deviation from the Code of Ethics.
        \item The authors should make sure to preserve anonymity (e.g., if there is a special consideration due to laws or regulations in their jurisdiction).
    \end{itemize}

\item {\bf Broader impacts}
    \item[] Question: Does the paper discuss both potential positive societal impacts and negative societal impacts of the work performed?
    \item[] Answer: \answerYes{} 
    \item[] Justification: We discuss potential positive and negative broader impacts of experience-guided LLM reasoning in Appendix~\ref{app:broader_impacts}.
    \item[] Guidelines:
    \begin{itemize}
        \item The answer \answerNA{} means that there is no societal impact of the work performed.
        \item If the authors answer \answerNA{} or \answerNo, they should explain why their work has no societal impact or why the paper does not address societal impact.
        \item Examples of negative societal impacts include potential malicious or unintended uses (e.g., disinformation, generating fake profiles, surveillance), fairness considerations (e.g., deployment of technologies that could make decisions that unfairly impact specific groups), privacy considerations, and security considerations.
        \item The conference expects that many papers will be foundational research and not tied to particular applications, let alone deployments. However, if there is a direct path to any negative applications, the authors should point it out. For example, it is legitimate to point out that an improvement in the quality of generative models could be used to generate Deepfakes for disinformation. On the other hand, it is not needed to point out that a generic algorithm for optimizing neural networks could enable people to train models that generate Deepfakes faster.
        \item The authors should consider possible harms that could arise when the technology is being used as intended and functioning correctly, harms that could arise when the technology is being used as intended but gives incorrect results, and harms following from (intentional or unintentional) misuse of the technology.
        \item If there are negative societal impacts, the authors could also discuss possible mitigation strategies (e.g., gated release of models, providing defenses in addition to attacks, mechanisms for monitoring misuse, mechanisms to monitor how a system learns from feedback over time, improving the efficiency and accessibility of ML).
    \end{itemize}
    
\item {\bf Safeguards}
    \item[] Question: Does the paper describe safeguards that have been put in place for responsible release of data or models that have a high risk for misuse (e.g., pre-trained language models, image generators, or scraped datasets)?
    \item[] Answer: \answerNA{} 
    \item[] Justification: This work does not release new pretrained models, image generators, scraped datasets, or other high-risk assets. The released materials are limited to code, prompts, configurations, and evaluation resources for public reasoning benchmarks.
    \item[] Guidelines:
    \begin{itemize}
        \item The answer \answerNA{} means that the paper poses no such risks.
        \item Released models that have a high risk for misuse or dual-use should be released with necessary safeguards to allow for controlled use of the model, for example by requiring that users adhere to usage guidelines or restrictions to access the model or implementing safety filters. 
        \item Datasets that have been scraped from the Internet could pose safety risks. The authors should describe how they avoided releasing unsafe images.
        \item We recognize that providing effective safeguards is challenging, and many papers do not require this, but we encourage authors to take this into account and make a best faith effort.
    \end{itemize}

\item {\bf Licenses for existing assets}
    \item[] Question: Are the creators or original owners of assets (e.g., code, data, models), used in the paper, properly credited and are the license and terms of use explicitly mentioned and properly respected?
    \item[] Answer: \answerYes{} 
    \item[] Justification: We credit the existing assets used in this work and report their licenses or terms of use in Appendix~\ref{app:implementation}.
    \item[] Guidelines:
    \begin{itemize}
        \item The answer \answerNA{} means that the paper does not use existing assets.
        \item The authors should cite the original paper that produced the code package or dataset.
        \item The authors should state which version of the asset is used and, if possible, include a URL.
        \item The name of the license (e.g., CC-BY 4.0) should be included for each asset.
        \item For scraped data from a particular source (e.g., website), the copyright and terms of service of that source should be provided.
        \item If assets are released, the license, copyright information, and terms of use in the package should be provided. For popular datasets, \url{paperswithcode.com/datasets} has curated licenses for some datasets. Their licensing guide can help determine the license of a dataset.
        \item For existing datasets that are re-packaged, both the original license and the license of the derived asset (if it has changed) should be provided.
        \item If this information is not available online, the authors are encouraged to reach out to the asset's creators.
    \end{itemize}

\item {\bf New assets}
    \item[] Question: Are new assets introduced in the paper well documented and is the documentation provided alongside the assets?
    \item[] Answer: \answerYes{} 
    \item[] Justification: We release the code and accompanying documentation at \url{https://anonymous.4open.science/r/HippoSpark/}, including instructions for reproducing the main experiments.
    \item[] Guidelines:
    \begin{itemize}
        \item The answer \answerNA{} means that the paper does not release new assets.
        \item Researchers should communicate the details of the dataset\slash code\slash model as part of their submissions via structured templates. This includes details about training, license, limitations, etc. 
        \item The paper should discuss whether and how consent was obtained from people whose asset is used.
        \item At submission time, remember to anonymize your assets (if applicable). You can either create an anonymized URL or include an anonymized zip file.
    \end{itemize}

\item {\bf Crowdsourcing and research with human subjects}
    \item[] Question: For crowdsourcing experiments and research with human subjects, does the paper include the full text of instructions given to participants and screenshots, if applicable, as well as details about compensation (if any)? 
    \item[] Answer: \answerNA{} 
    \item[] Justification: This work does not involve crowdsourcing experiments or research with human subjects.
    \item[] Guidelines:
    \begin{itemize}
        \item The answer \answerNA{} means that the paper does not involve crowdsourcing nor research with human subjects.
        \item Including this information in the supplemental material is fine, but if the main contribution of the paper involves human subjects, then as much detail as possible should be included in the main paper. 
        \item According to the NeurIPS Code of Ethics, workers involved in data collection, curation, or other labor should be paid at least the minimum wage in the country of the data collector. 
    \end{itemize}

\item {\bf Institutional review board (IRB) approvals or equivalent for research with human subjects}
    \item[] Question: Does the paper describe potential risks incurred by study participants, whether such risks were disclosed to the subjects, and whether Institutional Review Board (IRB) approvals (or an equivalent approval/review based on the requirements of your country or institution) were obtained?
    \item[] Answer: \answerNA{} 
    \item[] Justification: This work does not involve research with human subjects, so IRB approval or equivalent review is not applicable.
    \item[] Guidelines:
    \begin{itemize}
        \item The answer \answerNA{} means that the paper does not involve crowdsourcing nor research with human subjects.
        \item Depending on the country in which research is conducted, IRB approval (or equivalent) may be required for any human subjects research. If you obtained IRB approval, you should clearly state this in the paper. 
        \item We recognize that the procedures for this may vary significantly between institutions and locations, and we expect authors to adhere to the NeurIPS Code of Ethics and the guidelines for their institution. 
        \item For initial submissions, do not include any information that would break anonymity (if applicable), such as the institution conducting the review.
    \end{itemize}

\item {\bf Declaration of LLM usage}
    \item[] Question: Does the paper describe the usage of LLMs if it is an important, original, or non-standard component of the core methods in this research? Note that if the LLM is used only for writing, editing, or formatting purposes and does \emph{not} impact the core methodology, scientific rigor, or originality of the research, declaration is not required.
    \item[] Answer: \answerNA{} 
    \item[] Justification: LLMs were not used as an external tool to generate core methodology, data, or scientific claims; any use was limited to writing, editing, or formatting. We provide a statement in Appendix~\ref{app:llm_usage}.
    \item[] Guidelines:
    \begin{itemize}
        \item The answer \answerNA{} means that the core method development in this research does not involve LLMs as any important, original, or non-standard components.
        \item Please refer to our LLM policy in the NeurIPS handbook for what should or should not be described.
    \end{itemize}

\end{enumerate}

\end{document}